\documentclass[a4paper,fleqn]{cas-dc}
\usepackage{float}
\usepackage{array}
\usepackage{arydshln}
\usepackage{graphicx}
\usepackage{caption}
\usepackage{subcaption}
\usepackage{xcolor}
\usepackage{multirow}
\usepackage{tabularx}
\usepackage[numbers, sort&compress]{natbib}
\usepackage{chngcntr}  
\usepackage{float}
\usepackage{placeins}
\usepackage{xurl}
\usepackage{hyperref}
\usepackage{breakurl}
\usepackage{amssymb}
\usepackage{amsmath}
\usepackage{chngcntr}
\usepackage{lastpage}
\begin{document}
\renewcommand{\lastpage}{\pageref{LastPage}}
\setlength{\arrayrulewidth}{0.1pt}
\let\WriteBookmarks\relax
\def\floatpagepagefraction{1}
\def\textpagefraction{.001}

\shorttitle{Conditional Dual-Output Diffusion Transformer (DualDiT)}

\shortauthors{F. García-Torres et~al.}

\title [mode = title]{DualDiT: A Conditional Dual-Output Diffusion Transformer for Joint OCT Image and Segmentation Mask Generation}       
\author[1]{Fernando Garc\'ia-Torres}[orcid = 0000-0002-5337-8774]
\ead{fergart1@upv.es}
\cormark[1]

\credit{Conceptualization, Methodology, Data curation, Investigation, Formal analysis, Writing and Visualization}

\author[1,2]{Roc\'io del Amor}
\credit{Conceptualization, Methodology, Investigation, Formal analysis, Writing and Visualization}

\author[1]{Sandra Morales}
\credit{Conceptualization, Methodology, Investigation, Formal analysis, Writing and Visualization}

\author[3]{\'Alvaro Barroso}
\credit{Data curation, Review and editing}

\author[4]{Peter Heiduschka}
\credit{Data curation}

\author[3]{Björn Kemper}
\credit{Data curation, Review and editing}

\author[1,2]{Valery Naranjo}
\credit{Conceptualization, Review and editing, and Supervision}

\affiliation[1]{organization={Instituto Universitario de                Investigaci\'on en Tecnolog\'ia Centrada en el Ser Humano (HUMAN-tech), Universitat Politècnica de València},
            addressline={Camino de Vera, s/n}, 
            city={Valencia},
            postcode={46022}, 
            state={Comunidad Valenciana},
            country={Spain}}

\affiliation[2]{organization={Artikode Intelligence S.L},
            addressline={Camino de Vera, s/n}, 
            city={Valencia},
            postcode={46022}, 
            state={Comunidad Valenciana},
            country={Spain}}

\affiliation[3]{organization={Biomedical Technology Center of the Medical Faculty, University of Muenster},
            addressline={Mendelstraße 17},
            city={Münster},
            postcode={48149},
            state={North Rhine-Westphalia},
            country={Germany}}

\affiliation[4]{organization={Department of Ophthalmology, University of Muenster Medical Centre},
            addressline={Domagkstraße 15},
            city={Münster},
            postcode={48149},
            state={North Rhine-Westphalia},
            country={Germany}}

\cortext[1]{Corresponding author. Tel.: +34 664 883 363}
\begin{abstract}
\textbf{\textbf{Background and Objective:}} Generating realistic medical images along with anatomically accurate segmentation masks is a promising approach to address the shortage of annotated data in medical imaging, particularly in optical coherence tomography (OCT) of mouse eyes, where manual retinal layer delineation is very labour-intensive due to tiny structures and demands on expert knowledge, resulting in scarce available datasets. While diffusion models have shown strong performance in medical image synthesis, joint image-mask generation has primarily relied on U-Net-based denoisers, leaving diffusion transformers largely unexplored in this context.

\textbf{Methods:} We propose a conditional dual-output Diffusion Transformer (DualDiT) for the simultaneous synthesis of OCT B-scans and segmentation masks of the upper retinal cell layers (URCL) of ex vivo mouse retina. DualDiT encodes both modalities into a shared latent space via a pretrained variational auto-encoder, concatenates their latent representations, and performs conditional diffusion over the resulting joint tensor. We compared DualDiT against two adapted denoising diffusion baselines: a pixel-space conditional model (DDPM) and a latent-space conditional model (LDM). Generative quality was assessed using distribution similarity metrics (Fréchet Inception Distance, FID; spatial FID, sFID), practical utility was evaluated through synthetic data augmentation for downstream U-Net segmentation, and perceptual realism was assessed by a panel of three domain experts.

\textbf{Results:} DualDiT achieved the best generative quality, obtaining a FID of 56.14 and sFID of 114.35, outperforming DDPM (FID 164.55, sFID 254.52) and LDM (FID 102.21, sFID 150.66).  In the expert evaluation, panels incorrectly classified, on average, 46\% of synthetic samples as real and 42\% of real samples as synthetic. The additional use of DualDiT-generated images and masks increases Dice and IoU scores, as evaluated on a held-out test set of the segmentation model.

\textbf{Conclusions:} DualDiT demonstrates that transformer-based diffusion models can effectively learn the joint distribution of OCT images and segmentation masks, surpassing conventional DDPM- and LDM-based baselines in generative fidelity, downstream segmentation utility, and perceptual realism, highlighting its potential as a data augmentation strategy for annotation-scarce medical imaging applications.
\end{abstract}


\begin{highlights}
\item DualDiT generates simultaneously OCT images and anatomically aligned masks.
\item A shared latent space captures dependencies between images and masks.
\item DualDiT achieves the lowest FID and sFID among the evaluated models.
\item DualDiT-generated data improve segmentation performance on real OCT scans.
\item Experts misclassified 46\% of synthetic DualDiT OCT samples as real.
\end{highlights}


\maketitle

\section{Introduction}
\label{1_introduction}

Generative artificial intelligence (GenAI) is profoundly transforming the field of medical imaging. Generative models can learn the underlying distribution of complex biomedical data and generate images that are not only anatomically realistic but also include clinically and diagnostically relevant image content. Methods such as Generative Adversarial Networks (GANs), Variational Autoencoders (VAEs), Diffusion models and Transformers have been primarily employed for data augmentation, enhancing the diversity of training datasets and improving the generalisation of diagnostic algorithms when annotated data were scarce \cite{jimaging9040081, celard_survey_2023, chlap_review_2021}. Generative techniques have enabled researchers to synthesise anatomically coherent images that replicate key visual patterns across different imaging modalities (X-ray, histology, magnetic resonance, computed tomography, etc.), supporting tasks such as lesion detection and segmentation \cite{celard_survey_2023, islam_generative_2024}. These early applications established the foundation for using GenAI as a data-centric strategy to overcome the limitations of small and imbalanced datasets in healthcare \cite{celard_survey_2023, chlap_review_2021}. More recent developments are pushing this paradigm even further. Advances in diffusion models and foundation architectures have enabled high-fidelity, controllable image synthesis, where models learn general-purpose latent representations that capture the essence of complex biomedical data \cite{oulmalme_systematic_2025}. 

GANs are among the earliest and most influential approaches to generative modelling. They consist of a generator–discriminator pair trained in an adversarial setting, where the generator learns to produce realistic images while the discriminator distinguishes them from real samples \cite{islam_generative_2024}.  In medical imaging, these models have been successfully applied to data augmentation \cite{waheed2020covidgan}, modality translation \cite{isola_image--image_2017}, denoising \cite{yang2017dagan}, and super-resolution \cite{yang2017dagan}, improving image realism and diagnostic performance \cite{islam_generative_2024}. VAEs introduced a probabilistic framework that encodes input data into a latent distribution and reconstructs images through sampling \cite{kingma_introduction_2019}. Variants such as $\beta$-VAE (for disentangled representations), Conditional VAE (CVAE), and hybrid VAE–GAN architectures have expanded their use in medical imaging \cite{rguibi_medical_2023}. VAEs are valuable for anomaly detection, controllable synthesis, and representation learning, offering explicit uncertainty estimation—a key property for clinical interpretation \cite{rguibi_medical_2023}. Furthermore, their structured latent space and probabilistic nature make them particularly suitable for integration into modern diffusion-based frameworks.

Diffusion models represent the current state of the art in generative medical imaging. Unlike GANs and VAEs, they avoid mode collapse and training instability through a denoising score-matching objective, and produce sharper images with finer anatomical detail ~\cite{shi_diffusion_2025}. Their superiority has been empirically validated by Müller-Franzes et al.~\cite{muller2023multimodal}, who demonstrated substantially lower FID scores and greater diversity than GAN-based models on eye fundus, chest X-ray, and histopathology datasets. Diffusion models progressively denoise random noise into coherent images, achieving remarkable stability and visual fidelity. Notable architectures include Denoising Diffusion Probabilistic Models (DDPMs) \cite{ho_denoising_2020}, Latent Diffusion Models (LDMs) \cite{rombach_high-resolution_2022-1}, and Guided Diffusion variants for conditioning on modality, anatomy, or clinical priors. These models have achieved outstanding results in denoising, reconstruction, and cross-modality translation, often surpassing GAN-based methods \cite{wang_diffusion_2025}. Transformer-based architectures are the latest addition to generative imaging. Leveraging self-attention mechanisms, they model long-range spatial and contextual dependencies that convolutional models struggle to capture. Examples include Vision Transformers (ViT), TransGAN, and Diffusion Transformers (DiT), which combine attention modules with generative backbones to improve global coherence \cite{oulmalme_systematic_2025}. In medical imaging, transformers are being explored for conditional and multimodal generation, integrating imaging data with metadata or textual information such as clinical reports \cite{oulmalme_systematic_2025, chataut_generative_2025}. While research in this area is still emerging, transformer-based generators are expected to play a pivotal role in the next generation of foundation models, enabling large-scale, cross-modality synthesis and clinically interpretable generative pipelines.

In medical applications, optical coherence tomography (OCT) is a widely used imaging modality in ophthalmology, providing high-resolution cross-sectional images of retina. OCT enables clinicians to visualise the layered structure of the retina and is essential for diagnosing and monitoring diseases such as age-related macular degeneration, diabetic retinopathy, and glaucoma \cite{fercher2003optical}. In particular, the thickness between the embedding medium-retina interface and the interface of the inner plexiform layer (IPL) and inner nuclear layer(INL), from here on denoted as the upper retinal cell layers (URCL), has been used to study glaucoma-like changes in the retina \cite{mayer2010retinal}. Consequently, the automated segmentation of retinal layers from OCT scans is a critical task for quantitative disease assessment, as accurate delineation of retinal cell layers helps clinicians identify structural changes associated with early pathological processes \cite{morales_retinal_2021, amor_towards_2019}. 

While most publicly available OCT datasets consist of human retinal images, a highly research-relevant but underexplored domain is ex vivo mouse retinal OCT imaging. Mouse models are widely used in preclinical research to simulate retinal diseases such as diabetic retinopathy and glaucoma, as their retinal architecture shares key structural similarities with the human retina \cite{allen2020vivo}. Ex vivo OCT imaging offers notable advantages over in vivo imaging as it enables highly reproducible acquisitions across different OCT systems and experimental conditions \cite{tschernig2013elegant}. In recent work, methodologies have been developed for preserving mouse retinas in resin, enabling durable ex vivo OCT imaging that maintains biological properties and enhances reproducibility across different OCT systems \cite{barroso_durable_2024}. However, unlike human OCT data, no large-scale public databases exist for ex vivo mouse retinal OCT images, and their segmentation poses additional challenges due to tiny structures and differences in layer appearance compared with human in vivo retinas \cite{morales_retinal_2021}, differences in scale across embedding media, and limited availability of expert annotations. These particularities make manual annotation of mouse retinal layers especially time-consuming and reliant on highly specialised expert knowledge \cite{garcia2024using}, further restricting the development of robust deep learning-based segmentation models. These challenges make OCT an ideal domain for generative models capable of synthesising realistic images and their corresponding segmentation masks.

In this work, we introduce a conditional dual-output Diffusion Transformer (DualDiT) framework for joint image and segmentation mask generation. This framework can synthesise OCT retinal images from ex vivo mouse retina across different preservation media (\textit{physiological} and \textit{resin}), along with their corresponding URCL segmentation masks. To the best of the authors’ knowledge, this is the first work to formulate a DiT-based framework for paired mouse ex vivo OCT retinal image and mask synthesis, enabling the simultaneous generation of anatomically aligned images and segmentation annotations conditioned on the preservation medium domain. The main contributions of this paper are summarized as follows: (i) we introduce DualDiT, a novel DiT-based generative framework that extends diffusion transformers beyond single-output synthesis towards conditional paired image–mask generation through a unified dual-output formulation; (ii) we demonstrate its ability to synthesise domain-conditioned mouse ex vivo OCT retinal images together with anatomically consistent URCL segmentation masks; (iii) we provide a comprehensive quantitative evaluation of the generated data, including image realism assessment through the Fréchet Inception Distance (FID) and comparison against state-of-the-art diffusion-based generative frameworks; and (iv) we validate the practical and anatomical relevance of the generated image–mask pairs by showing their effectiveness as a data augmentation strategy in a downstream segmentation task, with improvements in Dice coefficient and Intersection over Union (IoU), and by conducting a systematic expert-based assessment of anatomical fidelity and perceived reliability of clinical and diagnostic relevant content.

\section{Related Work}
\label{2_related_work}

\subsection{Diffusion Models for OCT image synthesis and processing}

Diffusion models were introduced as Denoising Diffusion Probabilistic Models (DDPM) by Ho et al. in 2020 \cite{ho_denoising_2020}. The application of diffusion models to OCT has gained significant momentum, addressing inherent challenges such as speckle noise, scarcity of high-quality data, and the need for accurate structural synthesis. For example, OCTDiff \cite{tian_octdiff_2026} employs a bridged diffusion model to facilitate super-resolution in portable OCT devices, effectively reducing the quality gap with clinical-grade systems. This is complemented by physics-informed diffusion models \cite{abbasi_physics-informed_2025}, which incorporate the underlying optics of OCT acquisition into the generative process to achieve high-fidelity reconstruction. Beyond posterior segment imaging, diffusion models have also been successfully applied to remove noise and generate anterior segment (AS-OCT) images \cite{ahmed_denoising_2024}.

The primary motivation for synthetic OCT generation is often to improve subsequent clinical tasks. RetiDiff \cite{li_retidiff_2026} and other DDPM-based frameworks \cite{wu_retinal_2024} have demonstrated that generating synthetic scans with corresponding layer maps can significantly improve segmentation performance. This synthesis for segmentation paradigm has been extended to specific biomarkers. For example, recent benchmark tests show that diffusion models outperform traditional architectures in detecting fluid-filled regions for retinal analysis \cite{du_benchmarking_2025}. 

Although DDPMs achieve high-quality synthesis, operating directly in pixel space is computationally expensive due to the high dimensionality of images. To handle the high dimensionality of medical data without prohibitive costs, several studies have adopted Latent Diffusion Models (LDMs). Cascaded amortised LDMs \cite{huang_memory-efficient_2024} have been proposed for efficient in-memory synthesis of high-resolution human retinal OCT volumes. The versatility of latent space also enables multimodal tasks, such as translation from standard OCT to optical coherence tomography angiography (OCT-A) \cite{badhon_diffusion_2025}. These frameworks are particularly effective in diagnosing neovascularisation, as they provide high-resolution, biologically plausible synthetic OCT-A scans.

Finally, unified frameworks such as DiffusionDCI \cite{yang_diffusiondci_2024} represent the state of the art in dynamic OCT imaging, offering a single model for both generation and segmentation, suggesting a shift towards more integrated generative-discriminative architectures in ophthalmology. In contrast to these earlier works, which focus exclusively on human OCT data, our proposed DualDiT framework addresses the underexplored setting of mouse ex vivo OCT imaging, where no large-scale public datasets exist. Furthermore, unlike in vivo acquisition, ex vivo imaging requires tissue embedding, and domain differences across embedding media introduce additional challenges for image synthesis that DualDiT explicitly addresses.

\subsection{Diffusion models for joint image–mask generation}

While diffusion models were originally designed to generate a single image, they have recently been extended to structured output generation, particularly for joint image and segmentation mask synthesis. This setting is especially relevant in domains where annotated data is scarce or costly to obtain, such as medical imaging. Instead of generating images alone, these approaches aim to model the joint distribution $p_\theta(x, m)$, where $x$ denotes the image and $m$ its associated pixel-level label map.

Several recent works explicitly address simultaneous image-mask generation. In the field of satellite image processing, Toker et al. in \cite{toker_satsynth_2024} leverage DDPM with conditional super-resolution to augment aerial image-mask pairs for semantic segmentation. Mao et al. in \cite{mao_medsegfactory_2025} extend this idea by incorporating text guidance to generate paired medical images and masks, enabling more flexible, controllable data synthesis via a dual-stream diffusion with cross-attention between image and mask streams. Similarly, Frisch et al. \cite{frisch_gauda_2025} focus on uncertainty-guided diffusion-based augmentation for surgical segmentation, training an LDM with learned Vector Quatized-GAN-based latents. Other works explore pixel-level annotation synthesis in different contexts. Wu et al.  generate images with an LDM and derive semantic masks from the model’s cross-attention maps \cite{wu_diffumask_2023}. They exploit the cross-attention maps between text and image in the diffusion model to automatically generate high-resolution, class-specific semantic masks, enabling the training of segmentation models with synthetic data that performs comparably to real data. Li et al. \cite{li_open-vocabulary_2023} explore segmentation conditioned on flexible semantic prompts with LDM, bridging generative modelling and open-vocabulary recognition. 

Complementary approaches investigate segmentation-aware diffusion from different perspectives. Park et al. in \cite{park_seediff_2025} demonstrate that off-the-shelf Stable Diffusion models \cite{rombach_high-resolution_2022-1} can be adapted for seeded mask generation, highlighting the implicit segmentation capabilities encoded within pretrained diffusion backbones. Collectively, these works demonstrate the growing interest in diffusion-based structured generation, where image realism and label consistency must be jointly preserved. Compared to unconditional or purely class-conditional synthesis, joint image-mask generation imposes stronger structural constraints, requiring models to capture fine-grained spatial correspondences between visual content and pixel-level annotations.

In summary, prior work has relied primarily on U-Net-based DDPM and LDM architectures or on attention-derived masks from pretrained LDMs, while diffusion transformers that explicitly generate images and masks simultaneously, to the best of our knowledge, have not been explored.

\section{Methods}
\label{3_methods}

\begin{figure*}[H]
    \centering
    \includegraphics[width=\linewidth]{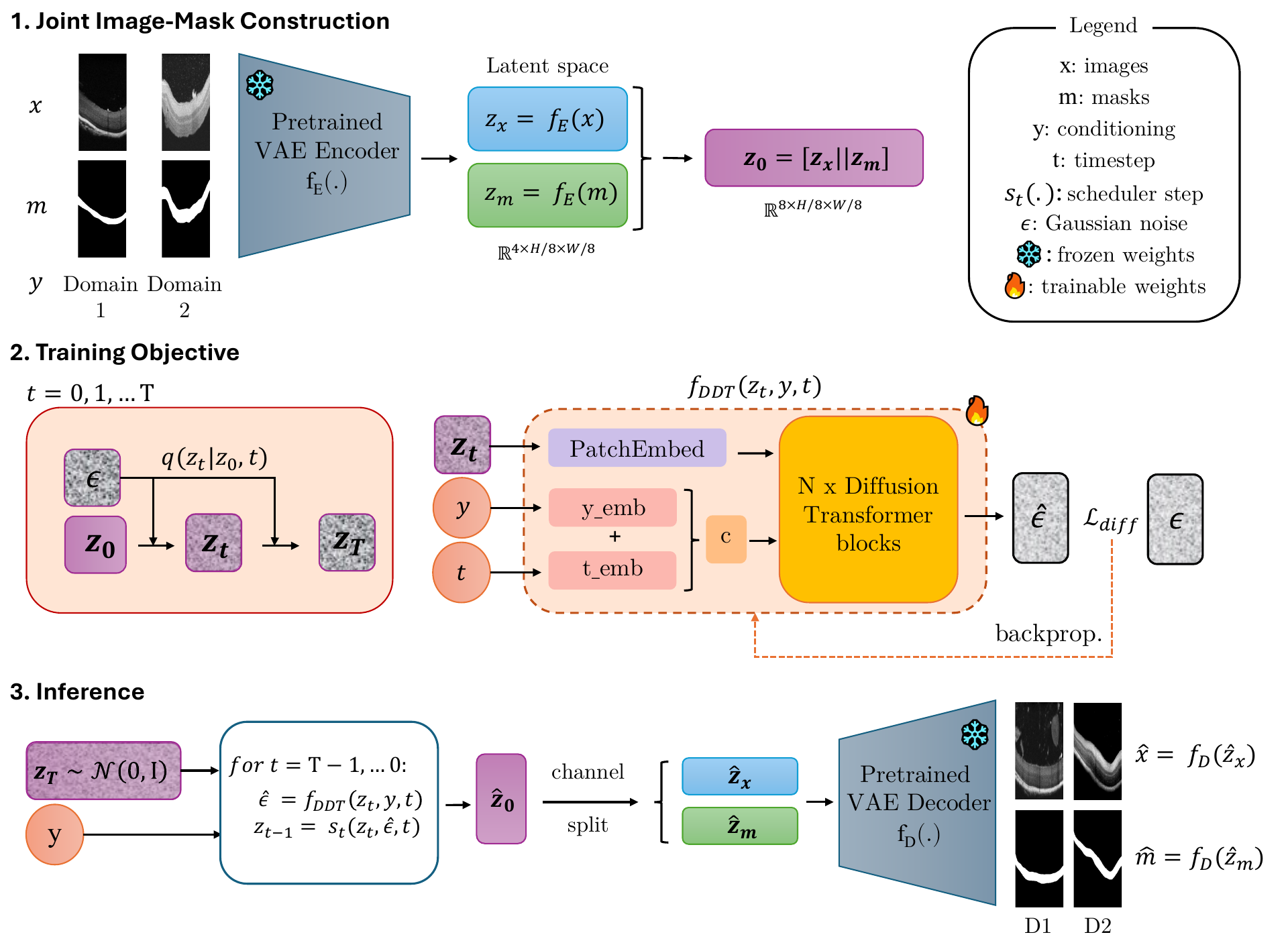}
    \caption{Overview of the DualDiT framework. Paired image and segmentation mask inputs ($x, m$) are encoded ($f_E(.)$) into a shared latent space and concatenated along the channel dimension to form a joint latent tensor $z_0 = [z_x \Vert z_m]$. During training, Gaussian noise 
    ($\epsilon$) is progressively added to $z_0$ over $T$ timesteps following the forward process $q(z_t \mid z_0, t)$. At each timestep $t$, the noisy latent $z_t$, the timestep embedding, and the conditioning class embedding are combined into $c$ and passed to $N$ Diffusion Transformer blocks ($f_{DDT}(z_t, y, t)$) to model cross-modal spatial dependencies between image appearance and mask topology. During training, the network learns to predict the added noise $\hat{\epsilon}$, while during inference, this denoising is applied iteratively starting from $z_T \sim \mathcal{N}(0, I)$ until a clean joint latent $\hat{z_0}$ is recovered. The joint latent $\hat{z_0}$ is then split along the channel dimension into image and mask components and decoded ($f_D(.)$) back to pixel space ($x', m'$).}
    \label{fig:dualdit}
\end{figure*}

Figure~\ref{fig:dualdit} presents an overview of the proposed \textbf{Conditional Dual-Output Diffusion Transformer (DualDiT)}. A detailed description of the different framework components is provided below.

\textbf{Problem formulation:}
The objective of this work is to train a conditional generative model capable of jointly synthesising anatomically consistent OCT images and segmentation masks conditioned on the embedding medium of the retina.  Each OCT image is paired with a segmentation mask delineating the URCL, a structure of interest for downstream analysis, and with a domain label indicating the embedding medium used during ex vivo tissue preparation. Let $\mathcal{D} = \{(x_i, m_i, y_i)\}_{i=1}^{N}$ denote an ex vivo mouse retinal OCT dataset, where $x_i$ $\in \mathbb{R}^{1 \times H \times W}$ represents a greyscale OCT image, $m_i$ $\in \mathbb{R}^{1 \times H \times W}$ its corresponding binary URCL segmentation mask, and $y_i \in \{0,\dots,K-1\}$ the associated class label of each OCT domain. 

\textbf{Joint latent representation learning:} To improve computational efficiency and reduce the dimensionality of the generative process, OCT images and segmentation masks are projected into a latent space using a pretrained VAE on natural images \cite{noauthor_stabilityaisd-vae-ft-ema_nodate}. Given an OCT image $x$ and its corresponding mask $m$, the encoder $f_E(.)$ maps both inputs into latent representations $z_x = f_E(x)$ and $z_m = f_E(m)$, respectively, each of spatial dimensions $\mathbb{R}^{4 \times H/8 \times W/8}$, 
where 4 corresponds to the latent channel dimensionality defined by the VAE 
architecture, and the spatial resolution is downsampled by a factor of 8. Since the VAE encoder and decoder are kept frozen during training, the 
learned latent representations are used as fixed projections. Both latent embeddings are concatenated along the channel dimension to construct a shared latent representation:
\begin{equation}
    z_0 = [z_x \, \Vert \, z_m], \in \mathbb{R}^{8 \times H/8 \times W/8},
\end{equation}
where $\Vert$ denotes channel-wise concatenation, yielding a joint latent tensor of dimensions $\mathbb{R}^{8 \times H/8 \times W/8}$. This joint latent formulation allows the model to learn the structural correspondence between retinal appearance and layer segmentation during the diffusion process while operating in a computationally efficient latent space.

\textbf{Conditional dual-output Diffusion Transformer:} The proposed framework introduces a conditional dual-output Diffusion Transformer architecture for the joint generation of OCT images and segmentation masks. Unlike previous DDPM- and LDM-based approaches relying on 
convolutional denoisers, the proposed model adopts a Transformer backbone to model long-range spatial dependencies through self-attention mechanisms.

The concatenated latent representation $z_0$ is progressively corrupted through a forward diffusion process over $T$ timesteps following $q(z_t \mid z_0, t)$, which denotes the forward diffusion kernel that corrupts the clean latent $z_0$ by adding Gaussian noise $\epsilon \sim \mathcal{N}(0,I)$ according to a variance schedule $\{\bar{\alpha}_t\}_{t=1}^T$, yielding the noisy latent representation $z_t$ at each timestep $t$ \cite{ho_denoising_2020}. The reverse denoising process models the conditional distribution $p_\theta(z_{t-1} \mid z_t, y)$, parameterized by a Transformer-based denoiser $f_{DDT}$ that takes as input the noisy latent $z_t \in \mathbb{R}^{8 \times H \times W}$, the domain label $y \in \{0,\dots,K-1\}$, and the diffusion timestep $t \in \{1,\dots,T\}$, and predicts the noise $\hat{\epsilon}$:
\begin{equation}
    \hat{\epsilon} = f_{DDT}(z_t,\, y,\, t).
\end{equation}

The timestep $t$ is mapped to a continuous vector representation $\mathbf{t}_\text{emb} = \phi_t(t) \in \mathbb{R}^d$ via sinusoidal embeddings followed by a Multilayer Perceptron (MLP), and the domain label $y$ is projected to $\mathbf{y}_\text{emb} = \phi_y(y) \in \mathbb{R}^d$ via a learned embedding table. Both are combined into a single conditioning vector $\mathbf{c} = \mathbf{t}_\text{emb} + \mathbf{y}_\text{emb} \in \mathbb{R}^d$. Following Peebles and Xie~\cite{peebles_scalable_2023}, each of the $N$ Transformer blocks conditions on $\mathbf{c}$ via adaLN-Zero. Each block regresses six modulation parameters (scale, shift, and gate for both the self-attention and MLP sub-layers) directly from $\mathbf{c}$, and applies them residually after layer normalisation. All parameters are zero-initialised so that each block acts as an identity at the start of training. In contrast to conventional DiT frameworks \cite{peebles_scalable_2023} designed for single-image synthesis, the proposed formulation jointly predicts both OCT and segmentation latent representations within a unified denoising trajectory. This dual-output design enforces anatomical consistency between retinal structures and layer masks during generation.

\textbf{Optimisation objective:} Since the VAE encoder and decoder are kept frozen during training, the model parameters are optimised using the standard diffusion noise prediction objective. Given a noisy latent sample $z_t$ and Gaussian noise $\epsilon \sim \mathcal{N}(0,I)$, the optimisation minimises the mean squared error between the predicted and actual noise:
\begin{equation}
    \mathcal{L}_{\text{diff}} =
    \mathbb{E}_{z_0, \epsilon, t}
    \left[
    \left\|
    \epsilon - f_{DDT}(z_t, y, t)
    \right\|_2^2
    \right].
\end{equation}
This objective enables progressive reconstruction of anatomically plausible OCT image-mask latent pairs throughout the reverse diffusion process.

\textbf{Joint OCT and mask synthesis:} During inference, the reverse diffusion process is initialised by sampling $z_T \sim \mathcal{N}(0, I)$ and iteratively denoising according to:
\begin{equation}
    z_{t-1} = s_t(z_t,\, \hat{\epsilon},\, t), \quad \hat{\epsilon} = f_{DDT}(z_t, y, t),
    \quad \text{for } t = T-1, \dots, 0,
\end{equation}
where $s_t(\cdot)$ denotes the scheduler step. After the reverse diffusion process completes, the joint latent $\hat{z_0}$ of dimensions $\mathbb{R}^{8 \times H/8 \times W/8}$ is split along the channel dimension into an image $\hat{z_x}$ and a mask component $\hat{z_m}$, each of dimensions $\mathbb{R}^{4 \times H/8 \times W/8}$. Both components are independently decoded through the pretrained VAE decoder $f_D(.)$ to reconstruct the synthetic OCT image and its corresponding segmentation mask:
\begin{equation}
    \hat{x} = f_D(\hat{z_x}) \in \mathbb{R}^{1 \times H \times W}, \qquad \hat{m} = f_D(\hat{z_m}) \in \mathbb{R}^{1 \times H \times W}.
\end{equation}

Since both outputs are generated from a shared latent diffusion trajectory, the synthesised image-mask pairs preserve spatial and anatomical coherence.

\section{Experimental settings}
\label{4_exp_settings}

\subsection{Dataset}
\label{41_Dataset}

The dataset used in this study comprises OCT B-scans of ex vivo mouse retinas acquired with a high-resolution OCT system (Thorlabs Ganymede Series, Thorlabs GmbH, Luebeck, Germany). Retina samples were prepared, and OCT data were acquired as part of a previous study conducted within a project approved by the local authorities \cite{barroso_durable_2023}. Retinas were collected from 7 male and 3 female mice aged 2 to 27.5 months (10.27 $\pm$ 9.09 months). Because several OCT volumes were derived from the same animal, sex and age are reported at the animal level, whereas volume counts in Table \ref{tab:table_data} reflect the total number of processed samples. Retina samples were prepared using two distinct embedding protocols, which serve as the primary classes for model conditioning. 

The first group included 10 volumes of retina maintained in a physiological liquid medium, specifically water or agarose gel, which preserves tissue hydration close to native physiological conditions. However, in a previous study, Agarose-embedded samples showed higher background noise/scattering and degraded after several weeks, whereas resin-embedded samples remained stable for several years \cite{barroso_durable_2024}. 

In contrast, the second group consisted of 12 volumes from retina embedded in resin. Resin embedding produces highly stable, durable samples, making it suitable for long-term preservation, repeated imaging, and phantom-based OCT evaluation. Barroso et al. \cite{barroso_durable_2024} showed that resin-embedded murine retina can preserve layered retinal structures comparable to those observed in gel-based preparations, while producing lower background scattering/noise than agarose-based media. However, resin preparation requires fixation, dehydration, and polymerisation steps, which are irreversible and may introduce subtle shrinkage or preparation-related artefacts.

The differences between the datasets define two complementary OCT imaging domains, with considerable variation in the visibility and delineation of retinal layers. This domain distinction motivates the conditional generation of both embedding types rather than modelling them as a single homogeneous distribution.
 
In addition to the embedding protocol, the dataset included both control retinas from untreated mice and corresponding retinas treated with N-Methyl-D-aspartate (NMDA) to simulate glaucoma-induced retina degradation. This treatment status was not used to define the conditioning domains, but it increases the morphological variability of the dataset by introducing changes in retinal layer thickness and structure.


\begin{table}[h]
\centering
\caption{Distribution of the number of animals, retinal OCT volumes, and manually segmented B-scans across the two experimental groups.}
\label{tab:table_data}
\resizebox{\columnwidth}{!}{
    \begin{tabular}{ccccc}
        \hline
        \textbf{Dataset} & \textbf{N animals} & \textbf{OCT volumes} & \textbf{Segmented B-Scans} \\ \hline
        \rule{0pt}{10pt}
        Resin & 5 & 12 &  137 \\
        Physiological & 5 & 10 &  203 \\
        \hline
        \textbf{Total} & \textbf{10} & \textbf{22} &  \textbf{340} \\
        \hline
    \end{tabular}
}
\end{table}

\begin{figure*}[H]
    \centering
    \begin{minipage}{0.45\textwidth}
        \centering
        \textbf{Physiological embedding} \\
        \vspace{0.5cm}
        \begin{minipage}{0.45\textwidth}
            \centering
            \includegraphics[width=1\linewidth]{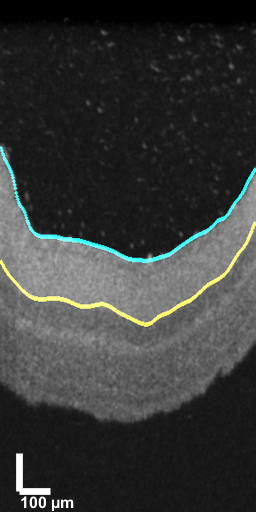} \\
            \vspace{0.3cm}
        \end{minipage}
        \hspace{0.3cm}
        \begin{minipage}{0.45\textwidth}
            \centering
            \includegraphics[width=1\linewidth]{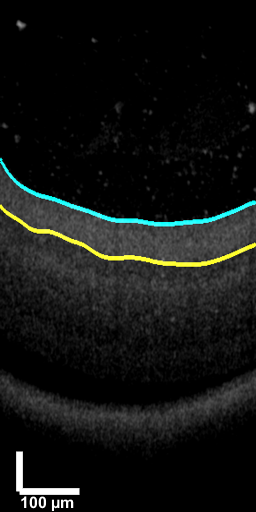} \\
            \vspace{0.3cm}
        \end{minipage}
    \end{minipage}
    \hspace{0.02\textwidth}
    \vrule width 0.5pt height 4cm
    \hspace{0.02\textwidth}
    \begin{minipage}{0.45\textwidth}
        \centering
        \textbf{Resin embedding} \\
        \vspace{0.5cm}
        
        \begin{minipage}{0.45\textwidth}
            \centering
            \includegraphics[width=1\linewidth]{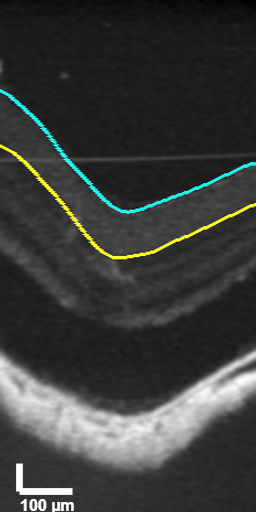} \\
            \vspace{0.3cm}
        \end{minipage}
        \hspace{0.3cm}
        \begin{minipage}{0.45\textwidth}
            \centering
            \includegraphics[width=1\linewidth]{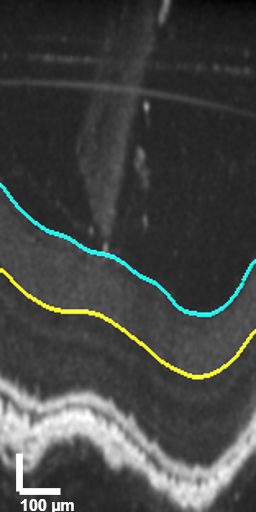} \\
            \vspace{0.3cm}
        \end{minipage}
    \end{minipage}
    \caption{Examples of URCL layer segmentation for representative B-scans of fluid and resin embedded samples. The images show the upper and lower boundaries of the URCL layer across the two experimental groups (see Table \ref{tab:table_data}).}
    \label{fig:URCL_delineation}
\end{figure*}

To generate the set of ground-truth masks, an expert manually segmented the upper retinal cell layer (URCL) in 203 B-scans from 10 physiologically-embedded volumes and 137 B-scans from 12 resin-embedded volumes, ensuring a highly accurate training database. Figure~\ref{fig:URCL_delineation} shows two representative examples of retinas embedded in the two different embedding media, along with the delineation of the URCL. Resin-embedded images exhibit sharper layer boundaries and higher signal contrast, albeit with occasional horizontal line artefacts arising from the air-resin interface, whereas physiologically-embedded images display a more uniform but lower-contrast signal with distributed speckle noise, resulting in more ambiguous layer transitions.

To ensure an unbiased evaluation of the downstream evaluation protocol that will be described in section~\ref{sec:432_downstream}, we divided the manually annotated B-scans into separate training, validation, and test sets, as summarised in Table~\ref{tab:table_partitions}. To prevent optimistic bias due to inter-slice correlations, splitting was performed at the volume level. For training generative models, we used only the training subset (236 B-scans: 95 from resin embedding and 141 from physiological embedding. This choice is critical because synthetic augmentation is evaluated through a downstream segmentation benchmark whose validation and test splits must remain strictly unseen during any generative training stage to maintain the independence of the augmentation assessment. Otherwise, the generative model could leak information from the held-out set into the segmentation model, compromising the fairness of the comparison. 

The validation set is used for the segmentation model selection and hyperparameter tuning, while the test set is kept separate for final performance evaluation. It is important to note that the downstream segmentation benchmark is evaluated only on the validation and test real images.

\begin{table}[h!]
\centering
\caption{Distribution of volumes, segmented B-scans across training, validation, and test partitions, categorised by embedding medium.}
\label{tab:table_partitions}
\begin{tabular}{cccc}
    \hline
    \textbf{Subset} & \textbf{Medium} & \textbf{Volumes} & \textbf{Segmented B-scans} \\
    \hline
    \multirow{3}{*}{\textbf{Training}} 
    & Resin & 9 & 95 \\ 
    & Physiological & 7 & 141 \\ 
    & \textbf{All} & \textbf{16}  & \textbf{236} \\
    \hline
    \multirow{3}{*}{\textbf{Validation}}
    & Resin & 1 & 23 \\ 
    & Physiological & 1 & 20 \\ 
    & \textbf{All} & \textbf{2} & \textbf{43} \\
    \hline
    \multirow{3}{*}{\textbf{Test}}
    & Resin & 2 & 19 \\ 
    & Physiological & 2 & 42 \\ 
    & \textbf{All} & \textbf{4} & \textbf{61} \\
    \hline
\end{tabular}
\end{table}

All images and masks were resized to $512\times256$ pixels to accommodate the variability in acquisition dimensions and because manual URCL delineations do not span the full retinal width. OCT B-scans are greyscale, and masks are binary (0: background, 1: URCL). 

\subsection{Generative Model Training}

All frameworks were implemented in PyTorch and trained on an NVIDIA DGX A100 system. All experiments were conducted using PyTorch 2.5 and Python 3.10. The code is publicly available at \url{https://github.com/cvblab/DualDiT}. 

\textbf{Training inputs:} For diffusion training, inputs were scaled to $[-1,1]$ to match the scale of the Gaussian noise added during the forward process and to ensure zero-centred inputs, which improves training stability. The image and mask were then encoded into a compressed latent space using the pretrained \texttt{sd-vae-ft-ema} VAE from StabilityAI \cite{noauthor_stabilityaisd-vae-ft-ema_nodate}, originally trained on OpenImages and subsequently fine-tuned on LAION-Aesthetics and LAION-Humans, and the standard latent scaling factor provided by the VAE configuration was applied. The encoder compresses the image and mask inputs by a factor of 8, producing latent tensors of spatial resolution $64\times32$. For joint image-mask generation, each training input was represented as a channel-wise concatenation of the image and its corresponding mask, yielding in a 8-channel latent representation. Conditioning used two classes corresponding to the embedding medium (\textit{resin} vs \textit{physiological}). 
 \vspace{0.25cm}

\textbf{DualDiT backbone:} The model was instantiated as a DiT-XL/2 backbone, processing the concatenated latent tensor through a patch embedding layer (patch size $2$) enriched with fixed 2D sine-cosine positional encodings, followed by 28 transformer blocks with hidden size 1152 and 16 attention heads.

\textbf{Model hyper-parameters:}  Optimisation was performed with Adam \cite{kingma_adam_2017} for 5000 epochs, with a learning rate of $10^{-4}$ and with batch size 16. The diffusion process used $T=1000$ timesteps and a linear noise schedule ($\beta_{\mathrm{start}}=1.5\times10^{-4}$, $\beta_{\mathrm{end}}=1.95\times10^{-2}$). The model was trained with classifier-free guidance \cite{ho_classier-free_nodate} with an unconditional probability of $0.1$. An exponential moving average (EMA) of the diffusion parameters with decay 0.9999 was maintained during training and used for sampling.

To evaluate the efficacy of the proposed DualDiT, two state-of-the-art diffusion architectures, conditionals DDPM and LDM, were adapted as baselines, ensuring a fair comparison by configuring them for joint image-mask generation:

\noindent \textbf{Conditional DDPM (Pixel Space) \cite{ho_denoising_2020}:} In this approach, the OCT image and its corresponding segmentation mask were treated as distinct channels of a single input tensor. Following normalisation, both components were concatenated along the channel dimension, resulting in a joint representation $x \in \mathbb{R}^{2 \times H \times W}$. The U-Net denoiser was trained to predict the combined noise of this dual-channel input, forcing the model to learn the joint distribution of the retinal anatomy and its pixel-level annotation directly in pixel space. The DDPM baseline employed a \texttt{UNet2DModel} from the \texttt{diffusers} library with channel widths of [32, 64, 64, 128, 128], one ResNet layer per block, and attention at the two deepest resolutions. Class conditioning was implemented by concatenating a learned class embedding of size 4 with the noisy input. The optimal hyper-parameter combination was achieved by training the DDPM for 6000 epochs. All remaining hyperparameters (diffusion timesteps $T$, learning rate, noise schedule, batch size, and classifier-free guidance probability) are shared with DualDiT as described above.

\noindent\textbf{Conditional LDM (Latent Space) \cite{rombach_high-resolution_2022-1}:} This implementation utilises a pre-trained VAE \cite{noauthor_stabilityaisd-vae-ft-ema_nodate} to perform diffusion in a compressed space. The resulting latent representations, $z_{x}$ and $z_{m} \in \mathbb{R}^{4 \times H/8 \times W/8}$, were then concatenated to form an 8-channel latent tensor. The LDM uses a U-Net denoiser as DDPM to model cross-modal spatial dependencies while leveraging the computational efficiency of the latent space. The LDM baseline employed a custom U-Net with channel widths of [64, 128, 256, 512, 512], two layers per stage, a bottleneck of [512, 512], 8 attention heads, Group Normalisation, and SiLU activations. The optimal hyper-parameters are shared with DualDiT as described above and were achieved training during 9500 epochs.

\subsection{Evaluation protocols}

\subsubsection{Generative quality metrics}
We evaluated the quality and diversity of the generated images using several standard metrics for generative models. 
First, we computed the Fréchet Inception Distance (FID) \cite{heusel_gans_2017}, which measures the distance between the feature distributions of real and generated images extracted from a pretrained Inception network. 
We represent the feature distributions of synthetic and real patches as 
$\mathcal{N}(\mu_{synth}, \Sigma_{synth})$ and $\mathcal{N}(\mu_{real}, \Sigma_{real})$, respectively.
The FID expression is given by:
\begin{equation}
\mathrm{FID} = \lVert \mu_{synth} - \mu_{real} \rVert^2 
+ \mathrm{Tr}\left( \Sigma_{synth} + \Sigma_{real} - 2 (\Sigma_{synth} \Sigma_{real})^{\frac{1}{2}} \right)
\end{equation}

Note that $\mathrm{FID} \in [0, +\infty)$, where lower values indicate higher similarity between real and generated distributions.

In addition, we report on the spatial FID (sFID), with $\mathrm{sFID} \in [0, +\infty)$, which evaluates the similarity of spatial features rather than grouped activations, providing a more sensitive assessment of structural consistency in the generated images.

All metrics were computed using the evaluation pipeline provided in the public implementation of Guided Diffusion models by OpenAI \cite{dhariwal_diffusion_2021}. The metrics were computed both globally and stratified by embedding medium to assess class-conditional fidelity.

\subsubsection{Downstream segmentation model training}
\label{sec:432_downstream}

To evaluate the clinical utility of the synthesized image-mask pairs, we perform a downstream segmentation task. We employ a standard U-Net architecture \cite{ronneberger_u-net_2015} as a baseline segmentation model, initially trained solely on the real training set ($\mathcal{D}$). We then evaluate the performance gain when the training set is augmented with synthetic image-mask pairs ($\mathcal{D'}$) generated by DDPM, LDM, and DualDiT. 

For real annotated images and masks, we used a standard combination of binary cross-entropy (BCE) and Dice loss~\cite{ma2021loss}, weighted equally ($\alpha = 0.5$):
\begin{equation}
    \mathcal{L}_{\mathrm{real}} = \alpha\,\mathcal{L}_{\mathrm{BCE}} + (1-\alpha)\,\mathcal{L}_{\mathrm{Dice}}
\end{equation}

When synthetic pairs were included, we adopted a confidence-guided teacher-student strategy to regulate their influence during training~\cite{9897435}. A teacher model trained exclusively on real data produces a pixel-wise confidence map $c_i = 2\left|p^{(T)}_i - 0.5\right|$, where $p^{(T)}_i = \sigma(f_T(x_s))_i$ denotes the teacher's predicted probability of a synthetic image ($x_s$) at pixel $i$. Thus, the confidence map down-weights uncertain synthetic regions in the BCE term while leaving the Dice loss unweighted, preserving a global structural penalty regardless of local mask quality. The total loss is $
    \mathcal{L}_{\mathrm{total}} = \mathcal{L}_{\mathrm{real}} + \mathcal{L}_{\mathrm{synth}}, $
where $\mathcal{L}_{\mathrm{synth}}$ combines confidence-weighted BCE and standard Dice over the synthetic branch.

\textbf{Model architecture}: The downstream segmentation model employed was a U-Net  fully convolutional network adapted for single-channel OCT B-scans and binary masks. The encoder consists of four convolutional blocks with 64, 128, 256, and 512 channels, followed by a bottleneck with 1024 channels. The decoder mirrors the encoder with upsampling and skip connections, and the output layer uses a 1×1 convolution to produce a single-channel probability map for URCL segmentation.
\vspace{0.25cm}

\textbf{Training protocol:} All models were trained for 300 epochs (batch size 16) using Adam ($\text{lr}=10^{-4}$, weight decay $10^{-4}$) with a \textit{ReduceLROnPlateau} scheduler (factor 0.5, patience 30 epochs) monitoring the validation Dice coefficient. These hyperparameters were fixed across the baseline and all augmentation strategies (DDPM, LDM, DualDiT) to ensure a fair comparison.

\subsubsection{External evaluation protocol}

As an additional evaluation method, a panel of experts with varying levels of expertise conducted a qualitative assessment of the synthesised images and masks. To this end, a visual evaluation test was prepared using image-mask pairs, in which the experts were required to distinguish real pairs from synthetic ones and to classify each pair according to the embedding medium (resin or physiological). In total, 100 images (50 real and 50 synthetic) were analysed. The two embedding classes were equally represented, yielding 25 images per class and per sample type (real or synthetic). 

\begin{figure}[]
    \centering
    \includegraphics[width=\linewidth]{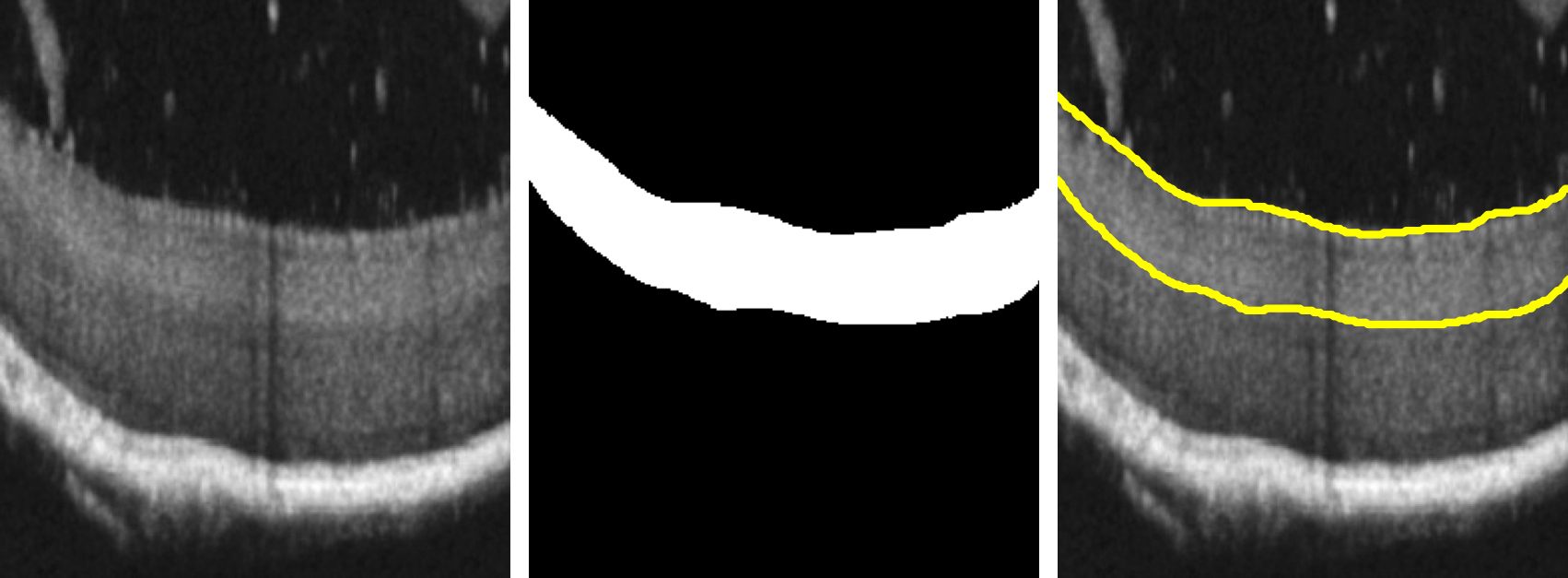}
    \caption{Example figure of a synthetic resin-embedded sample for expert evaluation: (Left) OCT B-Scan; (Middle) URCL segmentation mask; (Right) URCL delineation on the B-Scan.}
    \label{fig:expert_evaluation}
\end{figure}

To ensure the samples were assessed correctly, in each case, the B-scan, the URCL segmentation mask and the outline of the mask’s edges on the B-scan were displayed. As shown in Figure~\ref{fig:expert_evaluation}, this representation enabled the experts to evaluate not only the quality of the generated B-scans but also the correspondence between the B-scans and their corresponding segmentations.

\section{Results}
\label{6_results}

\subsection{Generative Quality Comparison}

\textbf{Quantitative results}: Table~\ref{tab:fid} presents the results for FID and sFID for the diffusion models examined in this study: DDPM\cite{ho_denoising_2020}, LDM \cite{rombach_high-resolution_2022-1}, and the proposed DualDiT. The \textit{Global} column of the table shows the results for the whole real data distribution. In contrast, the \textit{Resin} and \textit{Physiological} columns stratify the results for each embedding medium. Our framework achieves significant performance improvements over state-of-the-art models, with FID and sFID scores of $56.14$ and $114.35$, respectively. When analysing the results by embedding medium, we find that DualDiT outperforms the other frameworks across both metrics.

\begin{table*}[h]
    \centering
    \caption{Quantitative comparison of generative models across datasets.}
    \label{tab:fid}
    \small
    \resizebox{\textwidth}{!}{
    \begin{tabular}{|ll|ccc|ccc|ccc|}
        \hline
        & & \multicolumn{3}{c|}{Physiological} & \multicolumn{3}{c|}{Resin} & \multicolumn{3}{c|}{Global} \\
        & & DDPM \cite{ho_denoising_2020} & LDM \cite{rombach_high-resolution_2022-1} & DualDiT & DDPM \cite{ho_denoising_2020} & LDM \cite{rombach_high-resolution_2022-1} & DualDiT & DDPM \cite{ho_denoising_2020} & LDM \cite{rombach_high-resolution_2022-1} & DualDiT \\
        \hline
        \multirow{1}{*}{FID $\downarrow$} & & 175.02 & 124.79 & \textbf{66.96} & 208.99 & 109.60 & \textbf{59.83} & 164.55 & 102.21 & \textbf{56.14} \\
        sFID $\downarrow$ & & 278.87 & 157.93 & \textbf{122.15} & 329.91 & 153.70 & \textbf{114.72} & 254.52 & 150.66 & \textbf{114.35} \\
        \hline
    \end{tabular}
    }
\end{table*}

\begin{figure}[]
    \centering
    \setlength{\tabcolsep}{1pt}
    \renewcommand{\arraystretch}{1.0}
    \begin{tabular}{c cc!{\hspace{2mm}\vrule\hspace{2mm}}cc}
        & \multicolumn{2}{c}{\textbf{Physiological}} &
        \multicolumn{2}{c}{\textbf{Resin}} \\[2mm]

        \raisebox{1.25cm}{\scriptsize Real} &
        \includegraphics[width=0.18\columnwidth]{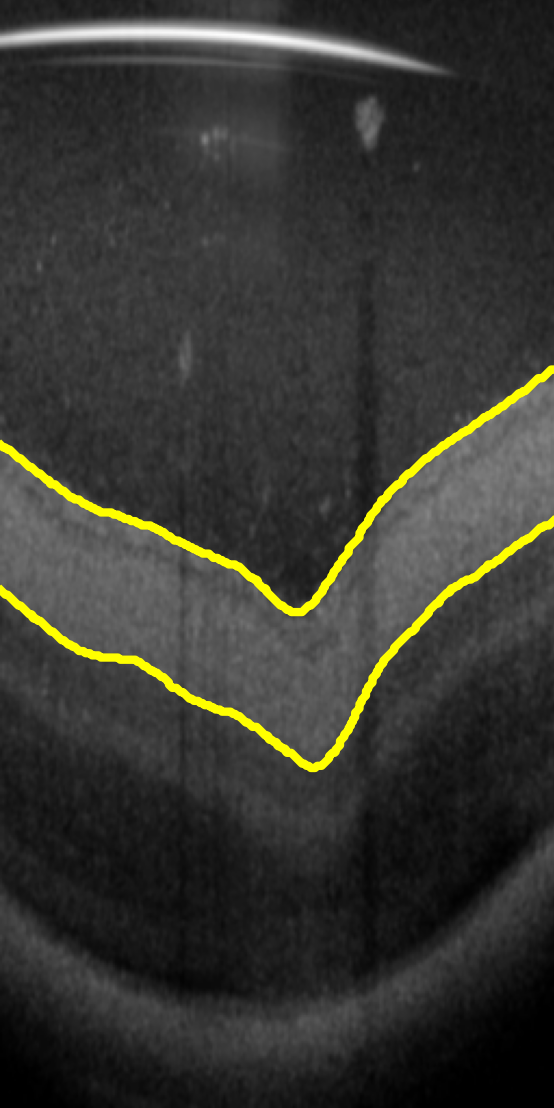} &
        \includegraphics[width=0.18\columnwidth]{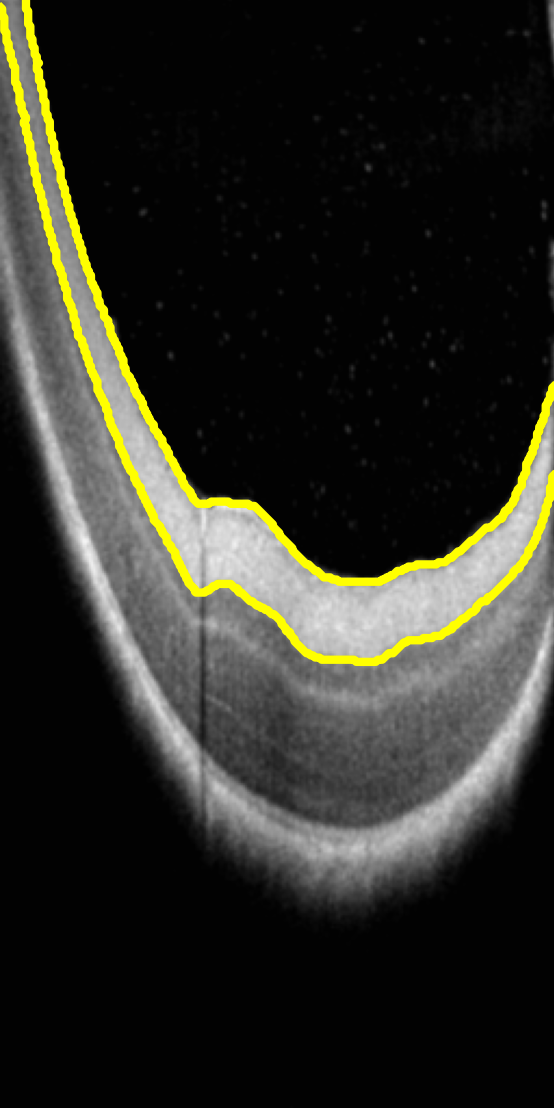} &
        \includegraphics[width=0.18\columnwidth]{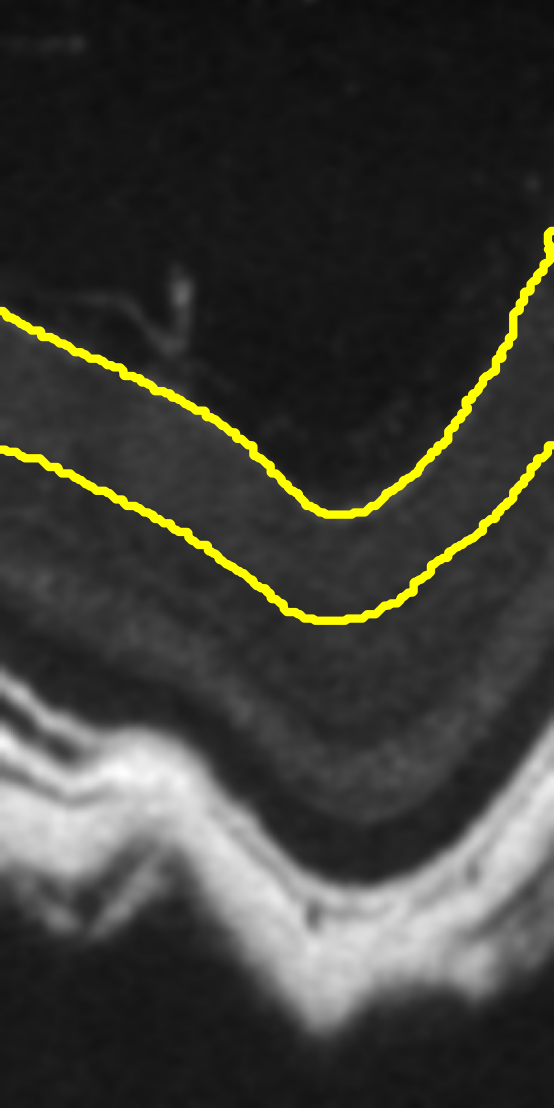} &
        \includegraphics[width=0.18\columnwidth]{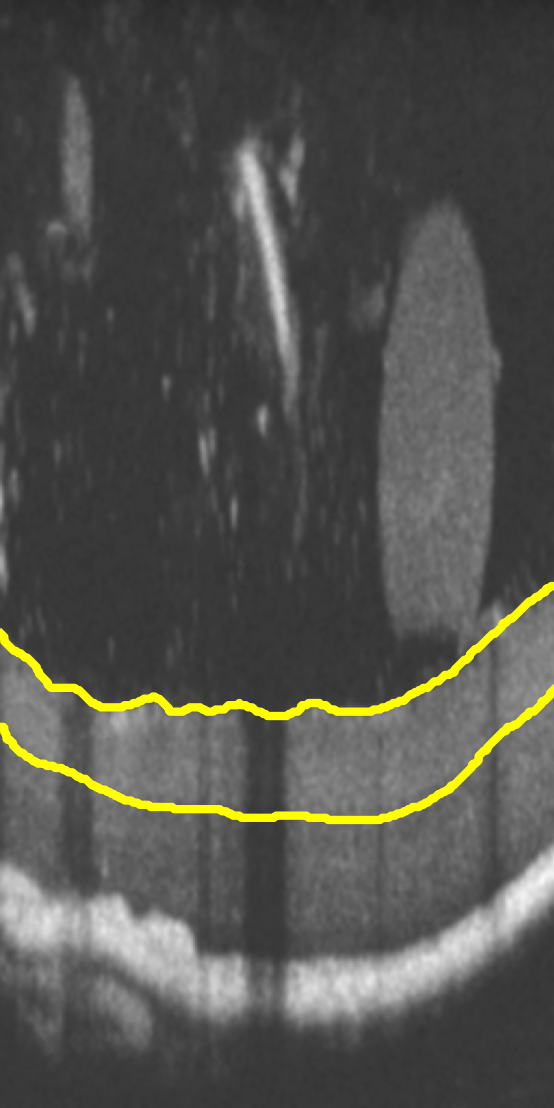} \\

        \raisebox{1.25cm}{\scriptsize DDPM \cite{ho_denoising_2020}} &
        \includegraphics[width=0.18\columnwidth]{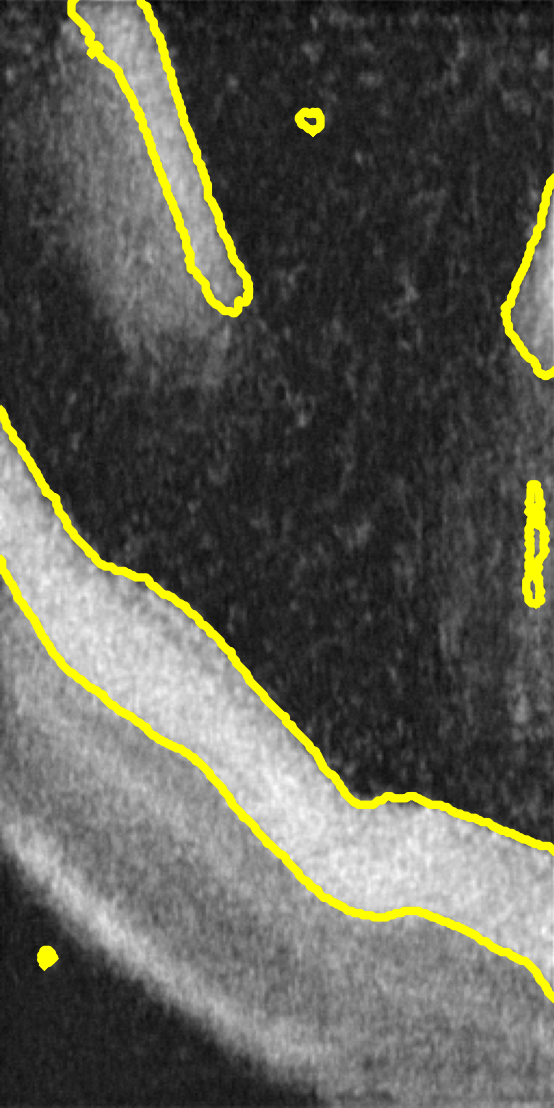} &
        \includegraphics[width=0.18\columnwidth]{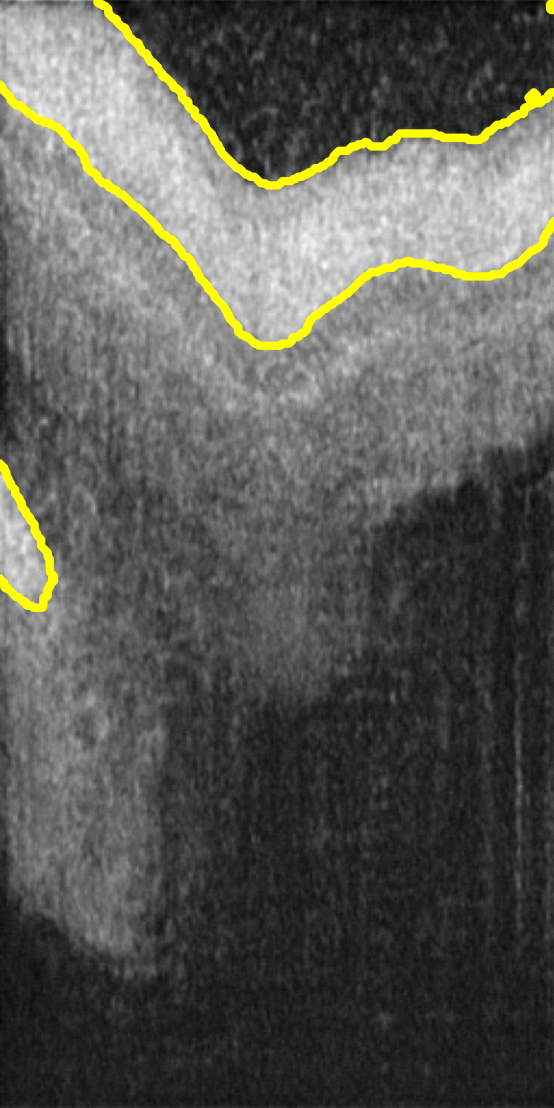} &
        \includegraphics[width=0.18\columnwidth]{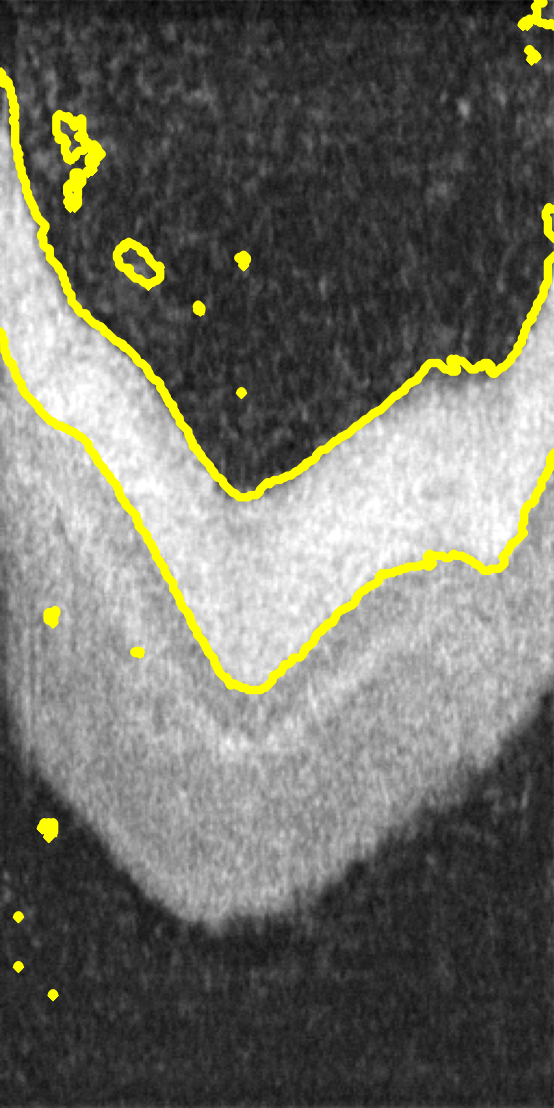} &
        \includegraphics[width=0.18\columnwidth]{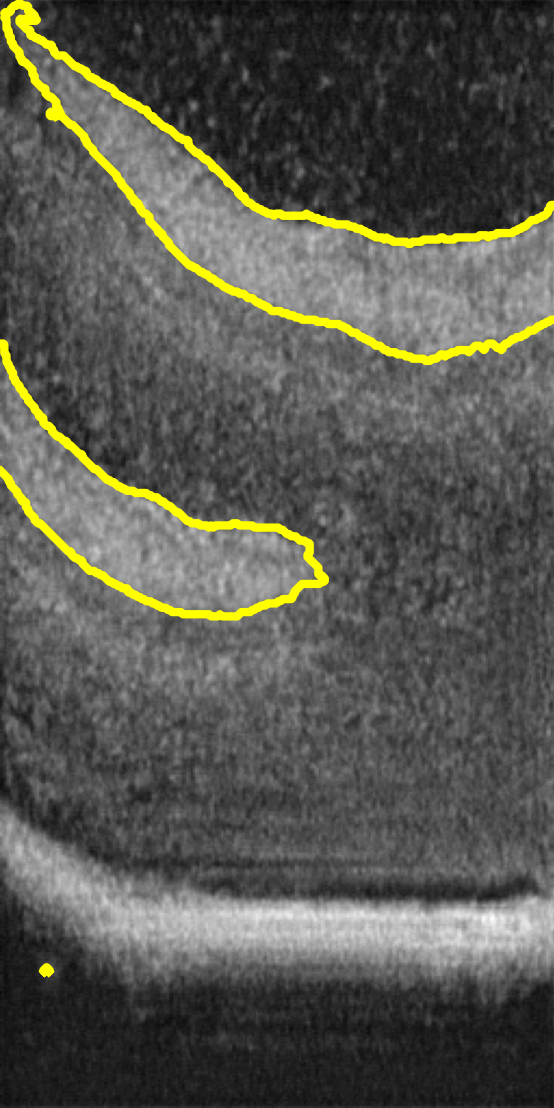} \\

        \raisebox{1.25cm}{\scriptsize LDM \cite{rombach_high-resolution_2022-1}} &
        \includegraphics[width=0.18\columnwidth]{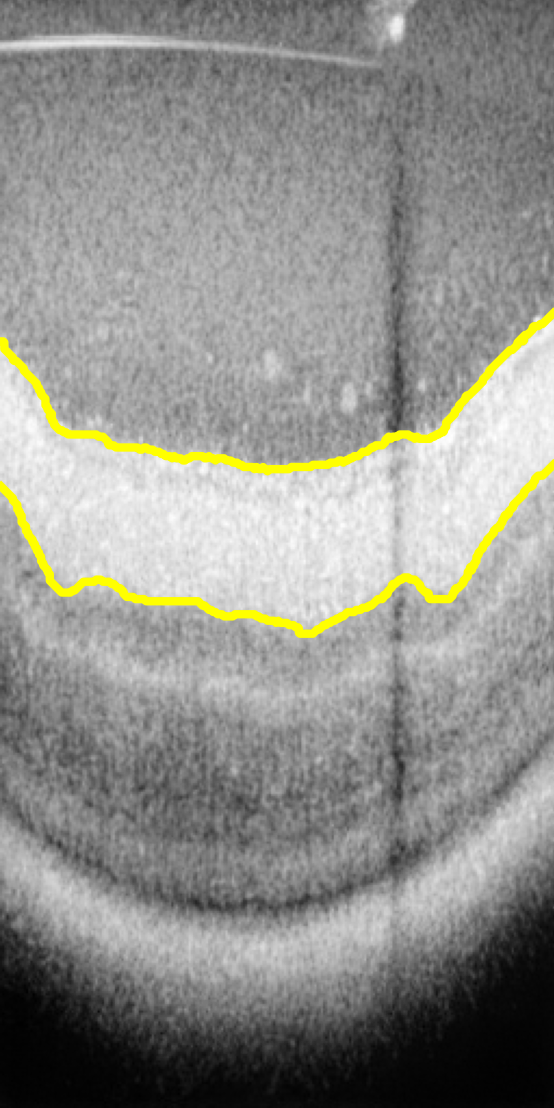} &
        \includegraphics[width=0.18\columnwidth]{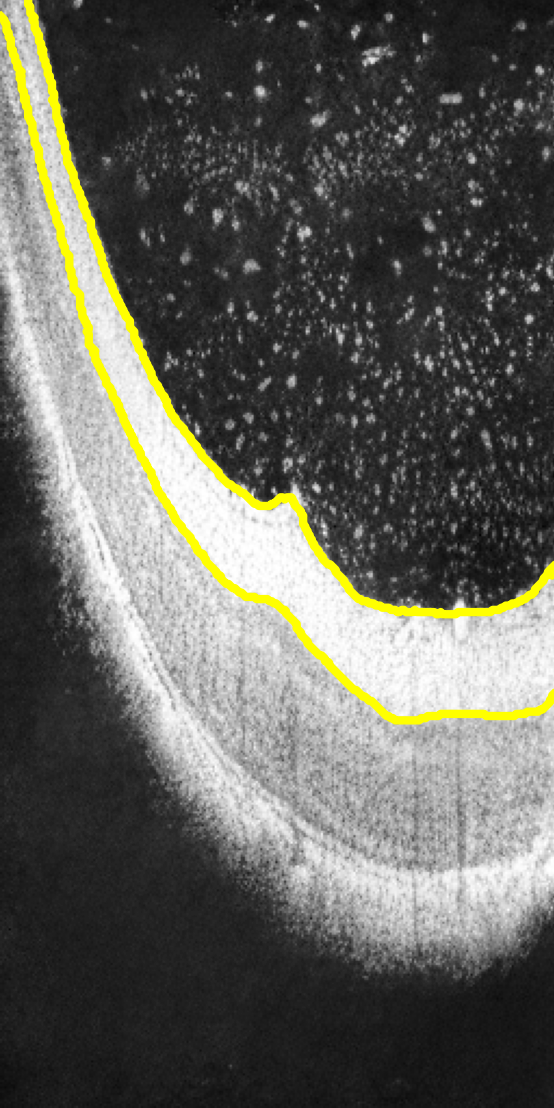} &
        \includegraphics[width=0.18\columnwidth]{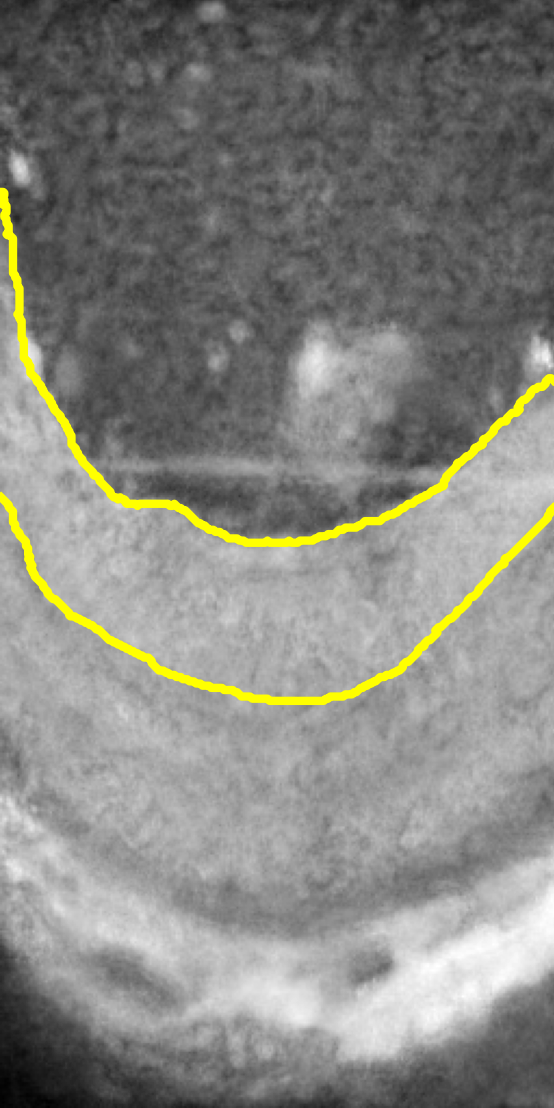} &
        \includegraphics[width=0.18\columnwidth]{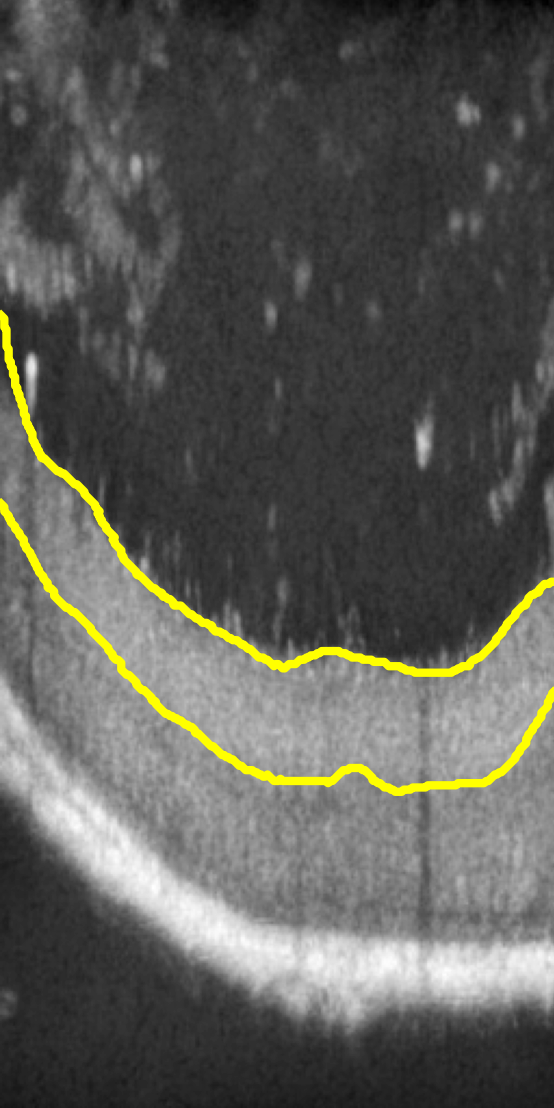} \\

        \raisebox{1.25cm}{\scriptsize
        \begin{tabular}{@{}c@{}}
            DualDiT\\
            (Ours)
        \end{tabular}} &
        \includegraphics[width=0.18\columnwidth]{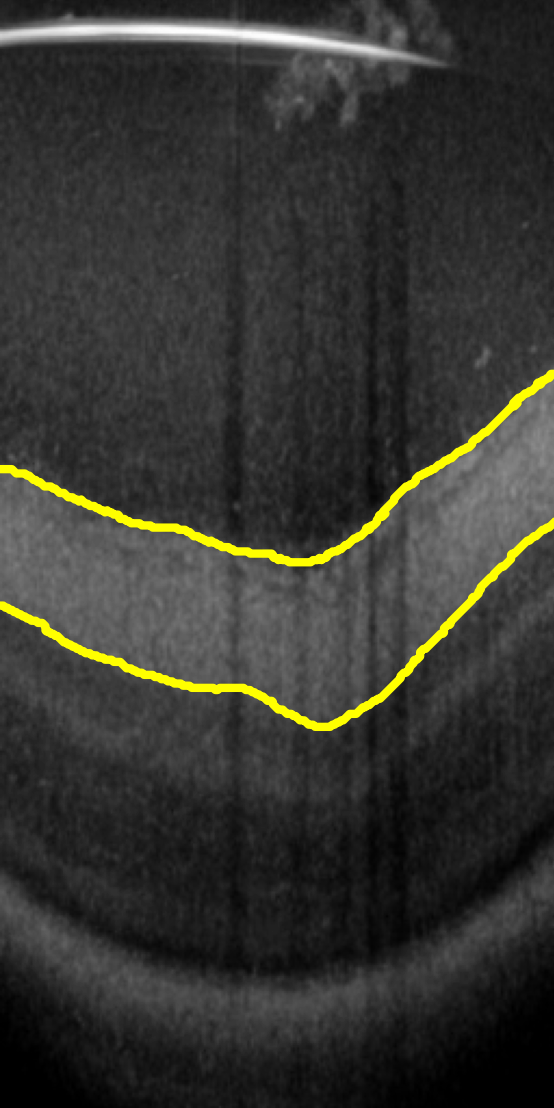} &
        \includegraphics[width=0.18\columnwidth]{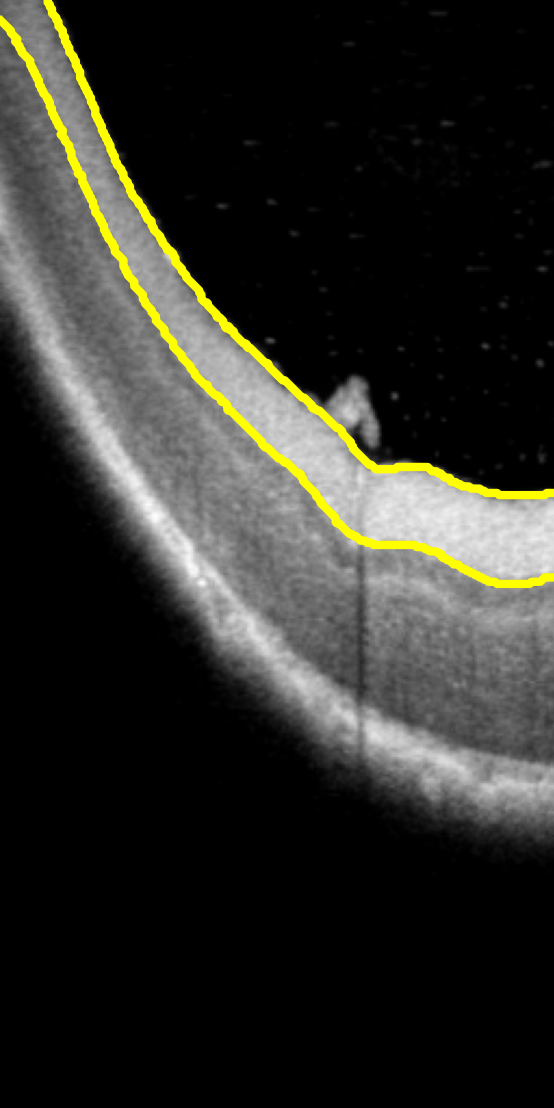} &
        \includegraphics[width=0.18\columnwidth]{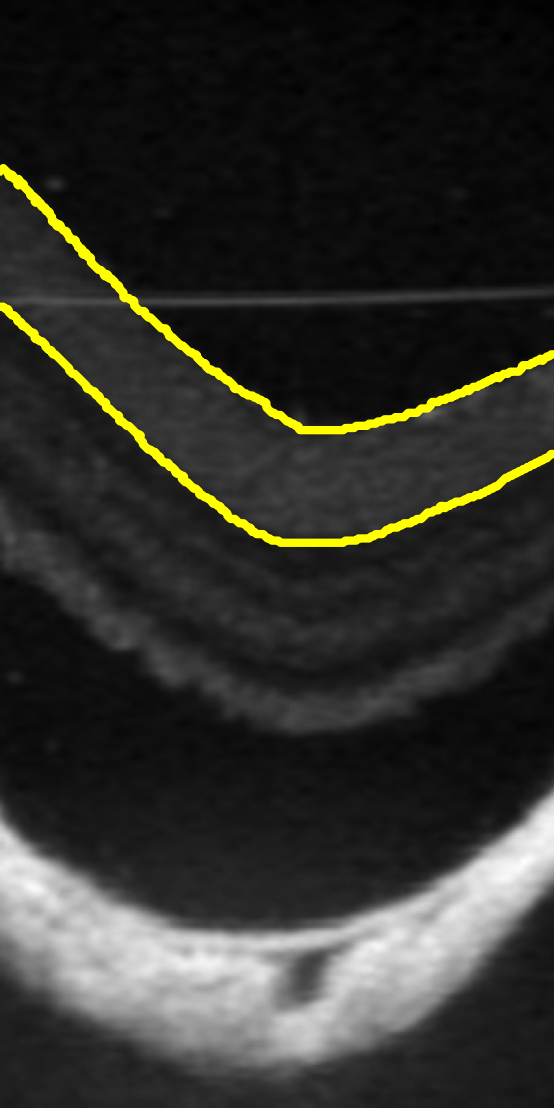} &
        \includegraphics[width=0.18\columnwidth]{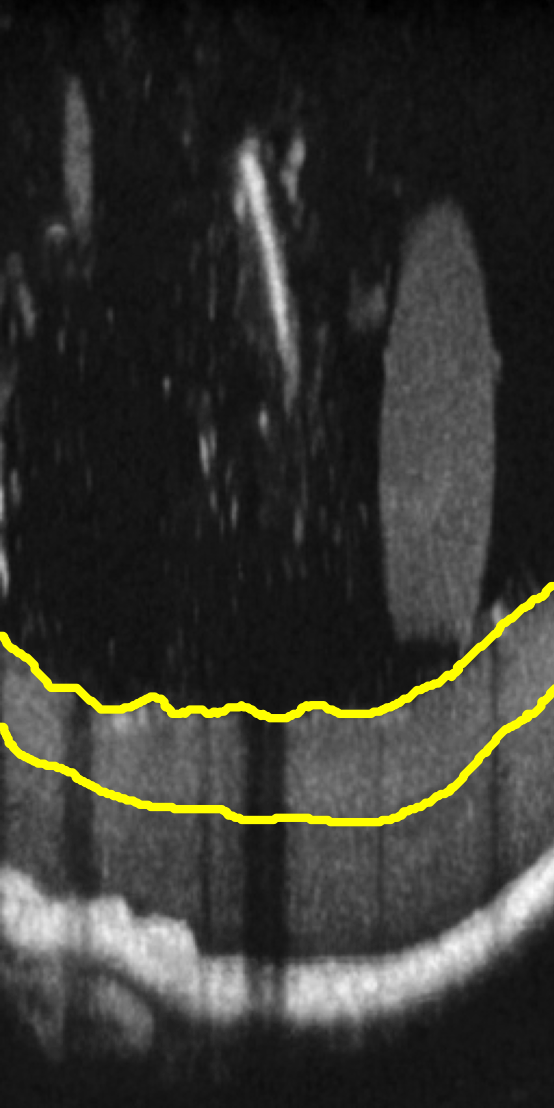} \\
    \end{tabular}
    \caption{Comparison of real and synthetic images with their corresponding URCL mask overlays. The left two columns show physiologically embedded samples, whereas the right two columns show resin-embedded samples.}
    \label{fig:qualitative_comparison}
\end{figure}

\textbf{Qualitative results and visual inspection:} To qualitatively evaluate the proposed method, we present synthesised images using the DDPM, LDM, and DualDiT frameworks. Figure~\ref{fig:qualitative_comparison} shows comparisons of real physiological and resin-embedded images with their corresponding mask overlays and the outputs of the different methods assessed: DDPM, LDM, and our proposed model (DualDiT). The DualDiT approach demonstrates greater resolution of the retinal layers and high reliability in reproducing common OCT artefacts and retinal defects observed in the actual data collected during experimental protocols and the embedding process.

For completeness, additional synthetic samples are provided in Appendix~\ref{app1} (Figures~\ref{fig:ddpm_0} and~\ref{fig:ddpm_1}), showing respectively, physiological and resin-embedded retinal images alongside their corresponding masks generated by the DDPM approach. This framework exhibits hallucinations in the shape of the retina for both resin and physiological embedding mediums. While it successfully captures the grey levels and distribution of the original data, it lacks the resolution required to differentiate between the URCL and the remaining retinal layers, and fails to reproduce the continuous structure of the retina and common OCT image artefacts such as speckle noise and refraction effects. Additionally, it often exaggerates intensity, resulting in overly bright white tones. With regard to mask synthesis, it can be observed how it segments artifacts outside the boundaries of the region of interest

Figures~\ref{fig:ldm_0} and~\ref{fig:ldm_1} show pairs of retinal images and masks generated by the LDM framework. This approach significantly improves upon the DDPM results, offering greater layer definition, higher overall resolution and contrast, more accurate replication of retinal shapes, and reduced output variability. However, there is still a perceived inconsistency in the resolution and detail of the retinal images. Similar to DDPM, LDM sometimes produces images that are overly saturated and bright, especially in physiological medium images (e.g., the second and fourth images in Figure~\ref{fig:ldm_1}).

Finally, Figures~\ref{fig:dit_0} and~\ref{fig:dit_1} display images and masks generated by the proposed DualDiT model. The synthetic images show a closer resemblance to real data for both resin and physiological embedding mediums. Qualitatively, the layers of the retina generated by DualDiT, particularly the URCL, appear to have higher resolution and contrast compared to the results from the DDPM and LDM models.

\subsection{Segmentation performance with synthetic augmentation}

This section presents the validation of the proposed method as a data augmentation strategy. To evaluate its effectiveness, we assess the segmentation model described in Section~\ref{sec:432_downstream} using the Dice Similarity Coefficient (DSC) and Intersection over Union (IoU) \cite{muller2022towards} on a strictly held-out real test set presented in Table~\ref{tab:table_partitions}. The reference model was trained on the training subset ($\mathcal{D}$), referred to as the \textit{Baseline} model. This model achieves a global Dice score of 0.908 $\pm$ 0.066 and an Intersection over Union (IoU) of 0.838 $\pm$ 0.106 on the test set. 

Table~\ref{tab:augmentation_results} summarises the segmentation performance obtained with the Baseline and the different synthetic-data augmentation strategies across the physiological, resin, and global test subsets. By augmenting the training dataset, as $\mathcal{D} \cup \mathcal{D'}$, with $\mathcal{D'}=50 $ synthetic images and masks, comprising 25 inputs from the resin class and 25 from the physiological class, generated by the DDPM, LDM, and DualDiT frameworks, we trained three additional segmentation models. As shown in Table~\ref{tab:augmentation_results}, among the three generative frameworks, DualDiT provides the most consistent results and is the only method to improve both global metrics relative to the Baseline. The model trained with DualDiT-generated samples achieves a global Dice score of 0.917 $\pm$ 0.058 and an IoU of 0.852 $\pm$ 0.095, compared with 0.908 $\pm$ 0.066 and 0.838 $\pm$ 0.106 for the Baseline. This improvement is consistent across embedding media, with DualDiT achieving Dice and IoU scores comparable to or higher than those of the Baseline in both the physiological and resin subsets. In contrast, DDPM substantially degrades performance in the resin subset, while LDM yields intermediate results but does not consistently surpass the Baseline. This degradation is consistent with the generative quality metrics reported in Table~\ref{tab:fid} as DDPM is the only model for which FID and sFID are markedly worse for the resin subset than for the physiological subset (+33.97 and +51.04 points, respectively), whereas both LDM and DualDiT achieve better resin than physiological scores despite the resin subset containing fewer training volumes (Table~\ref{tab:table_partitions}). This asymmetry suggests that the limited number of resin training samples is not, by itself, an insurmountable barrier to synthesis, but rather interacts with DDPM's more limited modelling capacity, which appears to hinder its ability to generalise from a comparatively smaller training set.

\begin{table*}[ht]
\caption{Comparison of segmentation performance between the baseline model and models trained with different numbers of synthetic image-mask pairs. Results are reported as mean and standard deviation for Dice and IoU. $\mathcal{D}$ denotes the original training set composed of 236 image-mask pairs.}
\centering
\label{tab:augmentation_results}
\resizebox{\textwidth}{!}{
\begin{tabular}{|c|c|cccccc|cccccc|}
\hline
                 & & \multicolumn{6}{c|}{Dice} & \multicolumn{6}{c|}{IoU} \\ \cline{3-14}
                 & & \multicolumn{2}{c|}{Physiological} & \multicolumn{2}{c|}{Resin} & \multicolumn{2}{c|}{Global}
                 & \multicolumn{2}{c|}{Physiological} & \multicolumn{2}{c|}{Resin} & \multicolumn{2}{c|}{Global} \\ \cline{3-14}
Model & Training images & \multicolumn{1}{c|}{Mean} & \multicolumn{1}{c|}{SD}
                 & \multicolumn{1}{c|}{Mean} & \multicolumn{1}{c|}{SD}
                 & \multicolumn{1}{c|}{Mean} & \multicolumn{1}{c|}{SD}
                 & \multicolumn{1}{c|}{Mean} & \multicolumn{1}{c|}{SD}
                 & \multicolumn{1}{c|}{Mean} & \multicolumn{1}{c|}{SD}
                 & \multicolumn{1}{c|}{Mean} & \multicolumn{1}{c|}{SD} \\ \hline

Baseline & $\mathcal{D}$ & 0.912 & 0.051 & 0.906 & 0.072 & 0.908 & 0.066 & 0.843 & 0.079 & 0.836 & 0.116 & 0.838 & 0.106 \\
\hline
DDPM \cite{ho_denoising_2020} & $\mathcal{D}+50$ & 0.917 & 0.073 & 0.528 & 0.312 & 0.796 & 0.258 & 0.855 & 0.116 & 0.415 & 0.266 & 0.718 & 0.270 \\
LDM \cite{rombach_high-resolution_2022-1} & $\mathcal{D}+50$ & 0.904 & 0.078 & 0.845 & 0.088 & 0.886 & 0.086 & 0.834 & 0.125 & 0.742 & 0.126 & 0.805 & 0.133 \\
DualDiT & $\mathcal{D}+50$ & 0.919 & 0.064 & 0.913 & 0.041 & 0.917 & 0.058 & 0.856 & 0.105 & 0.842 & 0.067 & 0.852 & 0.095 \\
\hline
DDPM \cite{ho_denoising_2020} & $\mathcal{D}+100$ & 0.921 & 0.070 & 0.603 & 0.199 & 0.822 & 0.193 & 0.860 & 0.112 & 0.459 & 0.199 & 0.735 & 0.235 \\
LDM \cite{rombach_high-resolution_2022-1} & $\mathcal{D}+100$ & 0.907 & 0.079 & 0.897 & 0.055 & 0.903 & 0.072 & 0.838 & 0.126 & 0.817 & 0.084 & 0.832 & 0.115 \\
DualDiT & $\mathcal{D}+100$ & 0.914 & 0.069 & 0.915 & 0.036 & 0.914 & 0.060 & 0.849 & 0.112 & 0.846 & 0.061 & 0.848 & 0.099 \\
\hline
DualDiT & $\mathcal{D}+200$ & 0.921 & 0.063 & 0.926 & 0.025 & 0.923 & 0.054 & 0.860 & 0.104 & 0.864 & 0.042 & 0.861 & 0.089 \\
DualDiT & $\mathcal{D}+400$ & 0.921 & 0.063 & 0.925 & 0.027 & 0.922 & 0.054 & 0.859 & 0.103 & 0.862 & 0.046 & 0.860 & 0.090 \\
DualDiT & $\mathcal{D}+600$ & 0.925 & 0.061 & 0.919 & 0.037 & 0.923 & 0.055 & 0.866 & 0.100 & 0.853 & 0.062 & 0.862 & 0.090 \\
DualDiT & $\mathcal{D}+800$ & 0.920 & 0.062 & 0.929 & 0.023 & 0.923 & 0.053 & 0.858 & 0.102 & 0.868 & 0.039 & 0.861 & 0.088 \\
DualDiT & $\mathcal{D}+1000$ & 0.927 & 0.054 & 0.919 & 0.033 & 0.925 & 0.049 & 0.869 & 0.090 & 0.852 & 0.056 & 0.864 & 0.082 \\
\textbf{DualDiT} & \textbf{$\mathcal{D}+1200$} & \textbf{0.927} & \textbf{0.055} & \textbf{0.927} & \textbf{0.027} & \textbf{0.927} & \textbf{0.048} & \textbf{0.869} & \textbf{0.092} & \textbf{0.866} & \textbf{0.046} & \textbf{0.868} & \textbf{0.081} \\
DualDiT & $\mathcal{D}+1400$ & 0.926 & 0.057 & 0.917 & 0.034 & 0.923 & 0.051 & 0.867 & 0.094 & 0.848 & 0.056 & 0.861 & 0.085 \\
\hline
\end{tabular}
}
\end{table*}

We also evaluated augmentation with 100 synthetic image-mask pairs, evenly split between the two classes. At this augmentation level, DualDiT again provides the strongest overall results among the three generative approaches, reaching a global Dice score of 0.914 $\pm$ 0.060 and an IoU of 0.848 $\pm$ 0.099. However, these values are only slightly above the Baseline and are lower than those obtained with larger DualDiT augmentation sets. DDPM continues to exhibit a marked degradation in the resin subset,  whereas LDM improves substantially over DDPM but does not provide a consistent global gain over the Baseline. These results suggest that synthetic image-mask pairs generated by DualDiT can provide a consistent benefit when augmenting a limited OCT segmentation dataset, although the magnitude of the improvement depends on the number of synthetic samples added.

Overall, DualDiT is the only augmentation framework that consistently improves the Baseline across multiple augmentation sizes. DDPM performs poorly in the resin subset and exhibits substantially larger variability, which may indicate limited generalisation or a domain mismatch in the generated samples. LDM produces more competitive results than DDPM, particularly for resin-embedded images, but its global performance remains close to or below the Baseline. By contrast, DualDiT generally improves the resin results while maintaining comparable performance in the physiological subset. 

Given the comparatively consistent behaviour of DualDiT, we extended the augmentation study from $\mathcal{D}+200$ up to $\mathcal{D}+1400$ synthetic image-mask pairs, following the progressive data-expansion strategy adopted in previous synthetic augmentation studies \cite{toker_satsynth_2024}. DualDiT yields global Dice and IoU values above the Baseline for nearly all evaluated augmentation sizes, although the improvement is not monotonic. The best overall performance is obtained with $\mathcal{D}+1200$, reaching a global Dice score of 0.927 $\pm$ 0.048 and an IoU of 0.868 $\pm$ 0.081. The gains are particularly evident in the resin subset, while the differences in the physiological subset are smaller, partly because the Baseline performance is already comparatively high. Similar improvements are observed for $\mathcal{D}+600$ and $\mathcal{D}+1000$, indicating that the benefit is not restricted to a single augmentation size in this particular dataset. This suggests that augmentation quality and diversity are more relevant than the absolute number of generated samples. Since the synthetic pairs are derived from the available training distribution, increasing their number may eventually introduce redundancy rather than additional variability. Nevertheless, this interpretation remains hypothetical and would require further experiments with larger datasets and independent test volumes. 

At the global level, the $\mathcal{D}+1200$ configuration increases the mean Dice and IoU scores by 2.09\% and 3.58\% relative to the Baseline, respectively, indicating a modest but consistent overall benefit across the complete test set. To complement these results, a per-B-scan analysis was conducted based on relative changes in Dice and IoU relative to the Baseline. Figure~\ref{fig:appendix_dualdit_results} presents the full per-B-scan comparison for the $\mathcal{D}+1200$ configuration. The Dice and IoU scatter plots in Figures~\ref{fig:scatter_dice} and~\ref{fig:scatter_iou}, respectively, show that most B-scans lie above the identity line, with the largest gains occurring mainly in cases where the Baseline performs poorly. Figure~\ref{fig:relative_changes} shows the relative percentage changes in Dice and IoU for all test B-scans, ordered by increasing IoU change. The results highlight that DualDiT outperforms the Baseline in the majority of cases, although performance decreases are observed for a small subset of B-scans.
    

\subsection{External validation protocol}

\begin{figure*}[]
    \centering
    \includegraphics[width=1\linewidth]{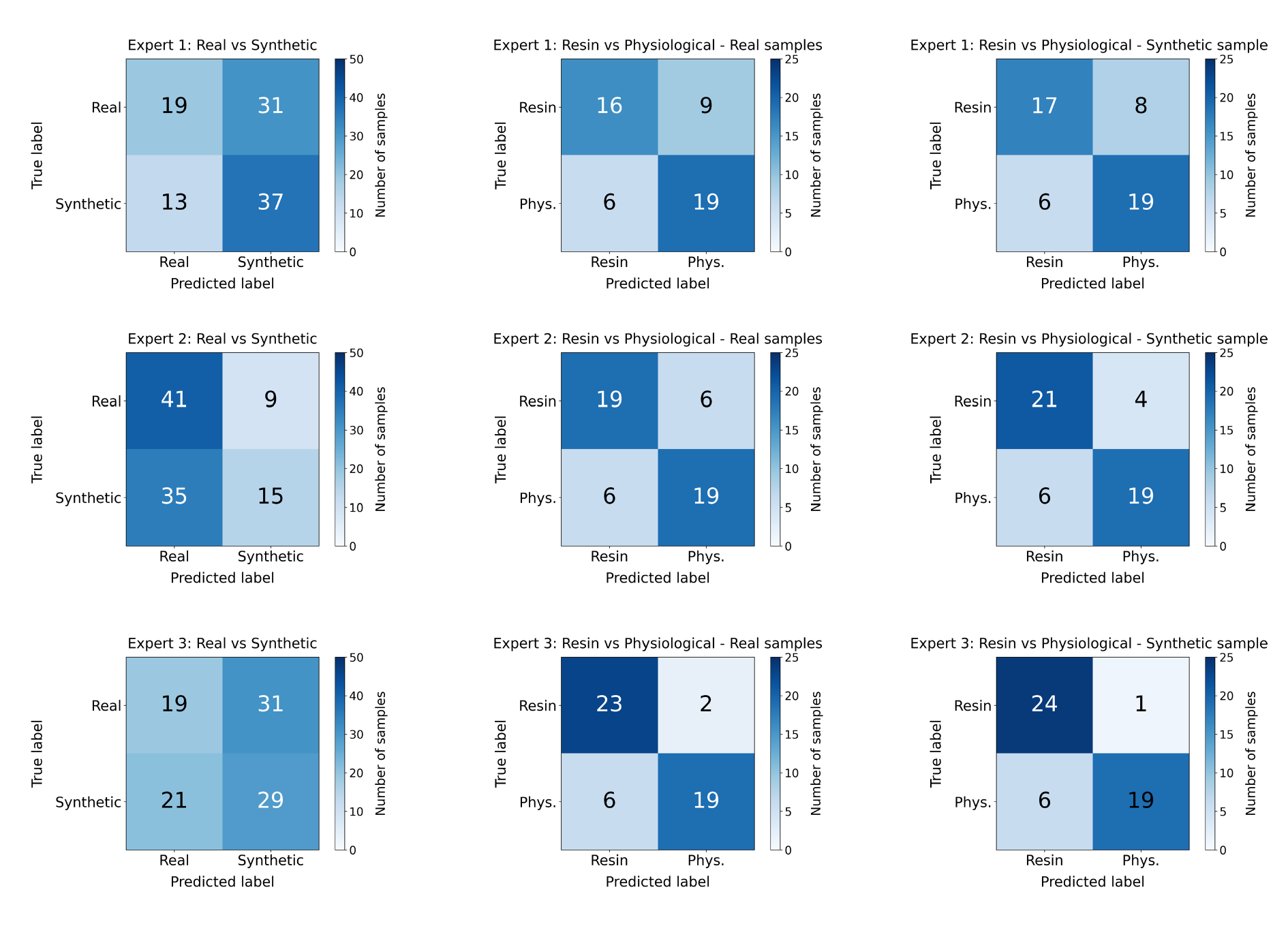}
    \caption{Confusion matrix for the classification performed by experts. (Left) Classification for real and synthetic samples; (Middle) Embedding medium classification for real samples; (Right) Embedding medium classification for synthetic pairs.}
    \label{fig:matrices}
\end{figure*}

To demonstrate the usefulness and realism of the proposed model's outputs, a panel of three experts evaluated the quality of the synthetic images. Figure~\ref{fig:matrices} shows the results obtained by every expert. Experts incorrectly classified, on average, 46\% of synthetic samples as real. On the other hand, 42\% of the real samples were deemed synthetic.

As regards the classification of embedding media, both resin and liquid embeddings were correctly identified in both real and synthetic samples, with similar success rates (76.7\% and 79.3\%, respectively). Additionally, we tested whether there was a statistically significant difference in classification performance between synthetic and real samples. For this purpose, the area under the receiver operating characteristic (ROC) curve (AUC) was calculated, a metric ranging from 0.5 (random performance) to 1 (perfect classification). Table~\ref{tab:auc_resin_fluid_experts} shows the AUC metric for identifying real (R) and synthetic (S) samples for each expert. In addition, the p-value (with $\alpha = 0.05$) is provided. Table~\ref{tab:auc_resin_fluid_experts} shows no statistically significant differences ($p-value > 0.5$) between the identification of media based on real or synthetic samples. This demonstrates that the developed model successfully distinguishes and represents both embedding materials.

\begin{table}[h]
\centering
\caption{AUC metrics for resin vs physiological classification on real and synthetic images, including the p-value for the comparison between AUC-R and AUC-S.}
\label{tab:auc_resin_fluid_experts}
\small
\vspace{0.3cm}
\begin{tabular}{lccc}
\hline
\textbf{Expert} & 
\textbf{AUC$_{\mathrm{real}}$} & 
\textbf{AUC$_{\mathrm{synth}}$} & 
\textbf{p-value} \\
\hline
Expert 1 & 0.700 & 0.720 & 0.852 \\
Expert 2 & 0.760 & 0.800 & 0.662 \\
Expert 3 & 0.840 & 0.860 & 0.801 \\
\hline
\end{tabular}
\end{table}

\section{Conclusions}
\label{7_conclusion}

In this study, we introduce a conditional dual-output Diffusion Transformer (DualDiT) that jointly synthesises OCT images and their corresponding segmentation masks in a single generative process. We evaluate its performance by generating realistic OCT B-scans of dissected ex vivo mouse retinas embedded in two distinct media, together with anatomically consistent URCL segmentation masks. The proposed model achieves an overall FID of 56.14, which is substantially lower than those of other tested diffusion models, such as DDPM and LDM, which achieved FIDs of 164.55 and 102.21, respectively. To validate the proposed method as a data augmentation strategy, we trained a URCL segmentation model and evaluated it on a separate dataset. The Baseline already achieved a relatively high global Dice score of 0.908 and an IoU of 0.838, leaving limited room for improvement. Nevertheless, DualDiT was the only evaluated augmentation framework that improved the Baseline across several augmentation sizes. The best configuration, obtained with $\mathcal{D}+1200$, reached a global Dice score of 0.927 and an IoU of 0.868, corresponding to relative improvements of 2.07\% and 3.50\%, respectively. These findings support the potential utility of augmenting limited training datasets with images and masks generated by DualDiT to aid the development of segmentation models from limited datasets. Regarding the quality inspection, the expert evaluation showed that synthetic images and masks were not always readily distinguishable from real samples. These findings suggest that DualDiT was able to reproduce both the general appearance of real OCT inputs and visual characteristics associated with the resin and physiological embedding media, with a degree of perceptual plausibility recognised by the evaluators.

Regarding the limitations of this study, the resolution and size of the generated images, as well as the amount of training data, can pose challenges to the model's applicability in clinical settings. Although the resolution used in this study ($512\times256$) proved adequate for the experiments conducted, real-world clinical scenarios may require higher resolutions to capture greater detail for diagnostic and planning tasks. It should be noted that this limitation is primarily due to computational constraints and that the proposed approach is scalable to higher resolutions (e.g., $1024\times512$) at the cost of increased model complexity and hardware resource requirements. 

Similarly, the limited size of the training dataset, together with the small number of annotated samples and independent test volumes, may constrain both the model's ability to generalise and the statistical strength of the downstream evaluation. The reported improvements, therefore, provide descriptive evidence of a consistent benefit, but further validation across larger datasets is required to establish statistical significance and generalisability. In particular, because B-scans within the same volume are highly correlated, the effective diversity of the dataset is smaller than the total number of annotated B-scans might suggest. Future research should evaluate the method across a wider range of acquisition conditions, specimens, and independent datasets. In this regard, it is important to note that obtaining datasets with pixel-level annotations of retinal layers is itself a significant challenge in medical imaging, as it is labour-intensive and requires expert knowledge.

Finally, the generation of synthetic data using diffusion models involves a trade-off between the diversity of the generated outputs and their fidelity to the real world. Although the proposed approach achieves a high degree of realism in the synthesised images, future research could focus on increasing the variability of the samples while maintaining this realism.

In conclusion, this work introduces a novel conditional DualDiT framework for jointly generating OCT images and segmentation masks, showing that it provides more favourable generative and downstream segmentation results than the evaluated DDPM- and LDM-based alternatives. Beyond its quantitative improvements, the proposed method highlights the potential of jointly modelling images and annotations as an effective data augmentation strategy in data-scarce medical scenarios. In general, these results suggest that transformer-based diffusion models constitute a promising direction for advancing synthetic data generation and supporting the development of robust medical image analysis systems.

\section{Acknowledgments and declarations}

\subsection*{Acknowledgments}

We gratefully acknowledge Steffi Ketelhut, from the Biomedical Technology Center of the Medical Faculty of the University of Muenster, for her collaboration in the animal experimentation.

We also thank the support from the Generalitat Valenciana (GVA) with the donation of the DGX A100 used for this work, an action co-financed by the European Union through the Operational Program of the European Regional Development Fund of the Comunitat Valenciana 2014-2020 (IDIFEDER/2020/030).

\subsection*{Funding}
This work was funded by Horizon Europe, the European Union's Framework Programme for Research and Innovation, under Grant Agreement No. 101070062 (SEQUOIA), by Horizon 2020 under Grant Agreement No. 732613 (GALAHAD), and by the Generalitat Valenciana under Grant CIPROM/2022/20 (COMTACTS2).

\subsection*{Ethics statement}
This study was conducted in accordance with the ARVO statement on the use of animals in ophthalmic and vision research.
This study was approved by the LANUV North Rhine-Westphalia, Recklinghausen, Germany, and the Animal Protection Office, University of Münster (Approval No. 84-02.04.2016.A395 and T24.036UMS).

\subsection*{Competing interests}

The authors declare no competing interests.

\subsection*{Declaration of generative AI and AI-assisted technologies in the manuscript preparation process}

During the preparation of this work, the authors used OpenAI ChatGPT (accessed in May 2026) to assist with language polishing and consistency checks. After using this tool/service, the authors reviewed and edited the content as needed and take full responsibility for the content of the published article.

\printcredits


\bibliographystyle{model1-num-names}

\bibliography{bibliography_1}

@Article{jimaging9040081,
AUTHOR = {Kebaili, Aghiles and Lapuyade-Lahorgue, Jérôme and Ruan, Su},
TITLE = {Deep Learning Approaches for Data Augmentation in Medical Imaging: A Review},
JOURNAL = {Journal of Imaging},
VOLUME = {9},
YEAR = {2023},
NUMBER = {4},
ARTICLE-NUMBER = {81},
PubMedID = {37103232},
ISSN = {2313-433X},
doi = {10.3390/jimaging9040081}
}

@article{celard_survey_2023,
	title = {A survey on deep learning applied to medical images: from simple artificial neural networks to generative models},
	volume = {35},
	issn = {1433-3058},
	doi = {10.1007/s00521-022-07953-4},
	shorttitle = {A survey on deep learning applied to medical images},
	pages = {2291--2323},
	number = {3},
	journaltitle = {Neural Computing and Applications},
	shortjournal = {Neural Comput \& Applic},
	author = {Celard, P. and Iglesias, E. L. and Sorribes-Fdez, J. M. and Romero, R. and Vieira, A. Seara and Borrajo, L.},
	urldate = {2025-10-24},
	date = {2023-01-01},
	year = {2023},
	langid = {english},
}

@article{islam_generative_2024,
	title = {Generative Adversarial Networks ({GANs}) in Medical Imaging: Advancements, Applications, and Challenges},
	volume = {12},
	issn = {2169-3536},
	doi = {10.1109/ACCESS.2024.3370848},
	shorttitle = {Generative Adversarial Networks ({GANs}) in Medical Imaging},
	pages = {35728--35753},
	journaltitle = {{IEEE} Access},
	author = {Islam, Showrov and Aziz, Md. Tarek and Nabil, Hadiur Rahman and Jim, Jamin Rahman and Mridha, M. F. and Kabir, Md. Mohsin and Asai, Nobuyoshi and Shin, Jungpil},
	urldate = {2025-10-24},
	date = {2024},
	year = {2024},
}

@article{oulmalme_systematic_2025,
	title = {A systematic review of generative {AI} approaches for medical image enhancement: Comparing {GANs}, transformers, and diffusion models},
	volume = {199},
	issn = {1386-5056},
	doi = {10.1016/j.ijmedinf.2025.105903},
	shorttitle = {A systematic review of generative {AI} approaches for medical image enhancement},
	pages = {105903},
	journaltitle = {International Journal of Medical Informatics},
	shortjournal = {International Journal of Medical Informatics},
	author = {Oulmalme, Chaimaa and Nakouri, Haïfa and Jaafar, Fehmi},
	urldate = {2025-10-24},
	date = {2025-07-01},
	year = {2025},
}

@article{kingma_introduction_2019,
	title = {An Introduction to Variational Autoencoders},
	volume = {12},
	issn = {1935-8237},
	url = {https://doi.org/10.1561/2200000056},
	doi = {10.1561/2200000056},
	pages = {307--392},
	number = {4},
	journaltitle = {Foundations and Trends in Machine Learning},
	author = {Kingma, Diederik P. and Welling, Max},
	urldate = {2026-06-23},
	date = {2019-11-28},
	year = {2019},
}

@article{chlap_review_2021,
	title = {A review of medical image data augmentation techniques for deep learning applications},
	volume = {65},
	rights = {© 2021 The Royal Australian and New Zealand College of Radiologists},
	issn = {1754-9485},
	doi = {10.1111/1754-9485.13261},
	pages = {545--563},
	number = {5},
	journaltitle = {Journal of Medical Imaging and Radiation Oncology},
	author = {Chlap, Phillip and Min, Hang and Vandenberg, Nym and Dowling, Jason and Holloway, Lois and Haworth, Annette},
	urldate = {2025-10-24},
	date = {2021},
	year = {2021},
	langid = {english},
}

@article{chataut_generative_2025,
	title = {Generative Artificial Intelligence in Healthcare: A Systematic Review of {GANs}, Diffusion Models, Large Language Models, and Variational Autoencoders for Medical Applications},
	volume = {3},
	issn = {2786-9342},
	doi = {10.59324/ejaset.2025.3(4).16},
	shorttitle = {Generative Artificial Intelligence in Healthcare},
	pages = {182--203},
	number = {4},
	journaltitle = {European Journal of Applied Science, Engineering and Technology},
	author = {Chataut, Sandeep and Bhatta, Srijana and Dahal, Bishwambhar and Ojha, Grishma and Raut, Srijana and Subedi, Bigyan and Bastakoti, Bijay},
	urldate = {2025-10-24},
	date = {2025-08-06},
	year = {2025},
	langid = {american},
}

@article{rguibi_medical_2023,
	title = {Medical variational autoencoder and generative adversarial network for medical imaging},
	volume = {32},
	doi = {10.11591/ijeecs.v32.i1.pp494-505},
	pages = {1∼1x},
	journaltitle = {Indonesian Journal of Electrical Engineering and Computer Science},
	shortjournal = {Indonesian Journal of Electrical Engineering and Computer Science},
	author = {Rguibi, Zakaria and Hajami, Abdelmajid and Zitouni, Dya and Yassine, Maleh and Elqaraoui, Amine},
	date = {2023-10-01},
	year = {2023},
}

@article{shi_diffusion_2025,
	title = {Diffusion Models for Medical Image Computing: A Survey},
	volume = {30},
	issn = {1007-0214},
	url = {https://ieeexplore.ieee.org/document/10676408/},
	doi = {10.26599/TST.2024.9010047},
	shorttitle = {Diffusion Models for Medical Image Computing},
	pages = {357--383},
	number = {1},
	journaltitle = {Tsinghua Science and Technology},
	author = {Shi, Yaqing and Abulizi, Abudukelimu and Wang, Hao and Feng, Ke and Abudukelimu, Nihemaiti and Su, Youli and Abudukelimu, Halidanmu},
	urldate = {2026-06-23},
	date = {2025-02},
	year = {2025},
}

@article{wang_diffusion_2025,
	title = {Diffusion model for medical image denoising, reconstruction and translation},
	volume = {124},
	issn = {0895-6111},
	doi = {10.1016/j.compmedimag.2025.102593},
	pages = {102593},
	journaltitle = {Computerized Medical Imaging and Graphics},
	shortjournal = {Computerized Medical Imaging and Graphics},
	author = {Wang, Wei and Xia, Jiayu and Luo, Gongning and Dong, Suyu and Li, Xiangyu and Wen, Jie and Li, Shuo},
	urldate = {2025-10-24},
	date = {2025-09-01},
	year = {2025},
}

@inproceedings{li_retidiff_2026,
	location = {Cham},
	title = {{RetiDiff}: Diffusion-Based Synthesis of Retinal {OCT} Images for Enhanced Segmentation},
	isbn = {978-3-032-04937-7},
	doi = {10.1007/978-3-032-04937-7_49},
	shorttitle = {{RetiDiff}},
	pages = {516--525},
	booktitle = {Medical Image Computing and Computer Assisted Intervention – {MICCAI} 2025},
	publisher = {Springer Nature Switzerland},
	author = {Li, Sicheng and Dan, Mai and Chu, Yuhui and Yu, Jiahui and Zhao, Yunpeng and Zhao, Pengpeng},
	editor = {Gee, James C. and Alexander, Daniel C. and Hong, Jaesung and Iglesias, Juan Eugenio and Sudre, Carole H. and Venkataraman, Archana and Golland, Polina and Kim, Jong Hyo and Park, Jinah},
	date = {2026},
	year = {2026},
	langid = {english},
}

@article{abbasi_physics-informed_2025,
	title = {A Physics-Informed Diffusion Model for Super-Resolved Reconstruction of Optical Coherence Tomography Data},
	volume = {72},
	issn = {1558-2531},
	doi = {10.1109/TBME.2025.3556794},
	pages = {2937--2946},
	number = {10},
	journaltitle = {{IEEE} Transactions on Biomedical Engineering},
	author = {Abbasi, Nima and Wong, Alexander and Bizheva, Kostadinka},
	urldate = {2026-02-11},
	date = {2025-10},
	year = {2025},
}

@inproceedings{wu_retinal_2024,
	title = {Retinal {OCT} Synthesis with Denoising Diffusion Probabilistic Models for Layer Segmentation},
	issn = {1945-8452},
	doi = {10.1109/ISBI56570.2024.10635836},
	eventtitle = {2024 {IEEE} International Symposium on Biomedical Imaging ({ISBI})},
	pages = {1--5},
	booktitle = {2024 {IEEE} International Symposium on Biomedical Imaging ({ISBI})},
	author = {Wu, Yuli and He, Weidong and Eschweiler, Dennis and Dou, Ningxin and Fan, Zixin and Mi, Shengli and Walter, Peter and Stegmaier, Johannes},
	urldate = {2026-02-11},
	date = {2024-05},
	year = {2024},
}

@article{tian_octdiff_2026,
	title = {{OCTDiff}: Bridged Diffusion Model for Portable {OCT} Super-Resolution and Enhancement},
	volume = {38},
	url = {https://proceedings.neurips.cc/paper_files/paper/2025/hash/3b13a29a93c85055e82e3b69881ebb37-Abstract-Conference.html},
	shorttitle = {{OCTDiff}},
	pages = {41445--41465},
	journaltitle = {Advances in Neural Information Processing Systems},
	author = {Tian, Ye and {McCarthy}, Angela and Gomide, Gabriel and Liddle, Nancy and Golebka, Jedrzej and Chen, Royce and Liebmann, Jeff and Thakoor, Kaveri},
	urldate = {2026-06-23},
	date = {2026-04-23},
	year = {2026},
	langid = {english},
}

@article{yang_diffusiondci_2024,
	title = {{DiffusionDCI}: A Novel Diffusion-Based Unified Framework for Dynamic Full-Field {OCT} Image Generation and Segmentation},
	volume = {12},
	issn = {2169-3536},
	doi = {10.1109/ACCESS.2024.3372863},
	shorttitle = {{DiffusionDCI}},
	pages = {37702--37714},
	journaltitle = {{IEEE} Access},
	author = {Yang, Bin and Li, Jianqiang and Wang, Jingyi and Li, Ruiqi and Gu, Ke and Liu, Bo},
	urldate = {2026-02-11},
	date = {2024},
	year = {2024},
}

@article{ahmed_denoising_2024,
	title = {Denoising of Optical Coherence Tomography Images in Ophthalmology Using Deep Learning: A Systematic Review},
	volume = {10},
	rights = {http://creativecommons.org/licenses/by/3.0/},
	issn = {2313-433X},
	doi = {10.3390/jimaging10040086},
	shorttitle = {Denoising of Optical Coherence Tomography Images in Ophthalmology Using Deep Learning},
	pages = {86},
	number = {4},
	journaltitle = {Journal of Imaging},
	publisher = {Multidisciplinary Digital Publishing Institute},
	author = {Ahmed, Hanya and Zhang, Qianni and Donnan, Robert and Alomainy, Akram},
	urldate = {2026-02-11},
	date = {2024-04},
	year = {2024},
	langid = {english},
}

@article{du_benchmarking_2025,
	title = {Benchmarking diffusion models against state-of-the-art architectures for {OCT} fluid biomarker segmentation},
	volume = {20},
	issn = {1932-6203},
	doi = {10.1371/journal.pone.0335615},
	pages = {e0335615},
	number = {10},
	journaltitle = {{PLOS} {ONE}},
	shortjournal = {{PLOS} {ONE}},
	publisher = {Public Library of Science},
	author = {Du, Katherine and Doshi, Utkarsh and {DiCenzo}, Benjamin and Jiang, Jessica and Wu, Ethan and Gadari, Adarsh and Vupparaboina, Sharat Chandra and Sadeghi, Elham and Bollepalli, Sandeep Chandra and Sahel, José-Alain and Chhablani, Jay and Vupparaboina, Kiran Kumar},
	urldate = {2026-02-11},
	date = {2025-10-29},
	year = {2025},
	langid = {english},
}

@inproceedings{huang_memory-efficient_2024,
	location = {Cham},
	title = {Memory-Efficient High-Resolution {OCT} Volume Synthesis with Cascaded Amortized Latent Diffusion Models},
	isbn = {978-3-031-72104-5},
	doi = {10.1007/978-3-031-72104-5_46},
	pages = {478--487},
	booktitle = {Medical Image Computing and Computer Assisted Intervention – {MICCAI} 2024},
	publisher = {Springer Nature Switzerland},
	author = {Huang, Kun and Ma, Xiao and Zhang, Yuhan and Su, Na and Yuan, Songtao and Liu, Yong and Chen, Qiang and Fu, Huazhu},
	editor = {Linguraru, Marius George and Dou, Qi and Feragen, Aasa and Giannarou, Stamatia and Glocker, Ben and Lekadir, Karim and Schnabel, Julia A.},
	date = {2024},
	year = {2024},
	langid = {english},
}

@article{badhon_diffusion_2025,
	title = {Diffusion model based {OCT} to {OCTA} translation},
	volume = {12},
	issn = {2296-858X},
	doi = {10.3389/fmed.2025.1655453},
	journaltitle = {Frontiers in Medicine},
	shortjournal = {Front. Med.},
	publisher = {Frontiers},
	author = {Badhon, Rashadul Hasan and Thompson, Atalie Carina and Lim, Jennifer I. and Leng, Theodore and Alam, Minhaj Nur},
	urldate = {2026-02-11},
	date = {2025-11-28},
	year = {2025},
}

@misc{rombach_high-resolution_2022-1,
	title = {High-Resolution Image Synthesis with Latent Diffusion Models},
	doi = {10.48550/arXiv.2112.10752},
	number = {{arXiv}:2112.10752},
	publisher = {{arXiv}},
	author = {Rombach, Robin and Blattmann, Andreas and Lorenz, Dominik and Esser, Patrick and Ommer, Björn},
	urldate = {2026-02-12},
	date = {2022-04-13},
	year = {2022},
	eprinttype = {arxiv},
	eprint = {2112.10752 [cs]},
}

@misc{dhariwal_diffusion_2021,
	title = {Diffusion Models Beat {GANs} on Image Synthesis},
	doi = {10.48550/arXiv.2105.05233},
	number = {{arXiv}:2105.05233},
	publisher = {{arXiv}},
	author = {Dhariwal, Prafulla and Nichol, Alex},
	urldate = {2026-02-12},
	date = {2021-06-01},
	year = {2021},
	eprinttype = {arxiv},
	eprint = {2105.05233 [cs]},
}

@misc{ho_classier-free_nodate,
      title={Classifier-Free Diffusion Guidance}, 
      author={Jonathan Ho and Tim Salimans},
      year={2022},
      eprint={2207.12598},
      archivePrefix={arXiv},
      primaryClass={cs.LG},
      url={https://arxiv.org/abs/2207.12598}, 
}

@inproceedings{li_open-vocabulary_2023,
	title = {Open-vocabulary Object Segmentation with Diffusion Models},
	url = {https://openaccess.thecvf.com/content/ICCV2023/html/Li_Open-vocabulary_Object_Segmentation_with_Diffusion_Models_ICCV_2023_paper.html},
	eventtitle = {Proceedings of the {IEEE}/{CVF} International Conference on Computer Vision},
	pages = {7667--7676},
	author = {Li, Ziyi and Zhou, Qinye and Zhang, Xiaoyun and Zhang, Ya and Wang, Yanfeng and Xie, Weidi},
	urldate = {2026-06-23},
	date = {2023},
	year = {2023},
	langid = {english},
}

@inproceedings{ho_denoising_2020,
	title = {Denoising Diffusion Probabilistic Models},
	volume = {33},
	pages = {6840--6851},
	booktitle = {Advances in Neural Information Processing Systems},
	publisher = {Curran Associates, Inc.},
	author = {Ho, Jonathan and Jain, Ajay and Abbeel, Pieter},
	urldate = {2026-02-16},
	date = {2020},
	year = {2020},
    url = {https://proceedings.neurips.cc/paper_files/paper/2020/file/4c5bcfec8584af0d967f1ab10179ca4b-Paper.pdf}
}

@inproceedings{mao_medsegfactory_2025,
	title = {{MedSegFactory}: Text-Guided Generation of Medical Image-Mask Pairs},
	url = {https://openaccess.thecvf.com/content/ICCV2025/html/Mao_MedSegFactory_Text-Guided_Generation_of_Medical_Image-Mask_Pairs_ICCV_2025_paper.html},
	shorttitle = {{MedSegFactory}},
	eventtitle = {Proceedings of the {IEEE}/{CVF} International Conference on Computer Vision},
	pages = {21525--21535},
	author = {Mao, Jiawei and Wang, Yuhan and Tang, Yucheng and Xu, Daguang and Wang, Kang and Yang, Yang and Zhou, Zongwei and Zhou, Yuyin},
	urldate = {2026-02-16},
	date = {2025},
	year = {2025},
	langid = {english},
}

@inproceedings{wu_diffumask_2023,
	title = {{DiffuMask}: Synthesizing Images with Pixel-level Annotations for Semantic Segmentation Using Diffusion Models},
	url = {https://openaccess.thecvf.com/content/ICCV2023/html/Wu_DiffuMask_Synthesizing_Images_with_Pixel-level_Annotations_for_Semantic_Segmentation_Using_ICCV_2023_paper.html},
	shorttitle = {{DiffuMask}},
	eventtitle = {Proceedings of the {IEEE}/{CVF} International Conference on Computer Vision},
	pages = {1206--1217},
	author = {Wu, Weijia and Zhao, Yuzhong and Shou, Mike Zheng and Zhou, Hong and Shen, Chunhua},
	urldate = {2026-02-16},
	date = {2023},
	year = {2023},
	langid = {english},
}

@inproceedings{toker_satsynth_2024,
	title = {{SatSynth}: Augmenting Image-Mask Pairs through Diffusion Models for Aerial Semantic Segmentation},
	url = {https://openaccess.thecvf.com/content/CVPR2024/html/Toker_SatSynth_Augmenting_Image-Mask_Pairs_through_Diffusion_Models_for_Aerial_Semantic_CVPR_2024_paper.html},
	shorttitle = {{SatSynth}},
	eventtitle = {Proceedings of the {IEEE}/{CVF} Conference on Computer Vision and Pattern Recognition},
	pages = {27695--27705},
	author = {Toker, Aysim and Eisenberger, Marvin and Cremers, Daniel and Leal-Taixé, Laura},
	urldate = {2026-02-16},
	date = {2024},
	year = {2024},
	langid = {english},
}

@article{park_seediff_2025,
	title = {{SeeDiff}: Off-the-Shelf Seeded Mask Generation from Diffusion Models},
	volume = {39},
	rights = {Copyright (c) 2025 Association for the Advancement of Artificial Intelligence},
	issn = {2374-3468},
	doi = {10.1609/aaai.v39i6.32686},
	shorttitle = {{SeeDiff}},
	pages = {6406--6415},
	number = {6},
	journaltitle = {Proceedings of the {AAAI} Conference on Artificial Intelligence},
	author = {Park, Joon Hyun and Jo, Kumju and Baik, Sungyong},
	urldate = {2026-02-16},
	date = {2025-04-11},
	year = {2025},
	langid = {english},
}

@inproceedings{frisch_gauda_2025,
	title = {{GAUDA}: Generative Adaptive Uncertainty-Guided Diffusion-Based Augmentation for Surgical Segmentation},
	issn = {2642-9381},
	doi = {10.1109/WACV61041.2025.00370},
	shorttitle = {{GAUDA}},
	eventtitle = {2025 {IEEE}/{CVF} Winter Conference on Applications of Computer Vision ({WACV})},
	pages = {3762--3771},
	booktitle = {2025 {IEEE}/{CVF} Winter Conference on Applications of Computer Vision ({WACV})},
	author = {Frisch, Yannik and Bornberg, Christina and Fuchs, Moritz and Mukhopadhyay, Anirban},
	urldate = {2026-02-16},
	date = {2025-02},
	year = {2025},
	note = {{ISSN}: 2642-9381},
}

@inproceedings{peebles_scalable_2023,
	title = {Scalable Diffusion Models with Transformers},
	url = {https://openaccess.thecvf.com/content/ICCV2023/html/Peebles_Scalable_Diffusion_Models_with_Transformers_ICCV_2023_paper.html},
	eventtitle = {Proceedings of the {IEEE}/{CVF} International Conference on Computer Vision},
	pages = {4195--4205},
	author = {Peebles, William and Xie, Saining},
	urldate = {2026-06-23},
	date = {2023},
	year = {2023},
	langid = {english},
}

@misc{kingma_adam_2017,
	title = {Adam: A Method for Stochastic Optimization},
	url = {http://arxiv.org/abs/1412.6980},
	doi = {10.48550/arXiv.1412.6980},
	shorttitle = {Adam},
	number = {{arXiv}:1412.6980},
	publisher = {{arXiv}},
	author = {Kingma, Diederik P. and Ba, Jimmy},
	urldate = {2026-06-23},
	date = {2017-01-30},
	year = {2017},
	eprinttype = {arxiv},
	eprint = {1412.6980 [cs.LG]},
}

@online{noauthor_stabilityaisd-vae-ft-ema_nodate,
	title = {stabilityai/sd-vae-ft-ema · Hugging Face},
	url = {https://huggingface.co/stabilityai/sd-vae-ft-ema},
	year = {2022},
	urldate = {2026-02-19},
}

@inproceedings{heusel_gans_2017,
	title = {{GANs} Trained by a Two Time-Scale Update Rule Converge to a Local Nash Equilibrium},
	volume = {30},
	url = {https://proceedings.neurips.cc/paper/2017/hash/8a1d694707eb0fefe65871369074926d-Abstract.html},
	booktitle = {Advances in Neural Information Processing Systems},
	publisher = {Curran Associates, Inc.},
	author = {Heusel, Martin and Ramsauer, Hubert and Unterthiner, Thomas and Nessler, Bernhard and Hochreiter, Sepp},
	urldate = {2026-06-23},
	date = {2017},
	year = {2017},
}

@article{morales_retinal_2021,
	title = {Retinal layer segmentation in rodent {OCT} images: Local intensity profiles \& fully convolutional neural networks},
	volume = {198},
	issn = {0169-2607},
	doi = {10.1016/j.cmpb.2020.105788},
	shorttitle = {Retinal layer segmentation in rodent {OCT} images},
	pages = {105788},
	journaltitle = {Computer Methods and Programs in Biomedicine},
	shortjournal = {Computer Methods and Programs in Biomedicine},
	author = {Morales, Sandra and Colomer, Adrián and Mossi, José M. and del Amor, Rocío and Woldbye, David and Klemp, Kristian and Larsen, Michael and Naranjo, Valery},
	urldate = {2024-06-21},
	date = {2021-01-01},
	year = {2021},
}

@inproceedings{barroso_durable_2024,
	title = {Durable ex vivo mouse retina 3D tissue models for optical coherence tomography},
	volume = {12854},
	doi = {10.1117/12.3002538},
	eventtitle = {Label-free Biomedical Imaging and Sensing ({LBIS}) 2024},
	pages = {21--23},
	booktitle = {Label-free Biomedical Imaging and Sensing ({LBIS}) 2024},
	publisher = {{SPIE}},
	author = {Barroso, \'{A}lvaro and Heiduschka, Peter and Nettels-Hackert, Gerburg and Ketelhut, Steffi and Amor, Rocío del and García-Torres, Fernando and Morales-Martínez, Sandra and Naranjo, Valery and Kemper, Björn and Schnekenburger, Jürgen},
	urldate = {2024-06-21},
	date = {2024-03-12},
	year = {2024},
}

@article{barroso_durable_2023,
	title = {Durable 3D murine ex vivo retina glaucoma models for optical coherence tomography},
	volume = {14},
	issn = {2156-7085},
	doi = {10.1364/BOE.494271},
	pages = {4421--4438},
	number = {9},
	journaltitle = {Biomedical Optics Express},
	shortjournal = {Biomed. Opt. Express, {BOE}},
	publisher = {Optica Publishing Group},
	author = {Barroso, \'{A}lvaro and Ketelhut, Steffi and Nettels-Hackert, Gerburg and Heiduschka, Peter and Amor, Rocío del and Naranjo, Valery and Kemper, Björn and Schnekenburger, Jürgen},
	urldate = {2024-06-21},
	date = {2023-09-01},
	year = {2023},
}

@inproceedings{amor_towards_2019,
	location = {A Coruna, Spain},
	title = {Towards Automatic Glaucoma Assessment: An Encoder-decoder {CNN} for Retinal Layer Segmentation in Rodent {OCT} images},
	isbn = {978-90-827970-3-9},
	doi = {10.23919/EUSIPCO.2019.8902794},
	shorttitle = {Towards Automatic Glaucoma Assessment},
	eventtitle = {2019 27th European Signal Processing Conference ({EUSIPCO})},
	pages = {1--5},
	booktitle = {2019 27th European Signal Processing Conference ({EUSIPCO})},
	publisher = {{IEEE}},
	author = {Amor, Rocio Del and Morales, Sandra and Colomer, Adria N and Mossi, Jose M. and Woldbye, David and Klemp, Kristian and Larsen, Michael and Naranjo, Valery},
	urldate = {2024-06-21},
	date = {2019-09},
	year = {2019},
	langid = {english},
}

@inproceedings{garcia2024using,
author = {Garc\'{\i}a-Torres, Fernando and del Amor, Roc\'{\i}o and Morales-Mart\'{\i}nez, Sandra and Barroso, \'{A}lvaro and Kemper, Bj\"{o}rn and Schnekenburger, J\"{u}rgen and Naranjo, Valery},
title = {Using Diffusion Models for Data Augmentation on Limited Rodent OCT Datasets},
year = {2024},
isbn = {978-3-031-77730-1},
publisher = {Springer-Verlag},
address = {Berlin, Heidelberg},
doi = {10.1007/978-3-031-77731-8_29},
booktitle = {Intelligent Data Engineering and Automated Learning – IDEAL 2024: 25th International Conference, Valencia, Spain, November 20–22, 2024, Proceedings, Part I},
pages = {313–324},
numpages = {12},
location = {Valencia, Spain}
}

@inproceedings{ronneberger_u-net_2015,
	location = {Cham},
	title = {U-Net: Convolutional Networks for Biomedical Image Segmentation},
	isbn = {978-3-319-24574-4},
	doi = {10.1007/978-3-319-24574-4_28},
	shorttitle = {U-Net},
	pages = {234--241},
	booktitle = {Medical Image Computing and Computer-Assisted Intervention – {MICCAI} 2015},
	publisher = {Springer International Publishing},
	author = {Ronneberger, Olaf and Fischer, Philipp and Brox, Thomas},
	editor = {Navab, Nassir and Hornegger, Joachim and Wells, William M. and Frangi, Alejandro F.},
	date = {2015},
	year = {2015},
	langid = {english},
}

@INPROCEEDINGS{9897435,
  author={del Amor, Rocío and Colomer, Adrián and Morales, Sandra and Pulgarín-Ospina, Cristian and Terradez, Liria and Aneiros-Fernandez, Jose and Naranjo, Valery},
  booktitle={2022 IEEE International Conference on Image Processing (ICIP)}, 
  title={A Self-Contrastive Learning Framework for Skin Cancer Detection Using Histological Images}, 
  year={2022},
  volume={},
  number={},
  pages={2291-2295},
  doi={10.1109/ICIP46576.2022.9897435}}

@article{ma2021loss,
  title={Loss odyssey in medical image segmentation},
  author={Ma, Jun and Chen, Jianan and Ng, Matthew and Huang, Rui and Li, Yu and Li, Chen and Yang, Xiaoping and Martel, Anne L},
  journal={Medical image analysis},
  volume={71},
  pages={102035},
  year={2021},
  publisher={Elsevier},
  doi = {https://doi.org/10.1016/j.media.2021.102035}
}

@article{muller2022towards,
  title={Towards a guideline for evaluation metrics in medical image segmentation},
  author={M{\"u}ller, Dominik and Soto-Rey, I{\~n}aki and Kramer, Frank},
  journal={BMC research notes},
  volume={15},
  number={1},
  pages={210},
  year={2022},
  publisher={Springer},
  doi = {https://doi.org/10.1186/s13104-022-06096-y}
}

@inproceedings{isola_image--image_2017,
	title = {Image-To-Image Translation With Conditional Adversarial Networks},
	url = {https://openaccess.thecvf.com/content_cvpr_2017/html/Isola_Image-To-Image_Translation_With_CVPR_2017_paper.html},
	eventtitle = {Proceedings of the {IEEE} Conference on Computer Vision and Pattern Recognition},
	pages = {1125--1134},
	author = {Isola, Phillip and Zhu, Jun-Yan and Zhou, Tinghui and Efros, Alexei A.},
	urldate = {2026-06-23},
	date = {2017},
	year = {2017},
}

@article{yang2017dagan,
  title={DAGAN: deep de-aliasing generative adversarial networks for fast compressed sensing MRI reconstruction},
  author={Yang, Guang and Yu, Simiao and Dong, Hao and Slabaugh, Greg and Dragotti, Pier Luigi and Ye, Xujiong and Liu, Fangde and Arridge, Simon and Keegan, Jennifer and Guo, Yike and others},
  journal={IEEE transactions on medical imaging},
  volume={37},
  number={6},
  pages={1310--1321},
  year={2017},
  publisher={IEEE},
  doi = {https://doi.org/10.1109/TMI.2017.2785879}
}

@article{waheed2020covidgan,
  title={Covidgan: data augmentation using auxiliary classifier gan for improved covid-19 detection},
  author={Waheed, Abdul and Goyal, Muskan and Gupta, Deepak and Khanna, Ashish and Al-Turjman, Fadi and Pinheiro, Pl{\'a}cido Rogerio},
  journal={Ieee Access},
  volume={8},
  pages={91916--91923},
  year={2020},
  publisher={Ieee},
  doi = {10.1109/ACCESS.2020.2994762}
}

@article{muller2023multimodal,
  title={A multimodal comparison of latent denoising diffusion probabilistic models and generative adversarial networks for medical image synthesis},
  author={M{\"u}ller-Franzes, Gustav and Niehues, Jan Moritz and Khader, Firas and Arasteh, Soroosh Tayebi and Haarburger, Christoph and Kuhl, Christiane and Wang, Tianci and Han, Tianyu and Nolte, Teresa and Nebelung, Sven and others},
  journal={Scientific reports},
  volume={13},
  number={1},
  pages={12098},
  year={2023},
  publisher={Nature Publishing Group UK London},
  doi = {https://doi.org/10.1038/s41598-023-39278-0}
}

@article{mayer2010retinal,
  title={Retinal nerve fiber layer segmentation on FD-OCT scans of normal subjects and glaucoma patients},
  author={Mayer, Markus A and Hornegger, Joachim and Mardin, Christian Y and Tornow, Ralf P},
  journal={Biomedical optics express},
  volume={1},
  number={5},
  pages={1358--1383},
  year={2010},
  publisher={Optical Society of America},
  doi = {https://doi.org/10.1364/BOE.1.001358}, 
}

@article{allen2020vivo,
  title={In vivo structural assessments of ocular disease in rodent models using optical coherence tomography},
  author={Allen, Rachael S and Bales, Katie and Feola, Andrew and Pardue, Machelle T},
  journal={Journal of visualized experiments: JoVE},
  number={161},
  pages={10--3791},
  year={2020},
  doi = {10.3791/61588}
}

@article{fercher2003optical,
  title={Optical coherence tomography-principles and applications},
  author={Fercher, Adolf F and Drexler, Wolfgang and Hitzenberger, Christoph K and Lasser, Theo},
  journal={Reports on progress in physics},
  volume={66},
  number={2},
  pages={239--303},
  year={2003},
  doi = {10.1088/0034-4885/66/2/204}
}

@article{tschernig2013elegant,
  title={An elegant technique for ex vivo imaging in experimental research—Optical coherence tomography (OCT)},
  author={Tschernig, T and Thrane, Lars and J{\o}rgensen, Thomas Martini and Thommes, J and Pabst, R and Yelbuz, TM},
  journal={Annals of Anatomy-Anatomischer Anzeiger},
  volume={195},
  number={1},
  pages={25--27},
  year={2013},
  publisher={Elsevier},
  doi={https://doi.org/10.1016/j.aanat.2012.07.005}
}

\appendix
\renewcommand{\thefigure}{A.\arabic{figure}}
\setcounter{figure}{0}

\section{Qualitative Results of Joint Synthetic Image and Mask Generation}
\label{app1}

\begin{figure*}[H]
    \centering
    \setlength{\tabcolsep}{2pt} 
    \renewcommand{\arraystretch}{1.1} 

    \begin{tabular}{c c c c c c}
        \raisebox{2cm}{(a)} &
        \includegraphics[width=0.17\textwidth]{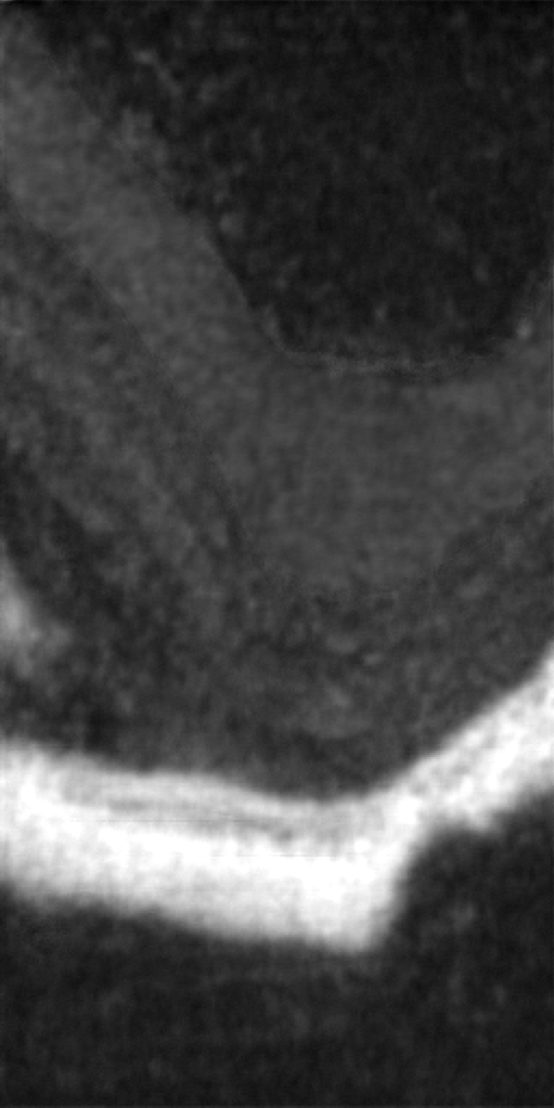} &
        \includegraphics[width=0.17\textwidth]{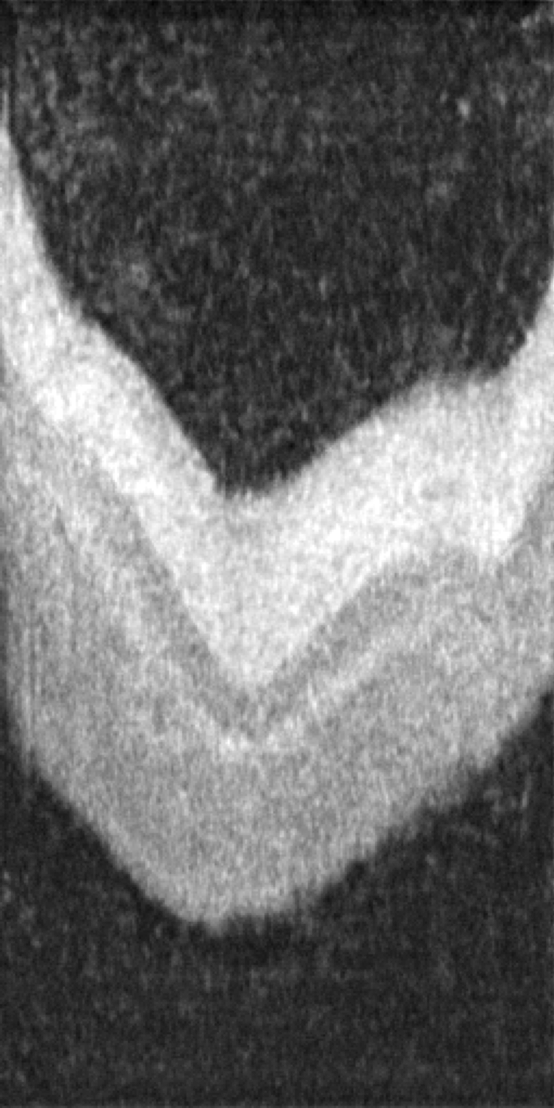} &
        \includegraphics[width=0.17\textwidth]{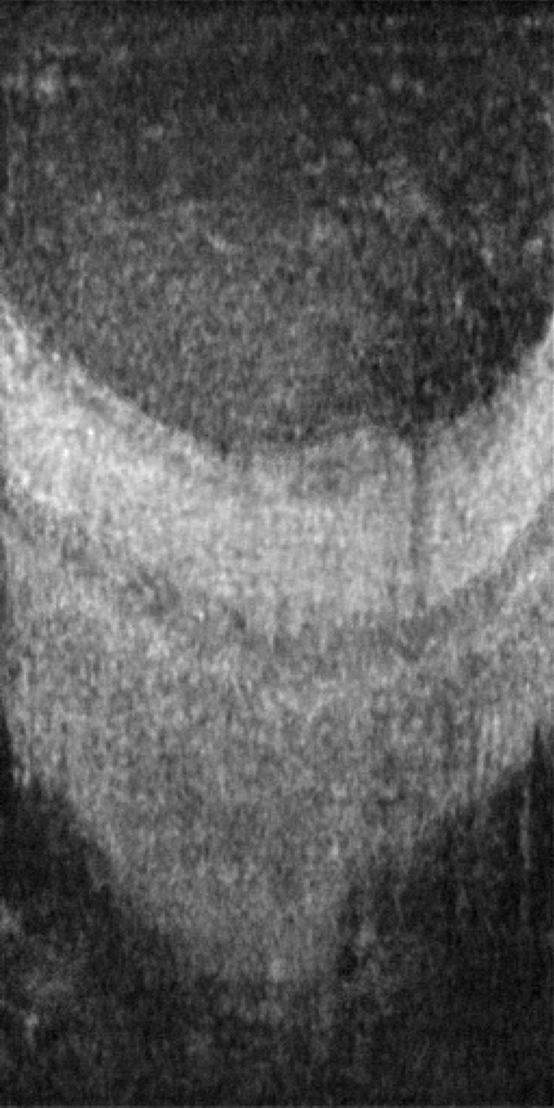} &
        \includegraphics[width=0.17\textwidth]{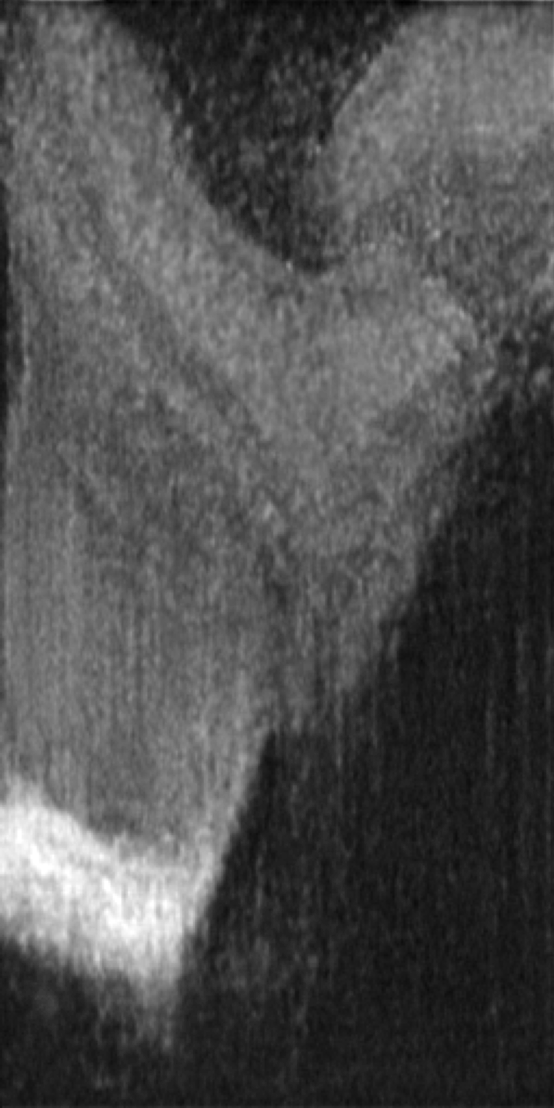} &
        \includegraphics[width=0.17\textwidth]{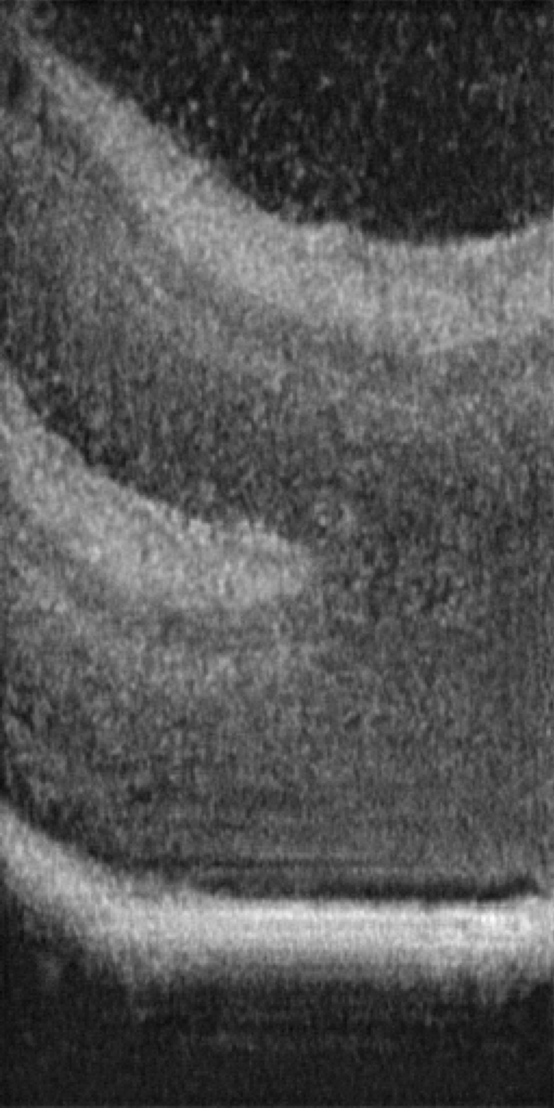} \\

        \raisebox{2cm}{(b)} &
        \includegraphics[width=0.17\textwidth]{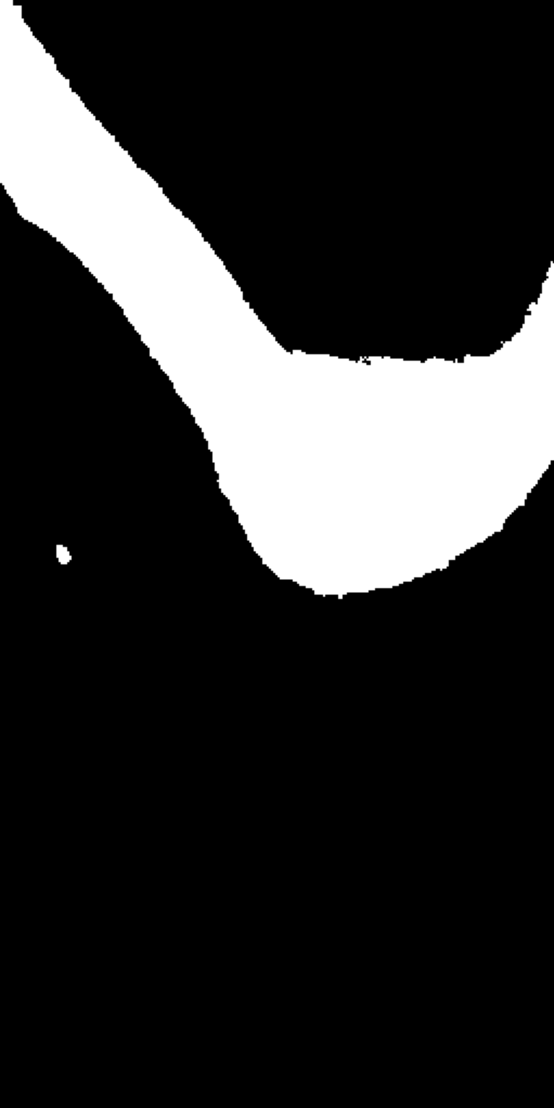} &
        \includegraphics[width=0.17\textwidth]{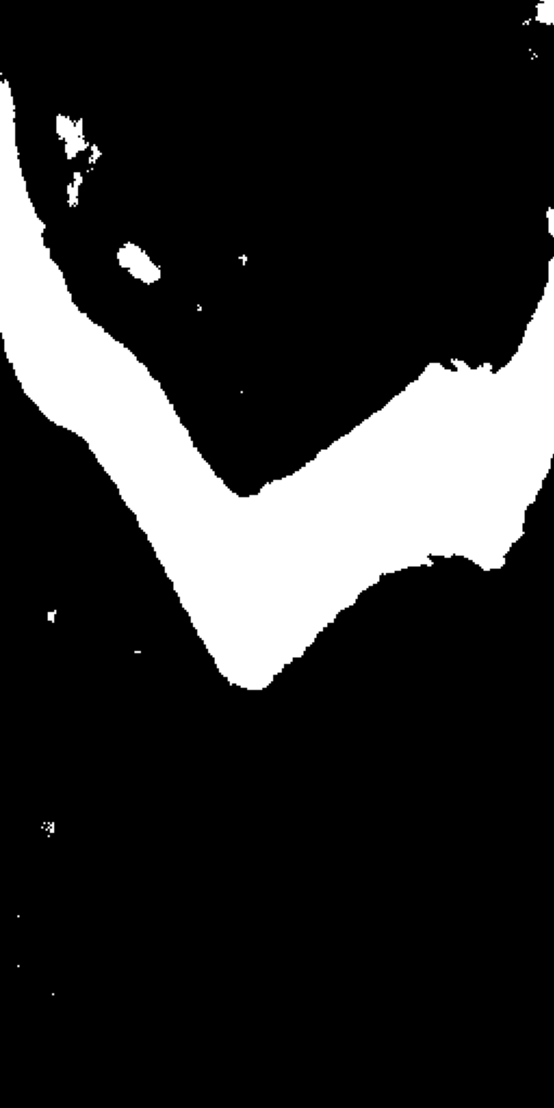} &
        \includegraphics[width=0.17\textwidth]{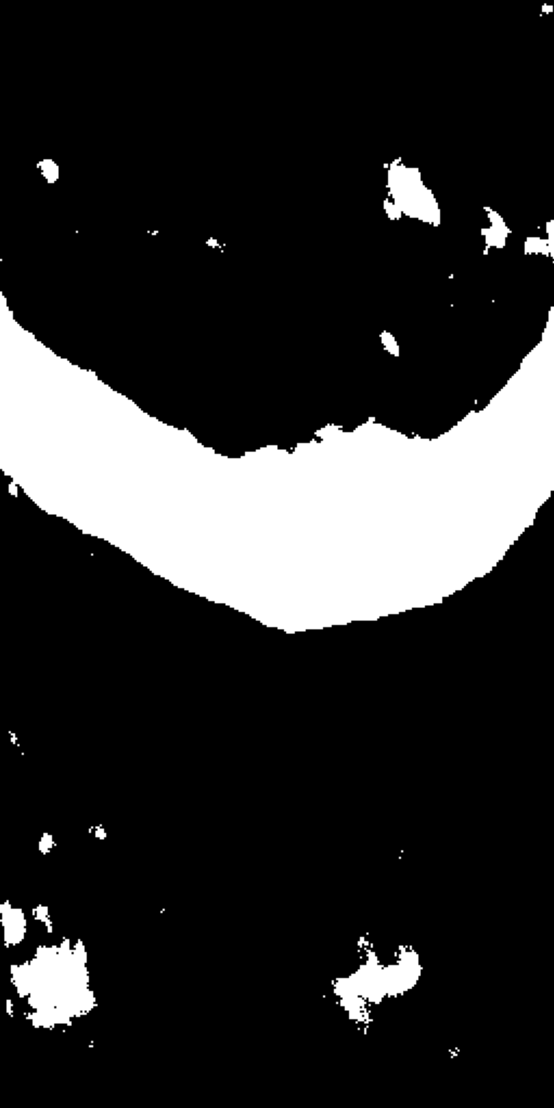} &
        \includegraphics[width=0.17\textwidth]{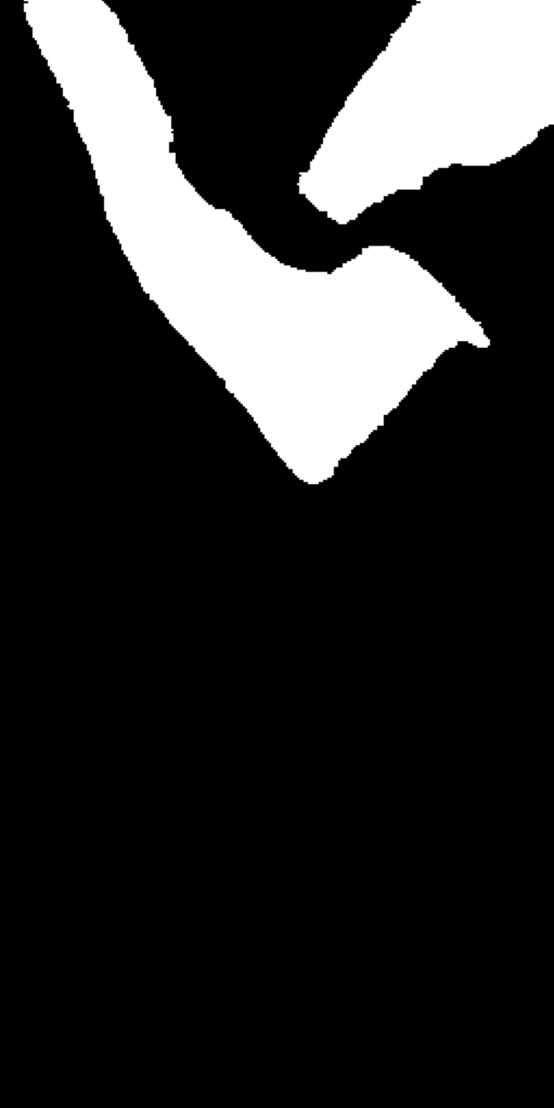} &
        \includegraphics[width=0.17\textwidth]{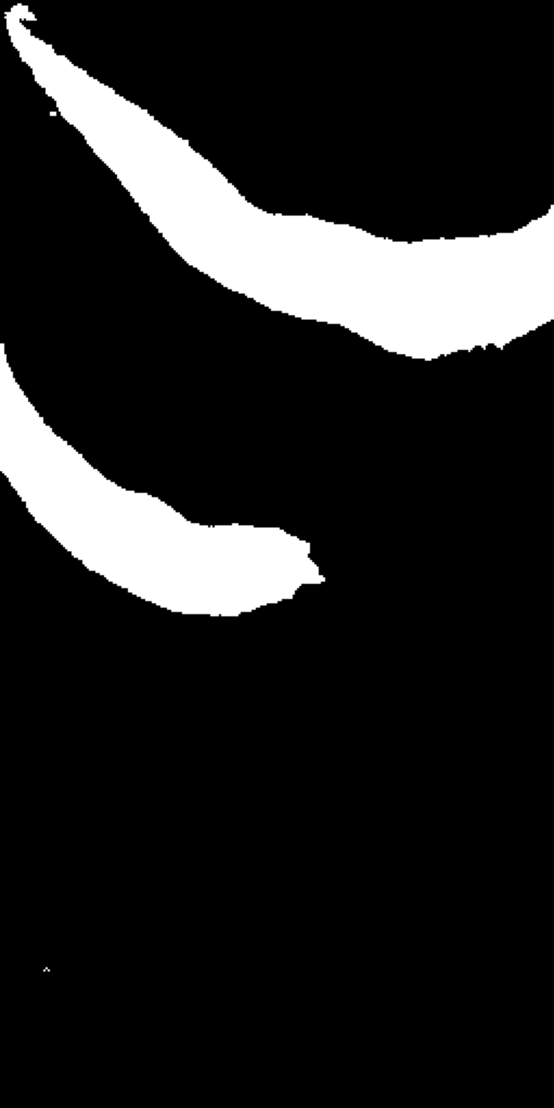} \\

        \raisebox{2cm}{(c)} &
        \includegraphics[width=0.17\textwidth]{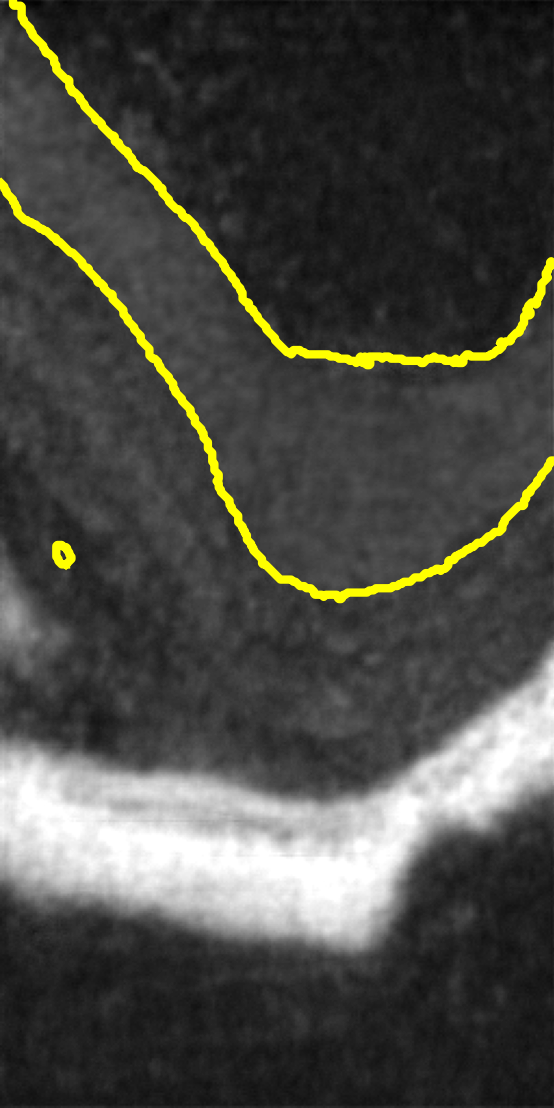} &
        \includegraphics[width=0.17\textwidth]{figs/appendix/DDPM/0/DDPM_overlay_3.png} &
        \includegraphics[width=0.17\textwidth]{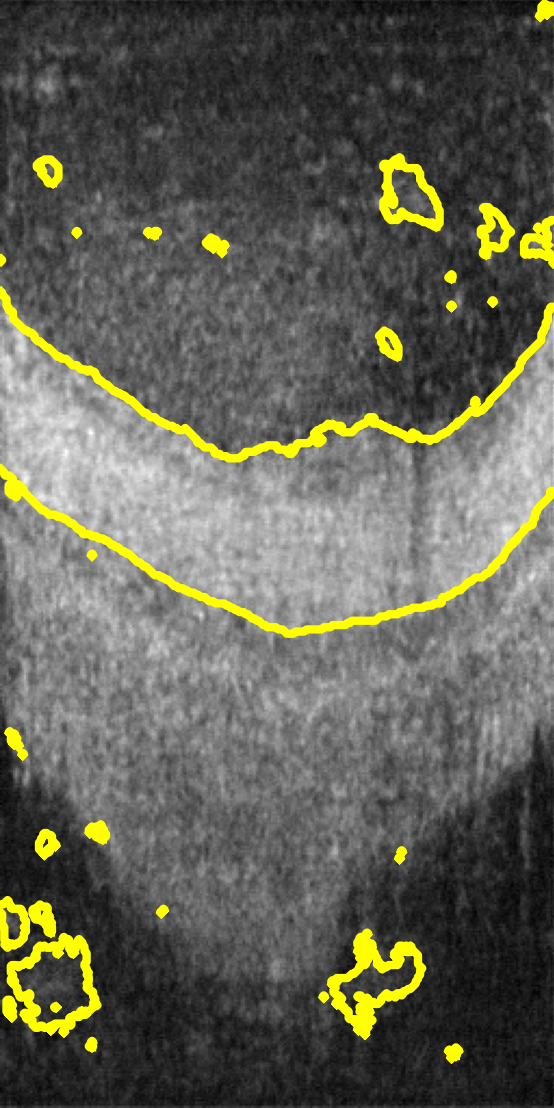} &
        \includegraphics[width=0.17\textwidth]{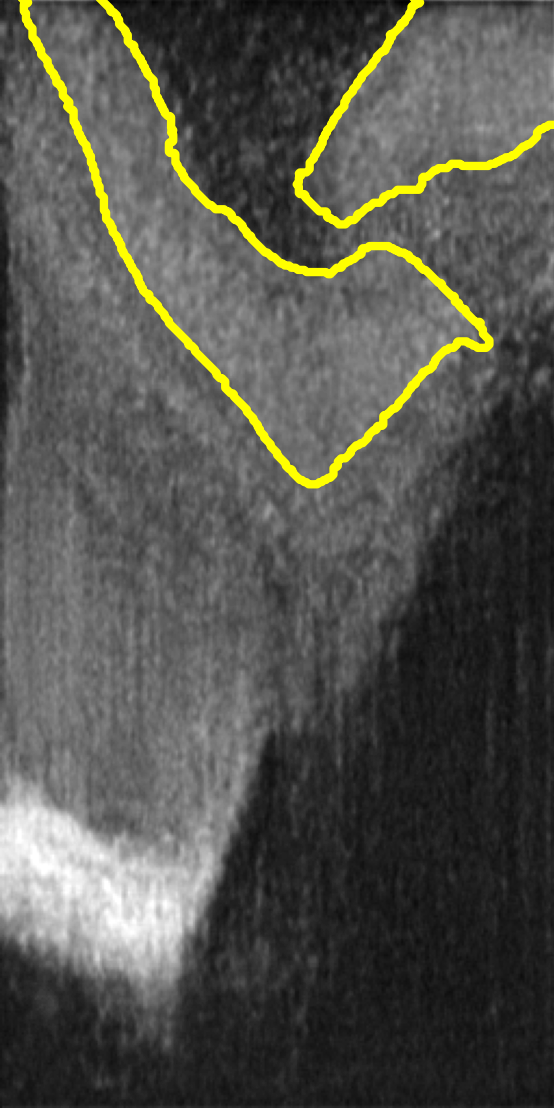} &
        \includegraphics[width=0.17\textwidth]{figs/appendix/DDPM/0/DDPM_overlay_6.png} \\
    \end{tabular}

    \caption{Synthetic resin-embedded images generated with DDPM \cite{ho_denoising_2020}. (a) Images; (b) masks; (c) overlaying the edges of the mask onto the image.}
    \label{fig:ddpm_0}
\end{figure*}

\begin{figure*}[H]
    \centering
    \setlength{\tabcolsep}{2pt}
    \renewcommand{\arraystretch}{1.1}

    \begin{tabular}{c c c c c c}
        \raisebox{2cm}{(a)} &
        \includegraphics[width=0.17\textwidth]{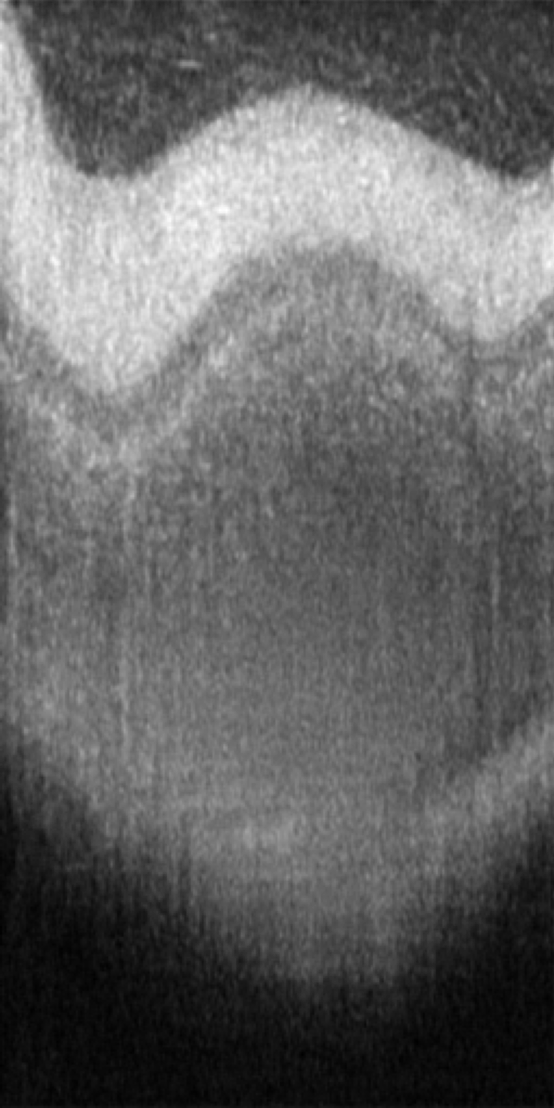} &
        \includegraphics[width=0.17\textwidth]{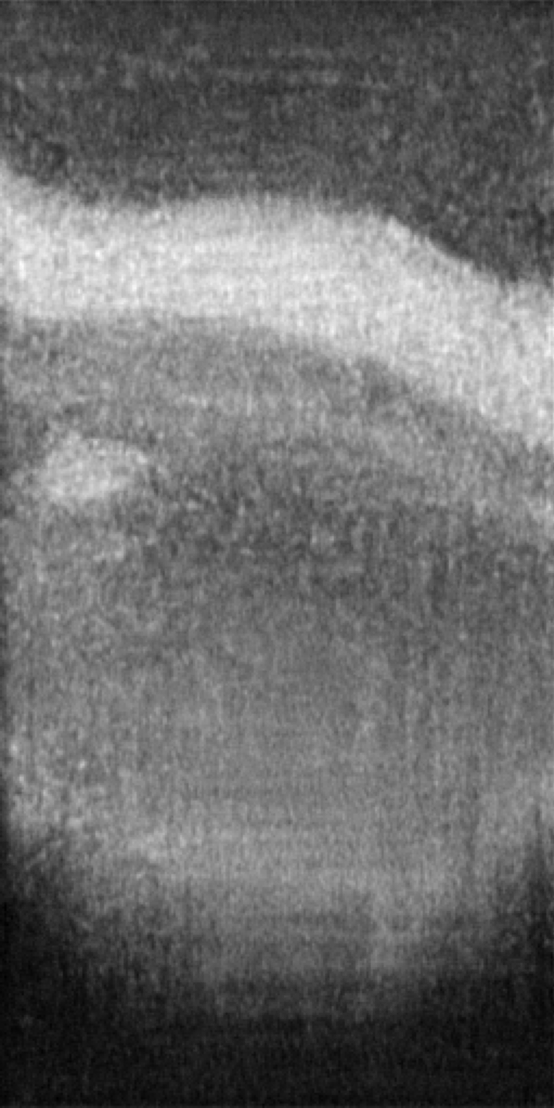} &
        \includegraphics[width=0.17\textwidth]{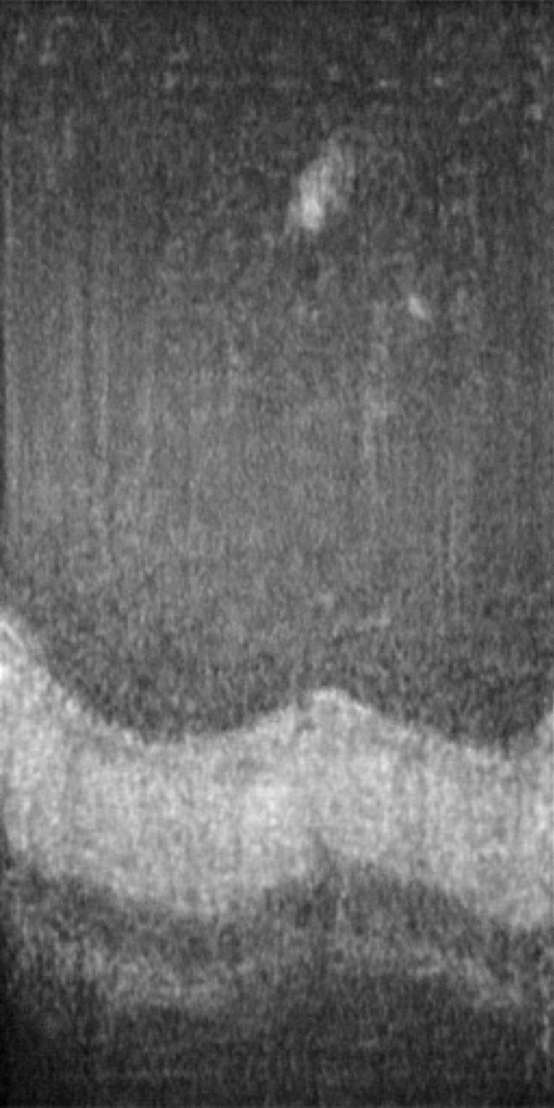} &
        \includegraphics[width=0.17\textwidth]{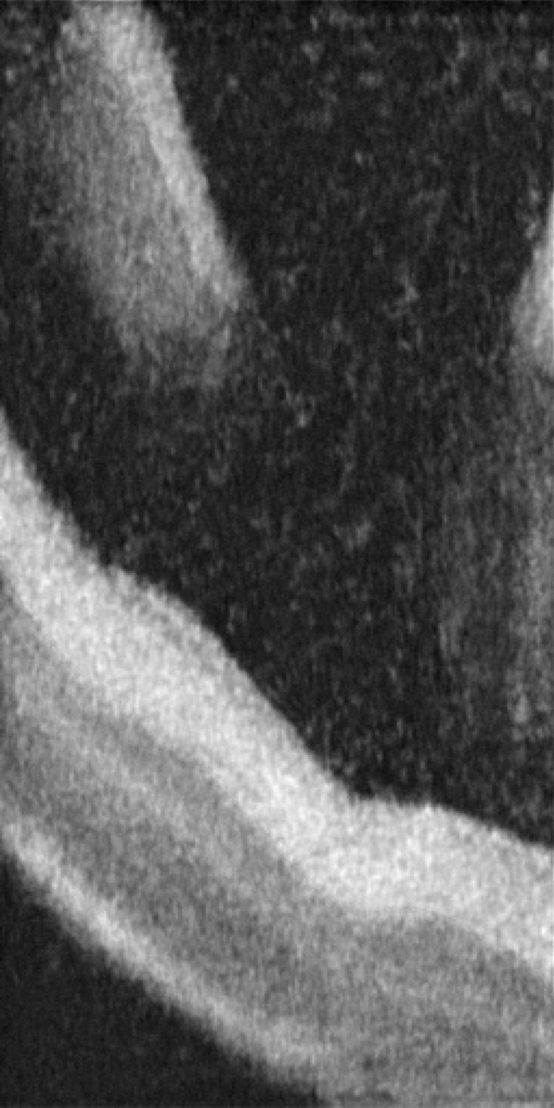} &
        \includegraphics[width=0.17\textwidth]{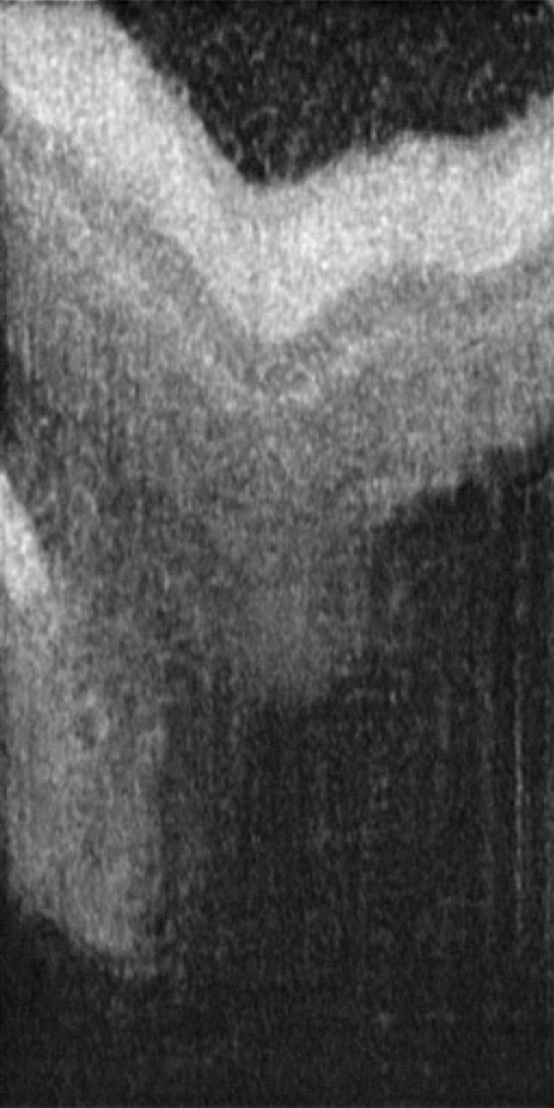} \\

        \raisebox{2cm}{(b)} &
        \includegraphics[width=0.17\textwidth]{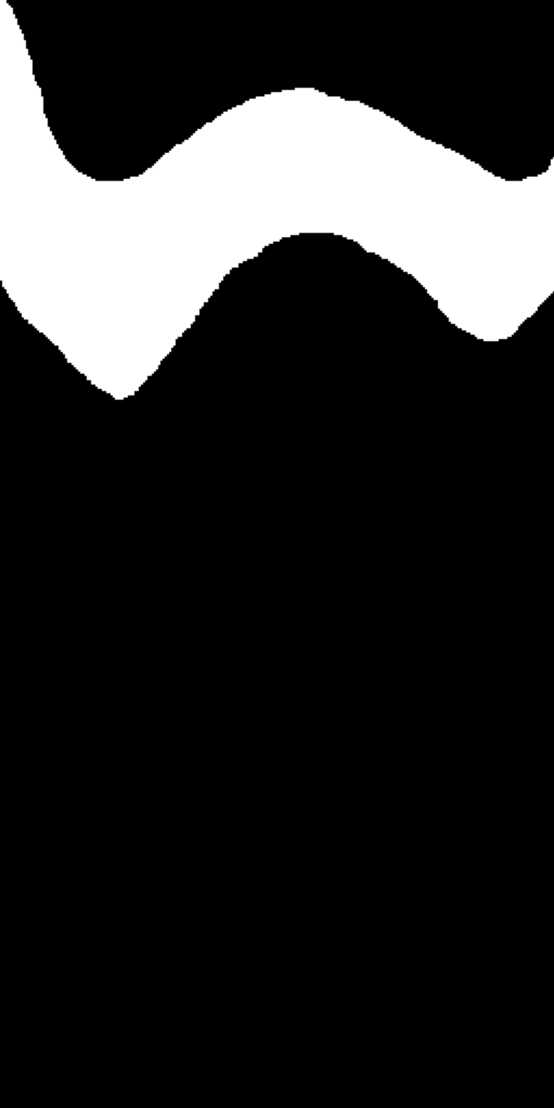} &
        \includegraphics[width=0.17\textwidth]{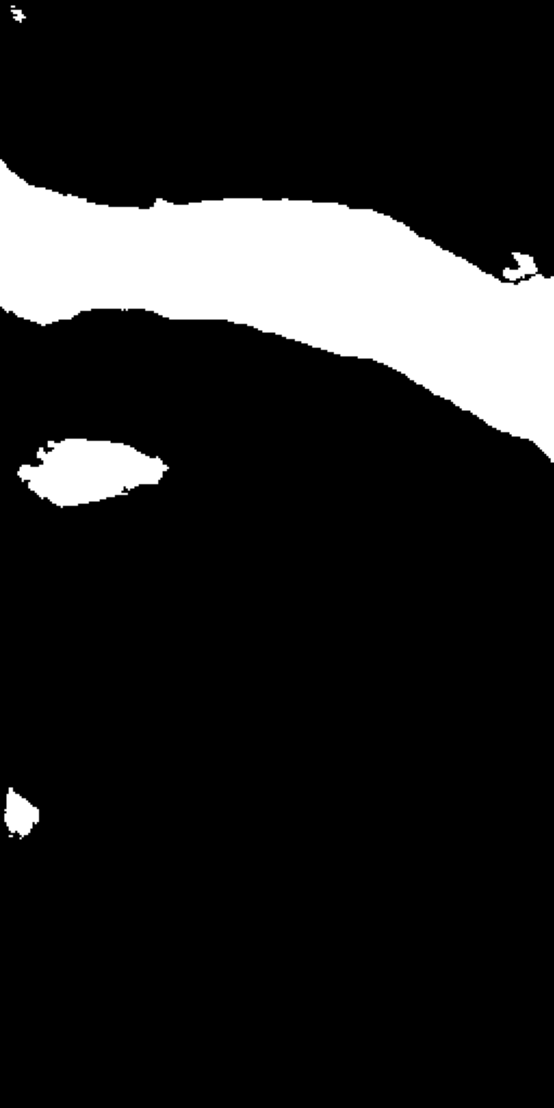} &
        \includegraphics[width=0.17\textwidth]{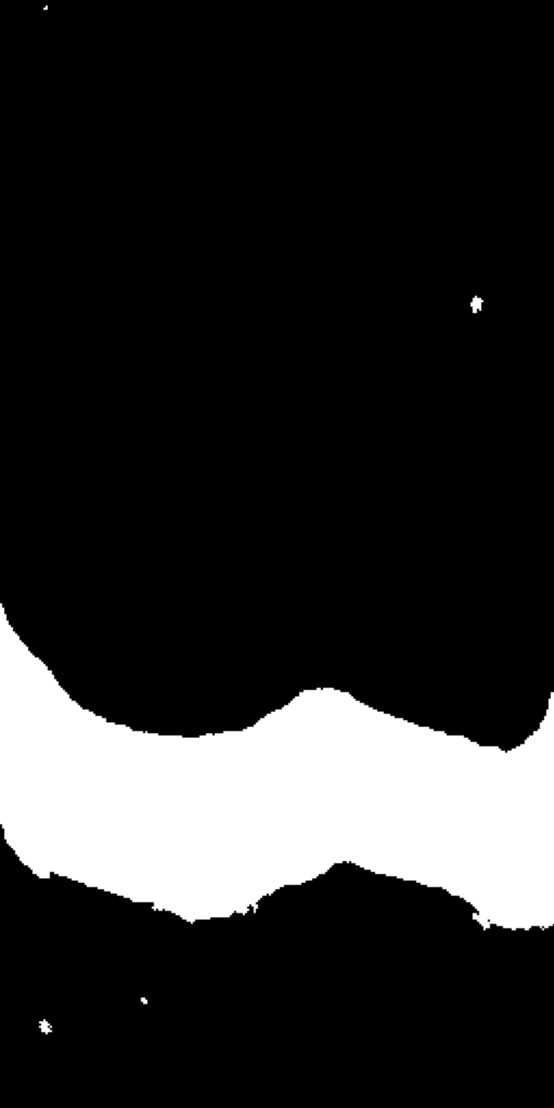} &
        \includegraphics[width=0.17\textwidth]{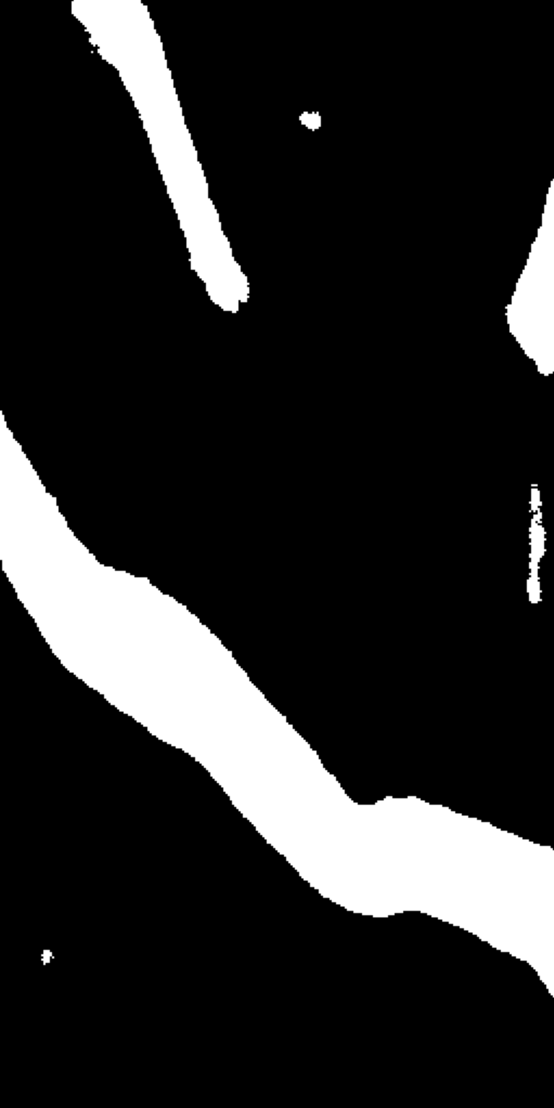} &
        \includegraphics[width=0.17\textwidth]{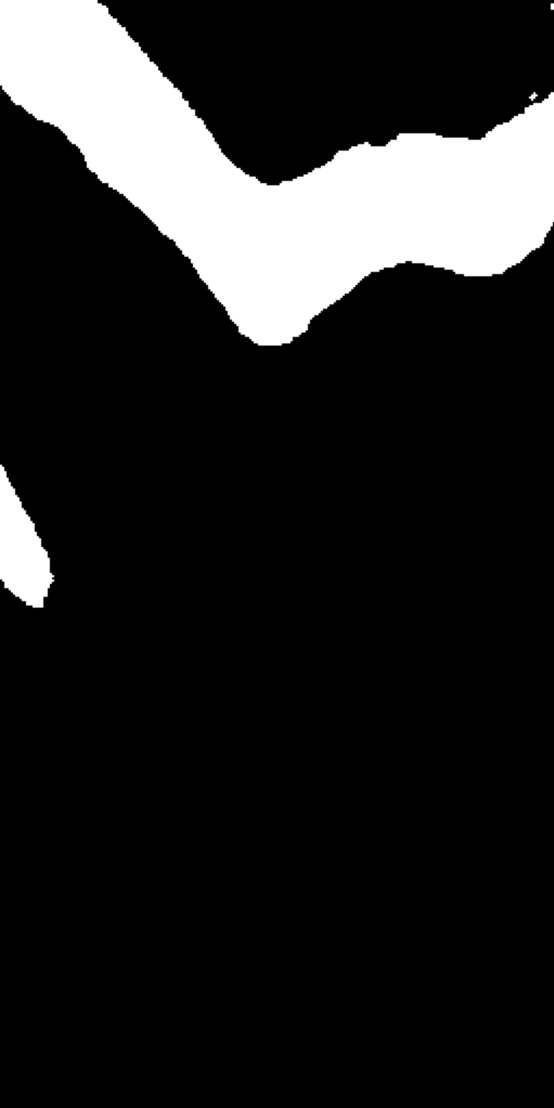} \\

        \raisebox{2cm}{(c)} &
        \includegraphics[width=0.17\textwidth]{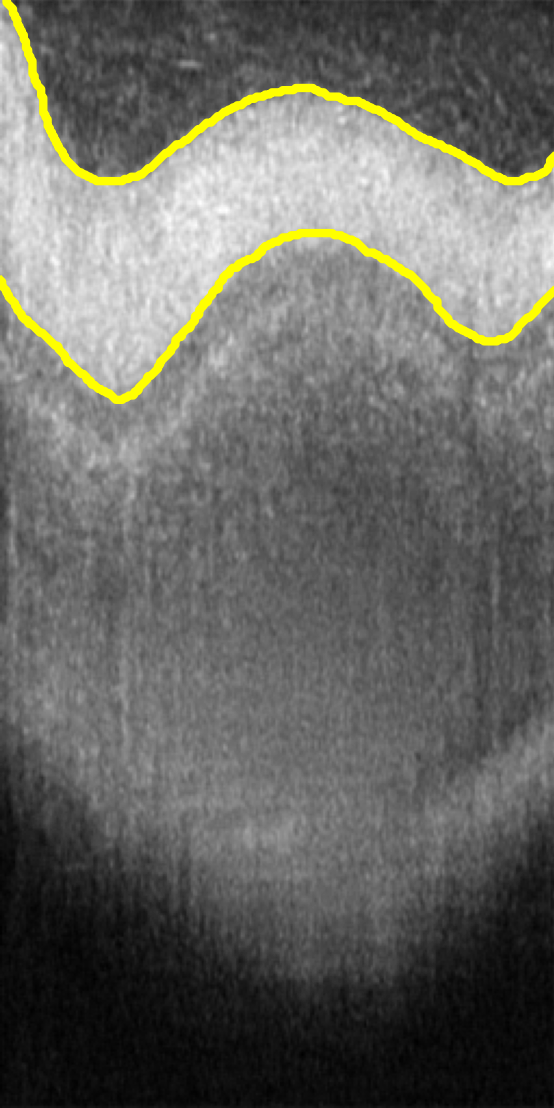} &
        \includegraphics[width=0.17\textwidth]{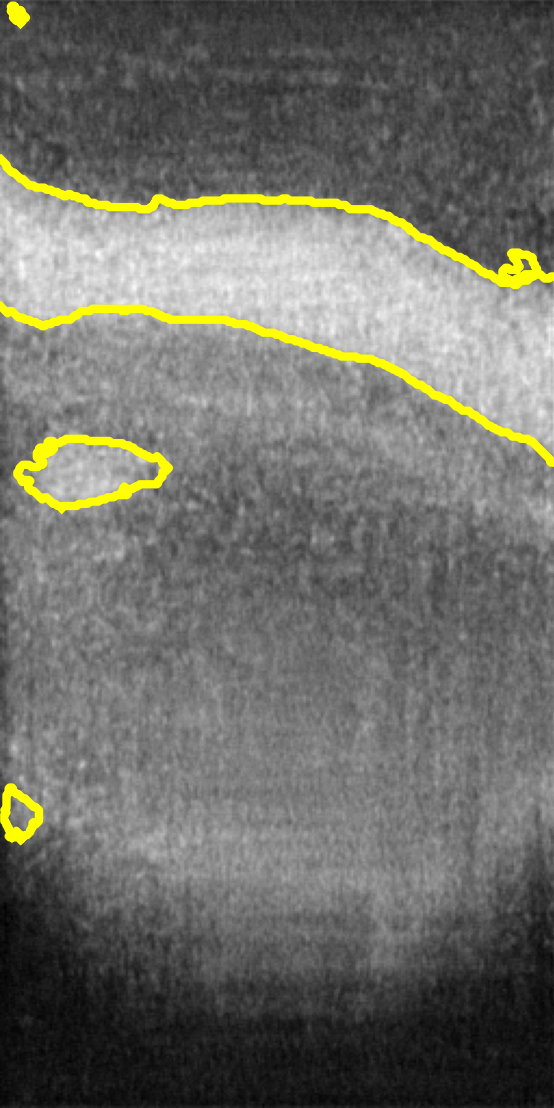} &
        \includegraphics[width=0.17\textwidth]{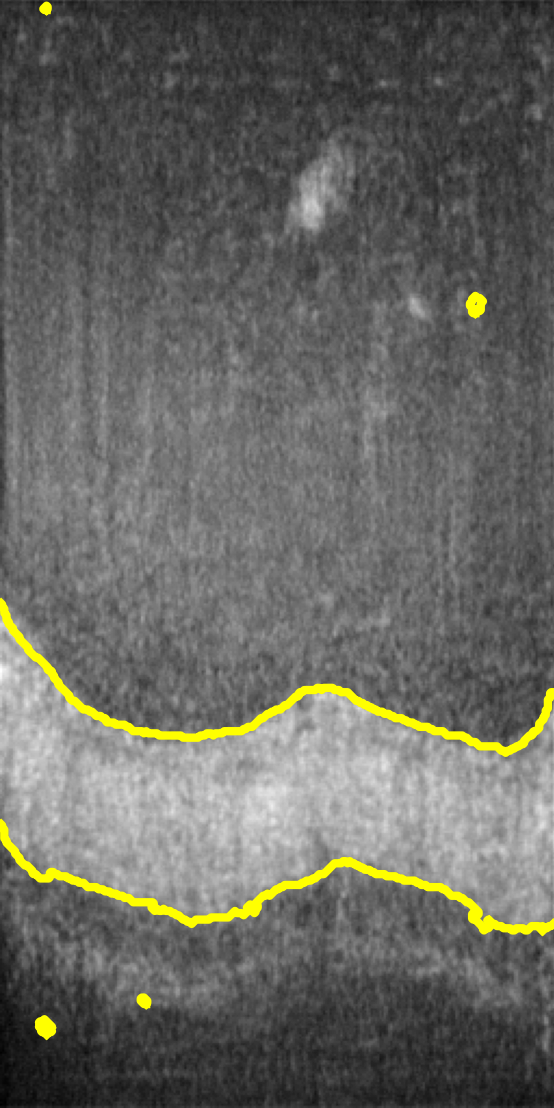} &
        \includegraphics[width=0.17\textwidth]{figs/appendix/DDPM/1/DDPM_overlay_5.png} &
        \includegraphics[width=0.17\textwidth]{figs/appendix/DDPM/1/DDPM_overlay_7.png} \\
    \end{tabular}

    \caption{Synthetic fluid-embedded images generated with DDPM \cite{ho_denoising_2020}. (a) Images; (b) masks; (c) overlaying the edges of the mask onto the image.}
    \label{fig:ddpm_1}
\end{figure*}

\begin{figure*}[H]
    \centering
    \setlength{\tabcolsep}{2pt} 
    \renewcommand{\arraystretch}{1.1} 

    \begin{tabular}{c c c c c c}
        \raisebox{2cm}{(a)} &
        \includegraphics[width=0.17\textwidth]{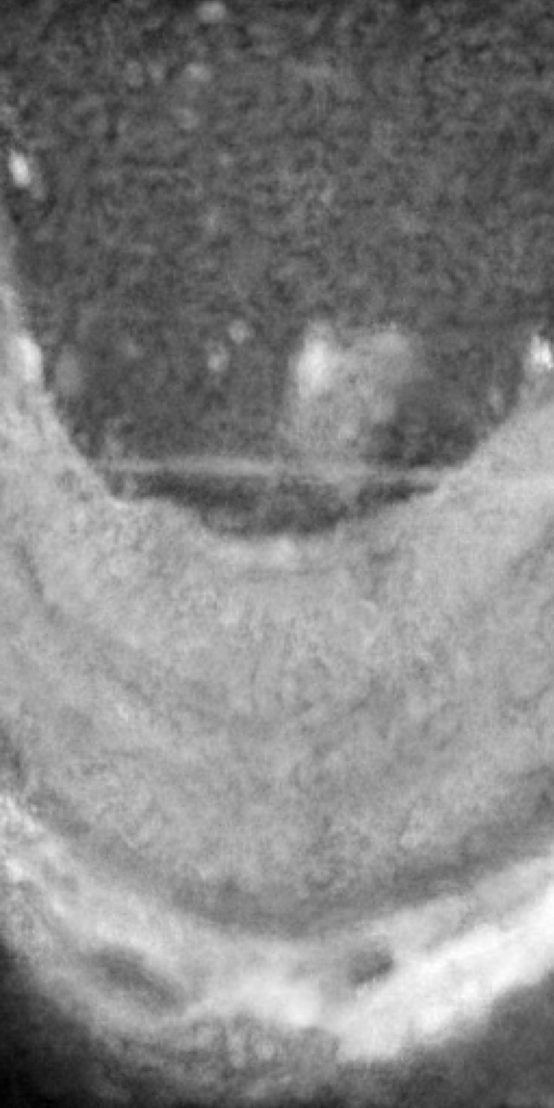} &
        \includegraphics[width=0.17\textwidth]{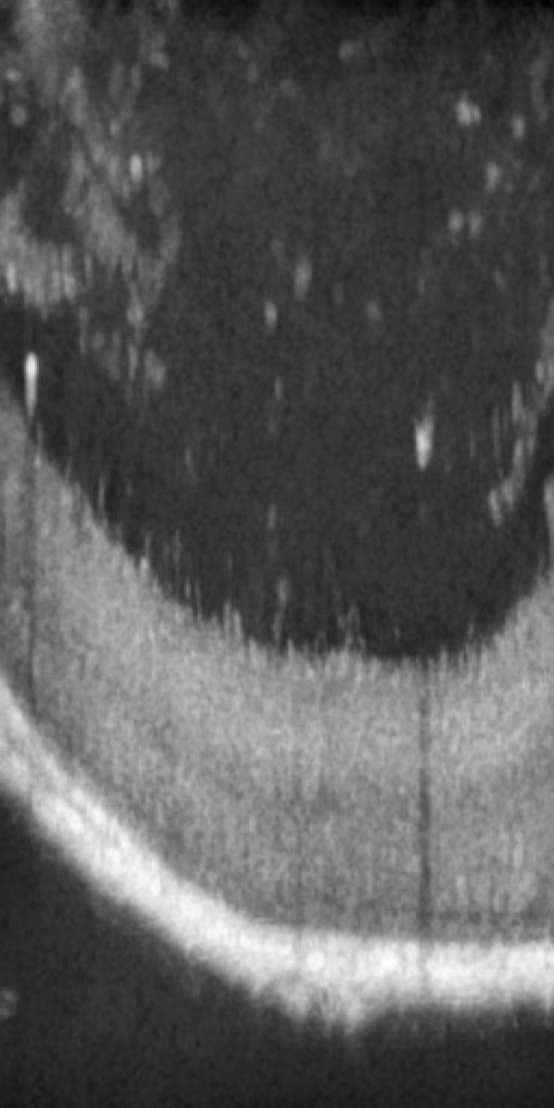} &
        \includegraphics[width=0.17\textwidth]{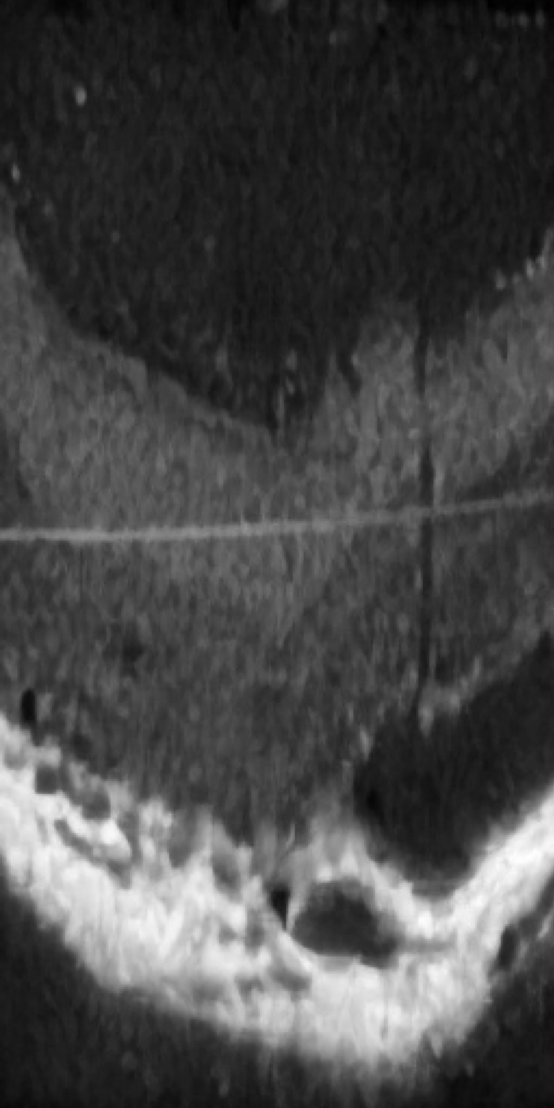} &
        \includegraphics[width=0.17\textwidth]{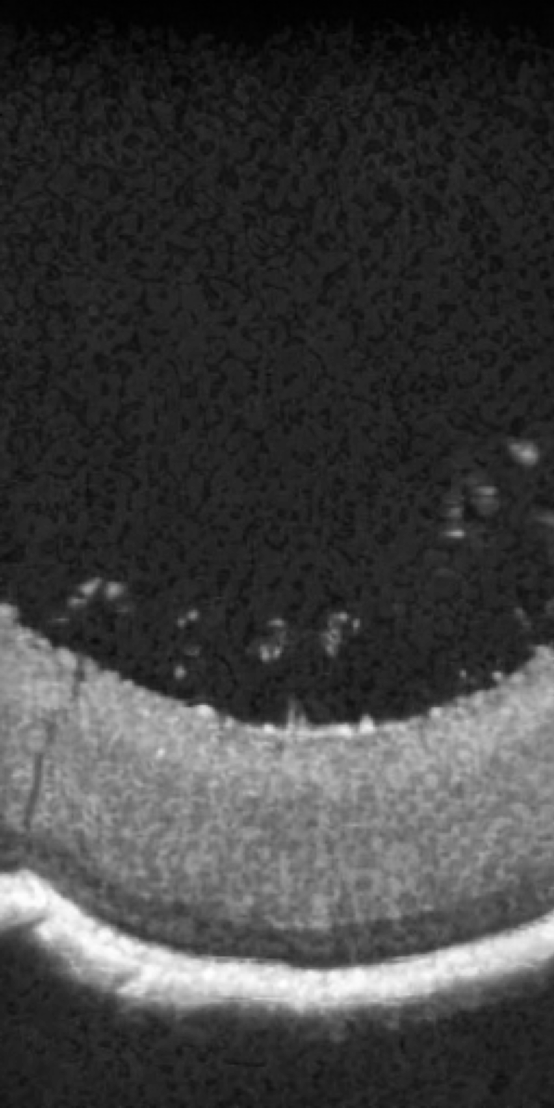} &
        \includegraphics[width=0.17\textwidth]{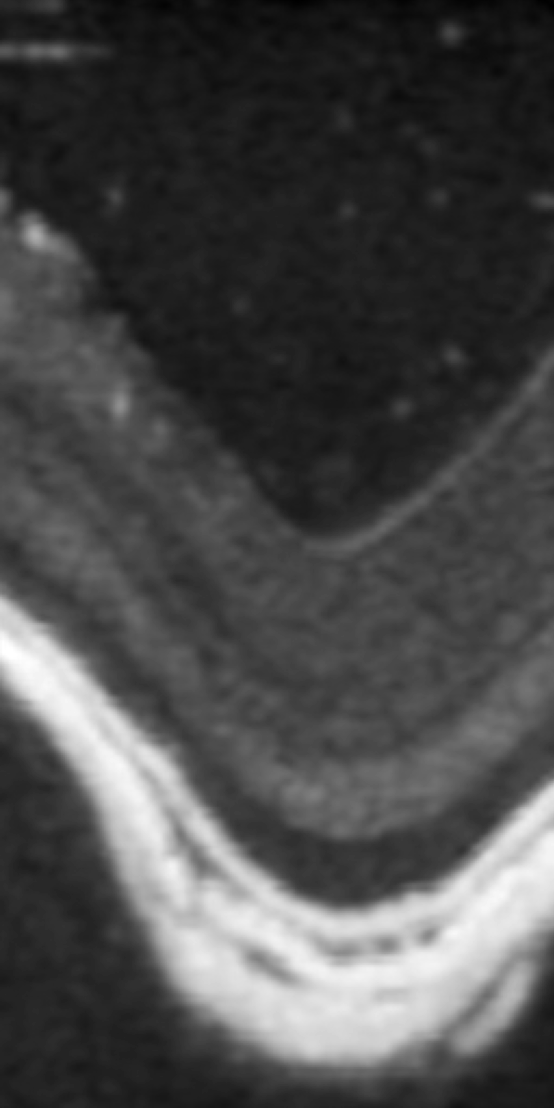} \\

        \raisebox{2cm}{(b)} &
        \includegraphics[width=0.17\textwidth]{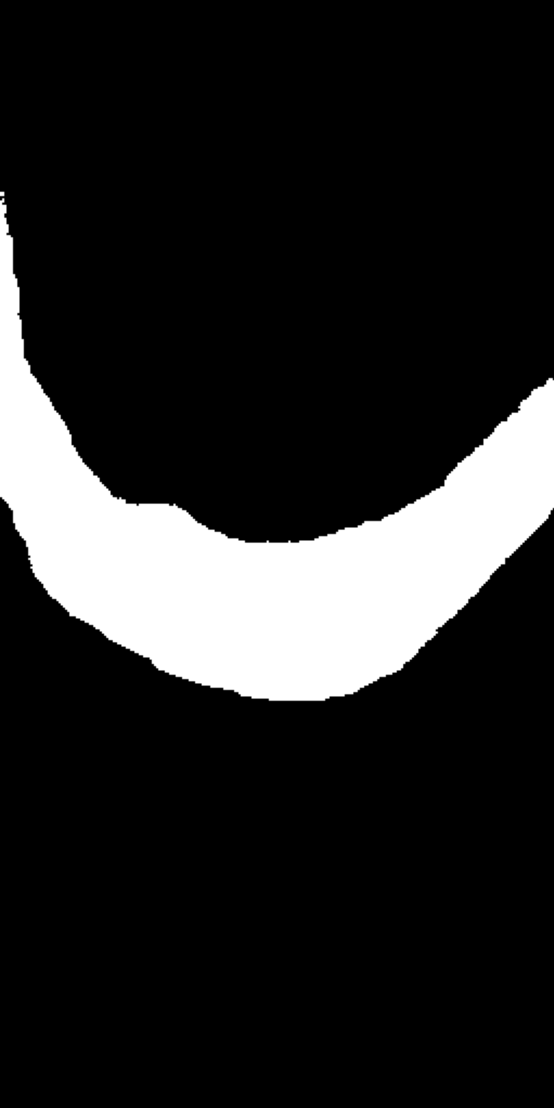} &
        \includegraphics[width=0.17\textwidth]{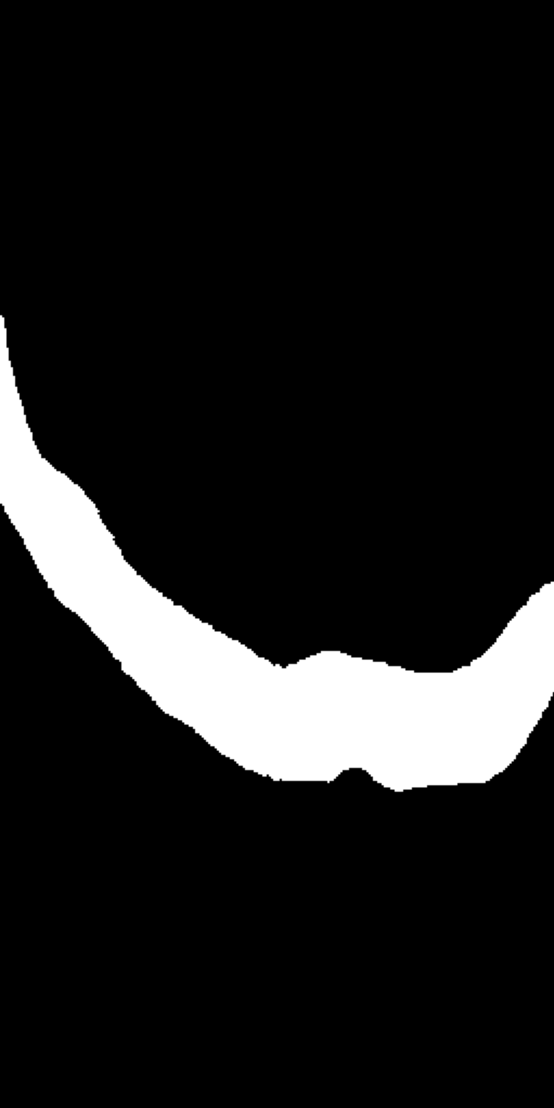} &
        \includegraphics[width=0.17\textwidth]{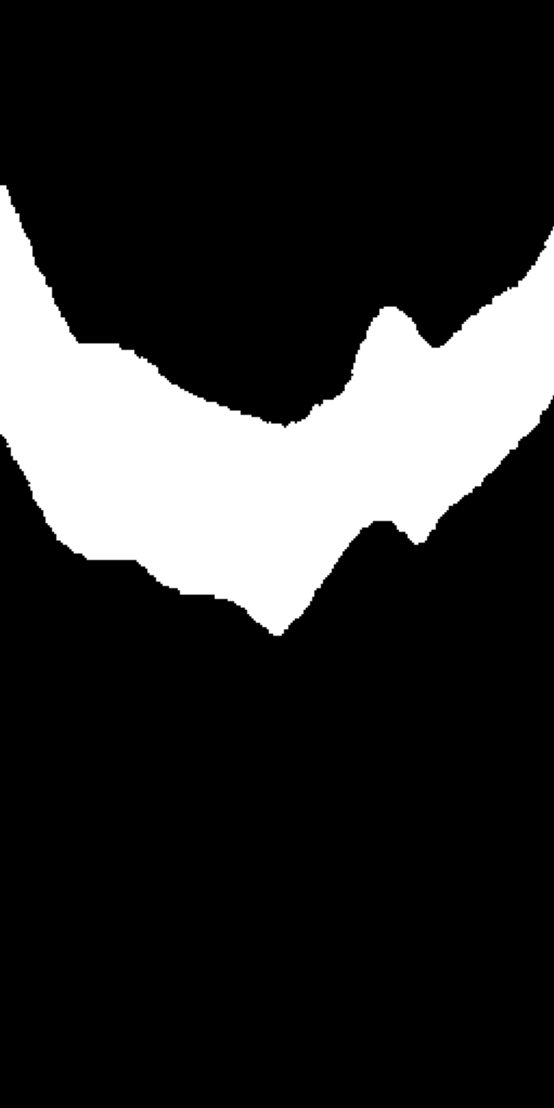} &
        \includegraphics[width=0.17\textwidth]{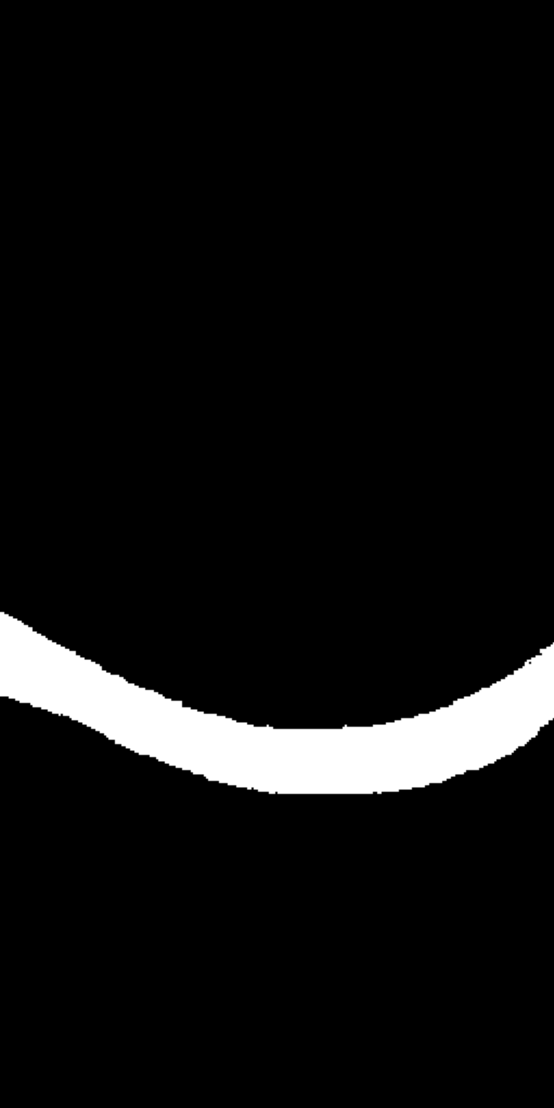} &
        \includegraphics[width=0.17\textwidth]{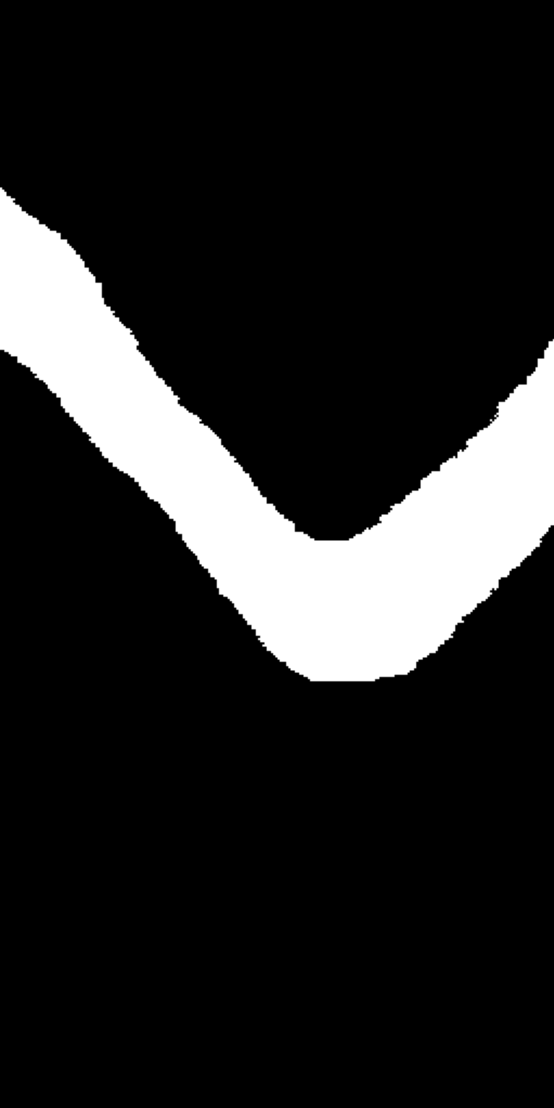} \\

        \raisebox{2cm}{(c)} &
        \includegraphics[width=0.17\textwidth]{figs/appendix/LDM/0/LDM_overlay_1.png} &
        \includegraphics[width=0.17\textwidth]{figs/appendix/LDM/0/LDM_overlay_3.png} &
        \includegraphics[width=0.17\textwidth]{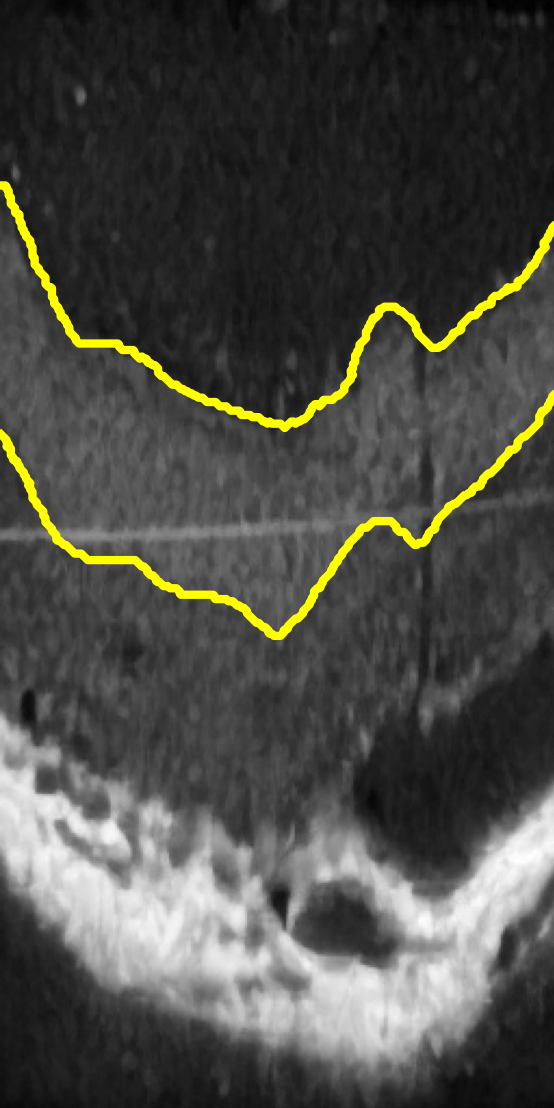} &
        \includegraphics[width=0.17\textwidth]{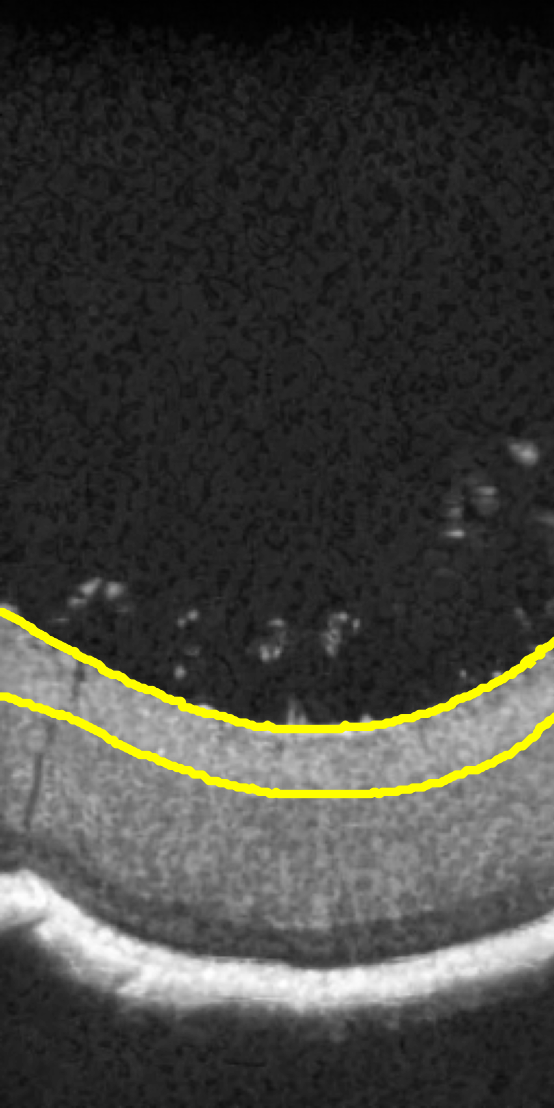} &
        \includegraphics[width=0.17\textwidth]{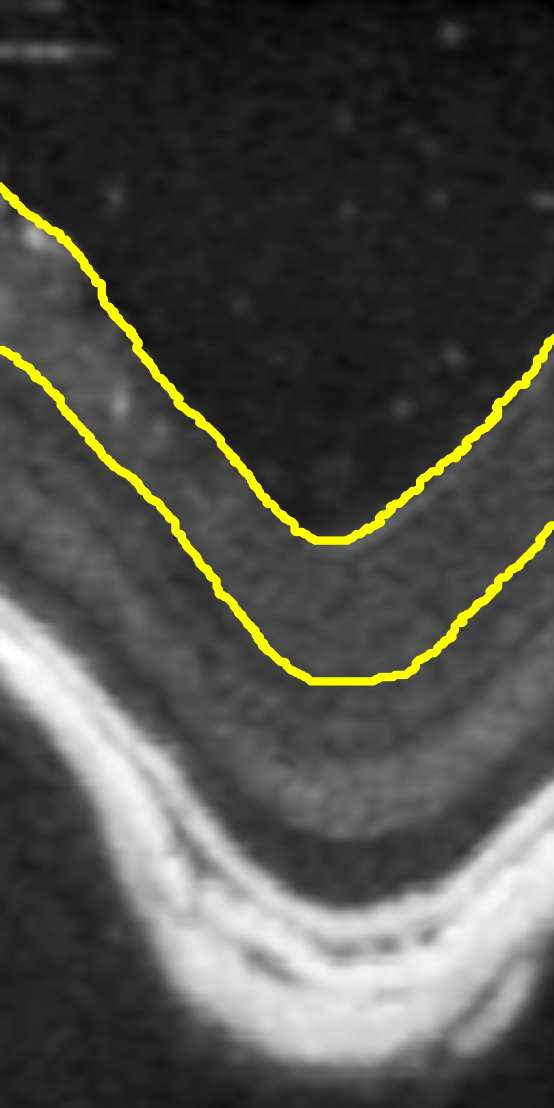} \\
    \end{tabular}

    \caption{Synthetic resin-embedded images generated with LDM \cite{rombach_high-resolution_2022-1}. (a) Images; (b) masks; (c) overlaying the edges of the mask onto the image.}
    \label{fig:ldm_0}
\end{figure*}

\begin{figure*}[H]
    \centering
    \setlength{\tabcolsep}{2pt} 
    \renewcommand{\arraystretch}{1.1} 

    \begin{tabular}{c c c c c c}
        \raisebox{2cm}{(a)} &
        \includegraphics[width=0.17\textwidth]{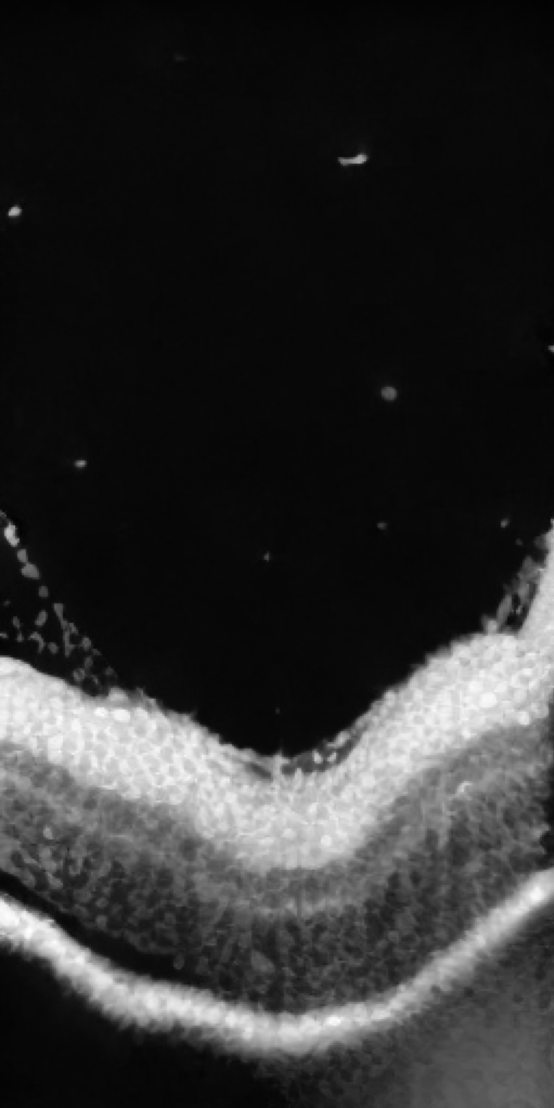} &
        \includegraphics[width=0.17\textwidth]{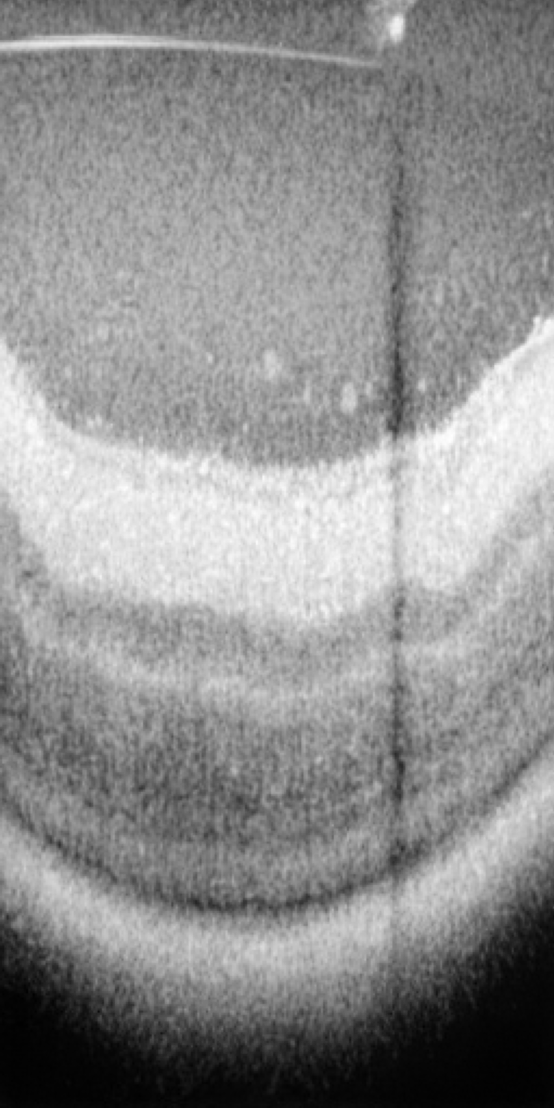} &
        \includegraphics[width=0.17\textwidth]{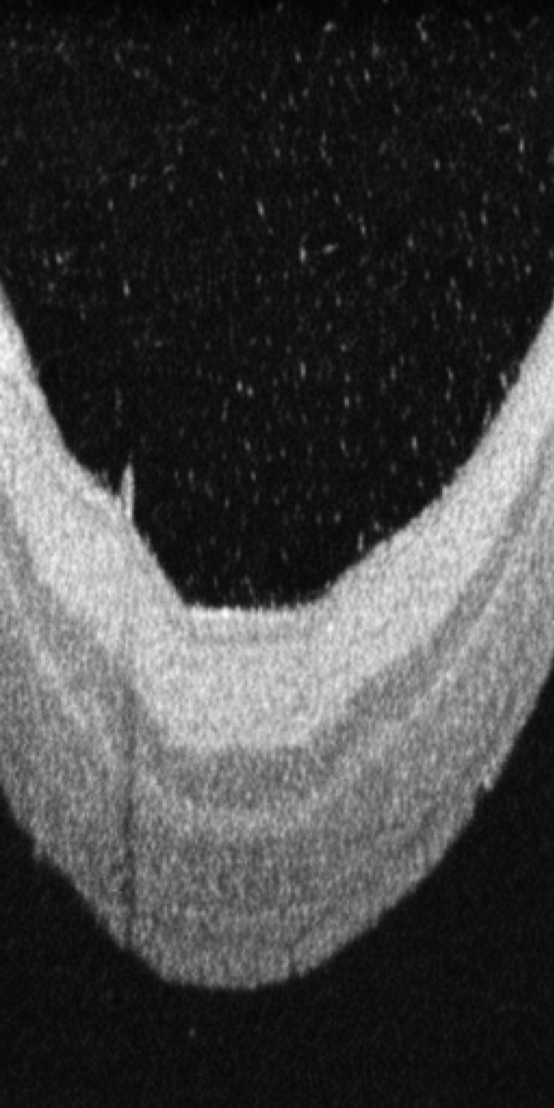} &
        \includegraphics[width=0.17\textwidth]{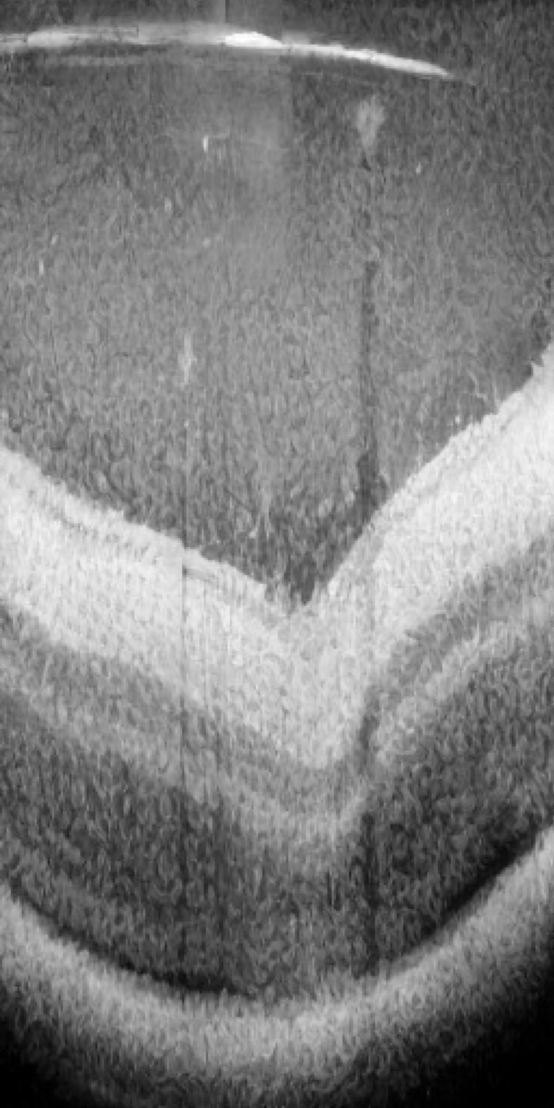} &
        \includegraphics[width=0.17\textwidth]{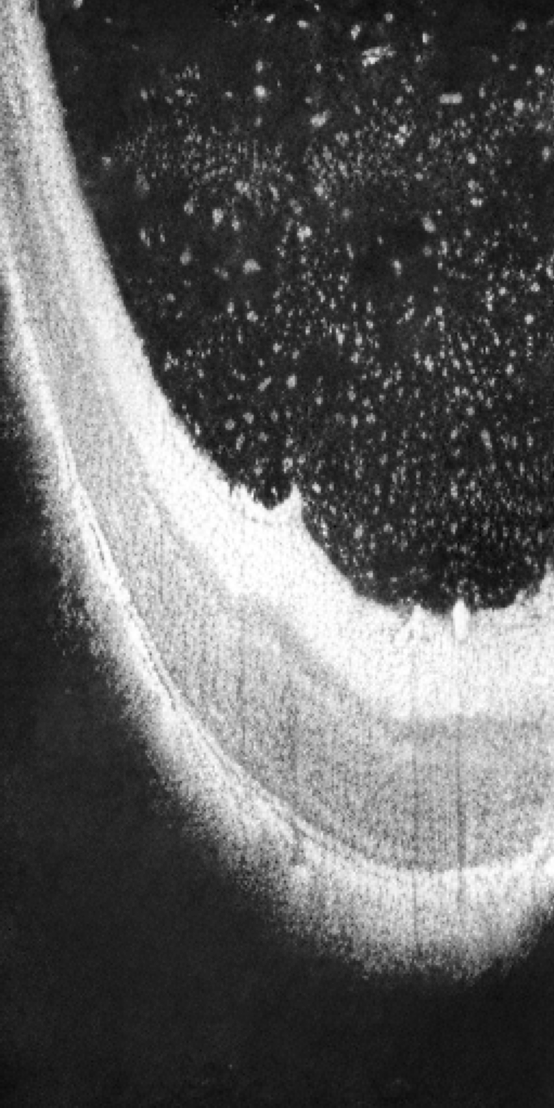} \\

        \raisebox{2cm}{(b)} &
        \includegraphics[width=0.17\textwidth]{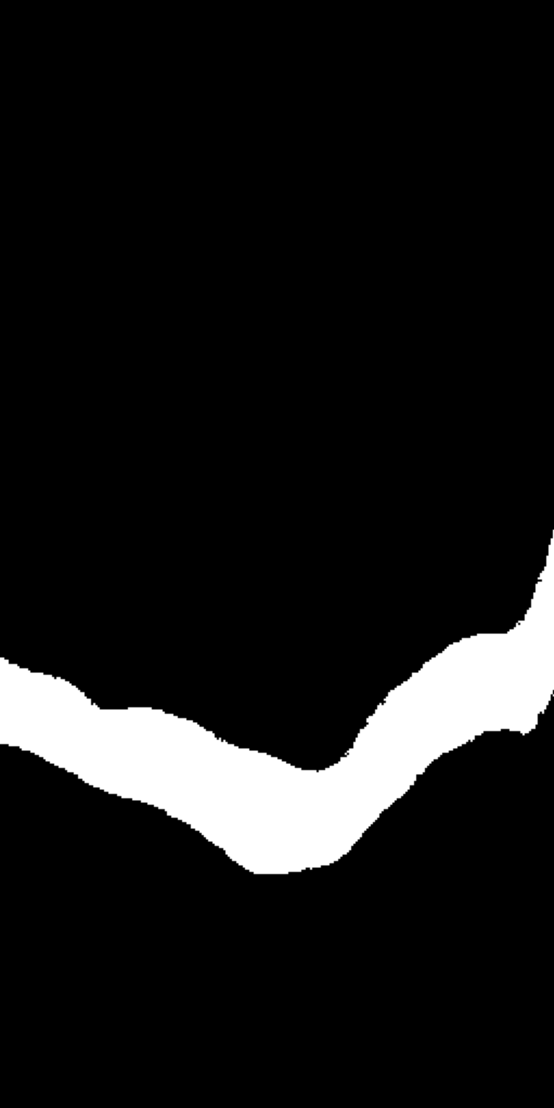} &
        \includegraphics[width=0.17\textwidth]{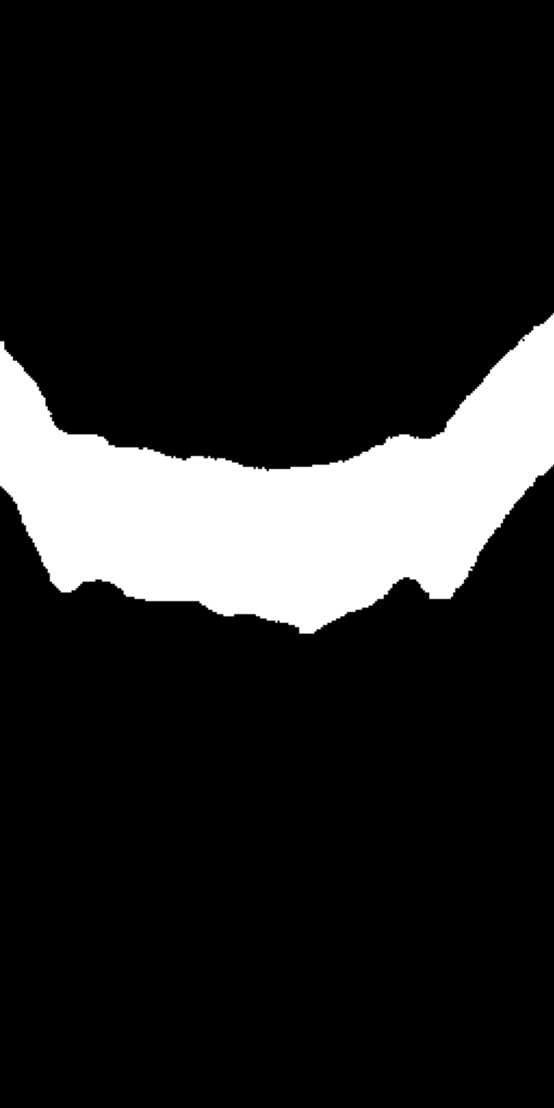} &
        \includegraphics[width=0.17\textwidth]{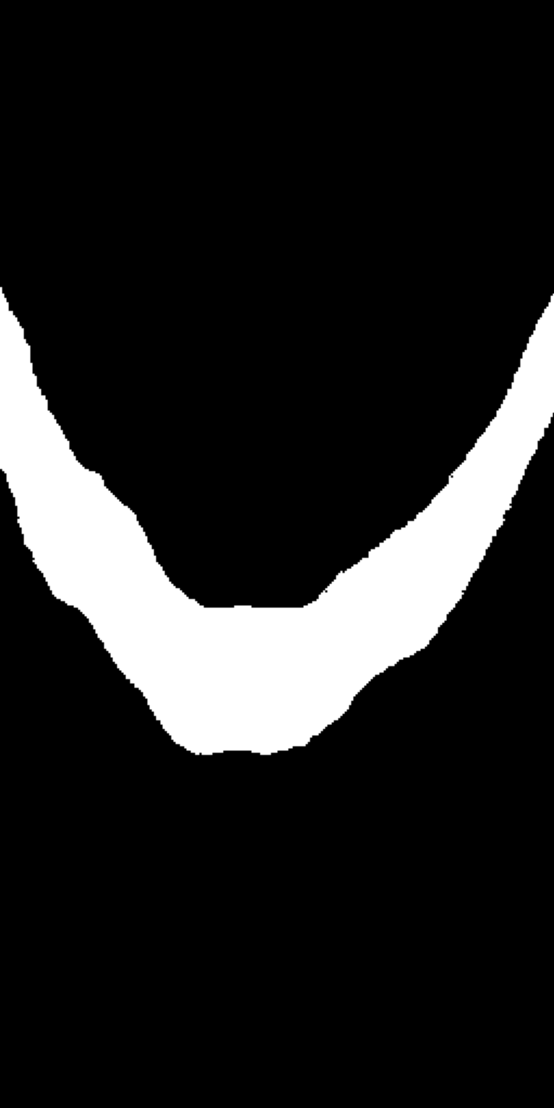} &
        \includegraphics[width=0.17\textwidth]{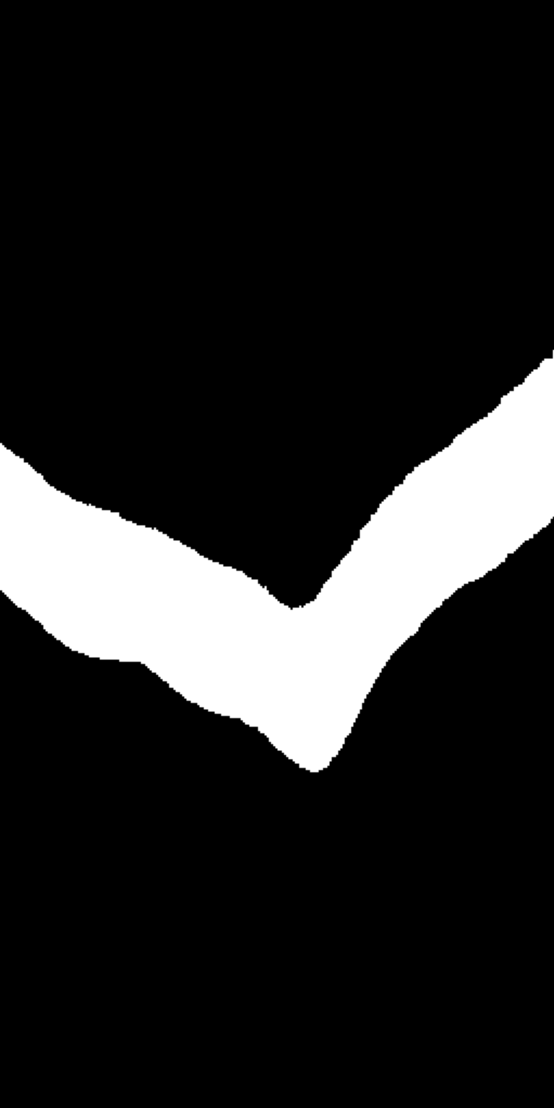} &
        \includegraphics[width=0.17\textwidth]{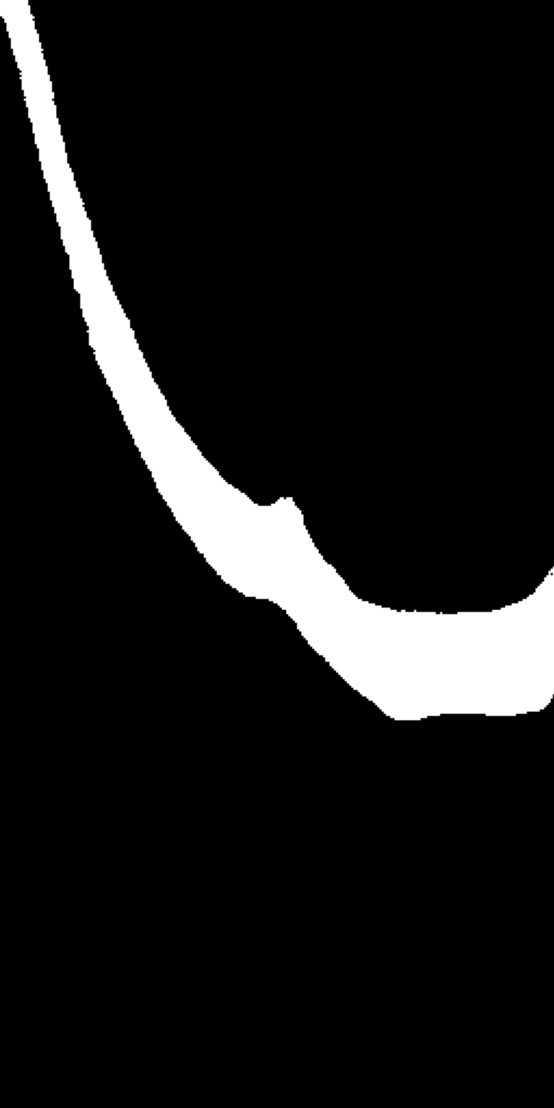} \\

        \raisebox{2cm}{(c)} &
        \includegraphics[width=0.17\textwidth]{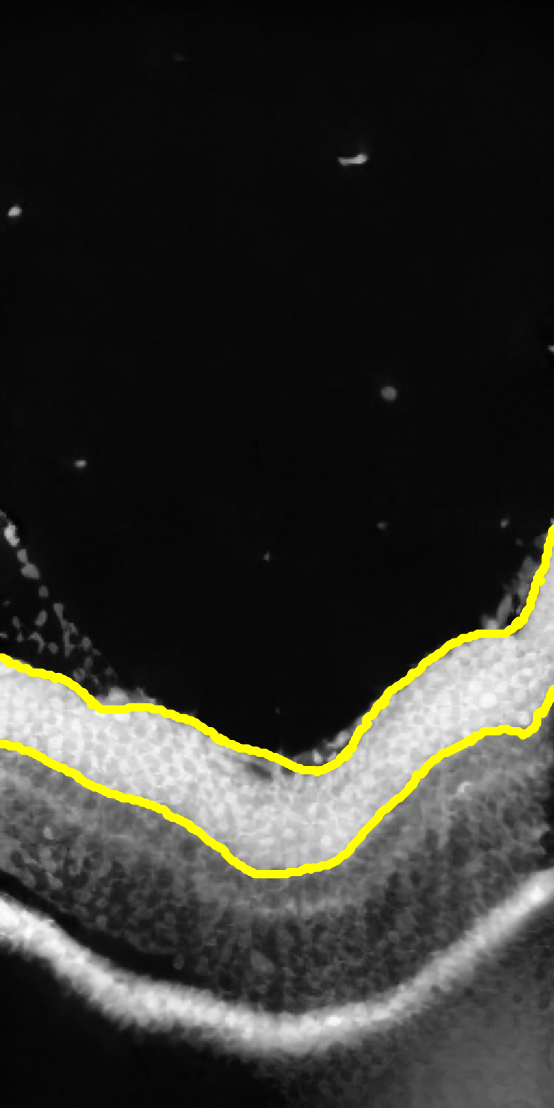} &
        \includegraphics[width=0.17\textwidth]{figs/appendix/LDM/1/LDM_overlay_3.png} &
        \includegraphics[width=0.17\textwidth]{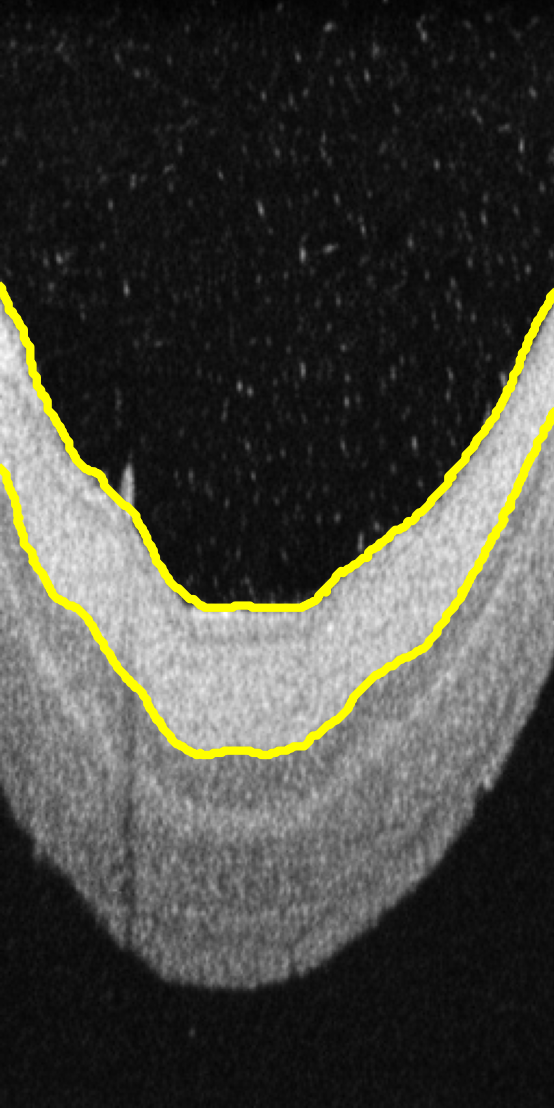} &
        \includegraphics[width=0.17\textwidth]{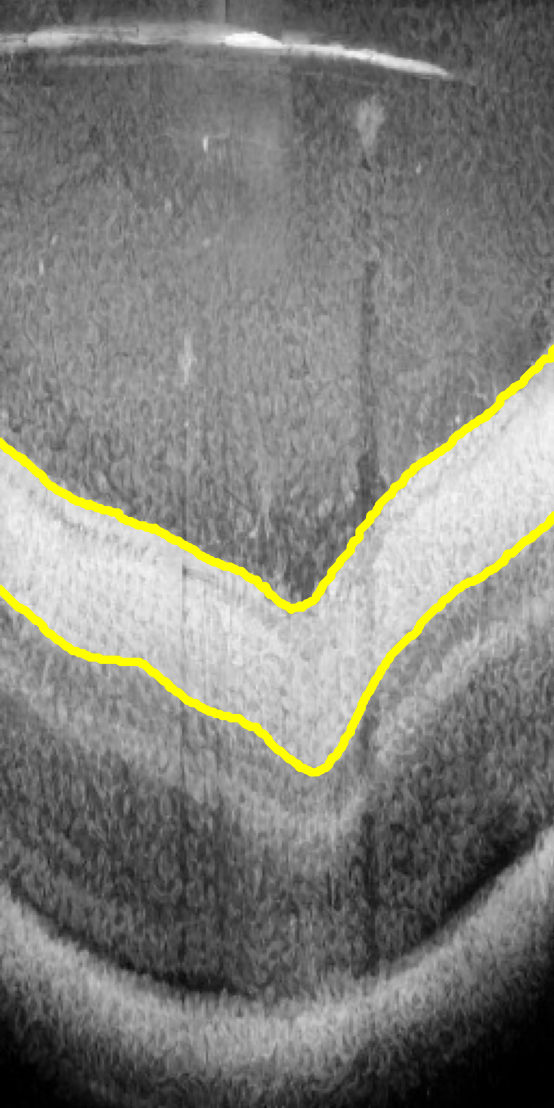} &
        \includegraphics[width=0.17\textwidth]{figs/appendix/LDM/1/LDM_overlay_6.png} \\
    \end{tabular}

    \caption{Synthetic fluid-embedded images generated with LDM \cite{rombach_high-resolution_2022-1}. (a) Images; (b) masks; (c) overlaying the edges of the mask onto the image.}
    \label{fig:ldm_1}
\end{figure*}

\begin{figure*}[H]
    \centering
    \setlength{\tabcolsep}{2pt} 
    \renewcommand{\arraystretch}{1.1} 

    \begin{tabular}{c c c c c c}
        \raisebox{2cm}{(a)} &
        \includegraphics[width=0.17\textwidth]{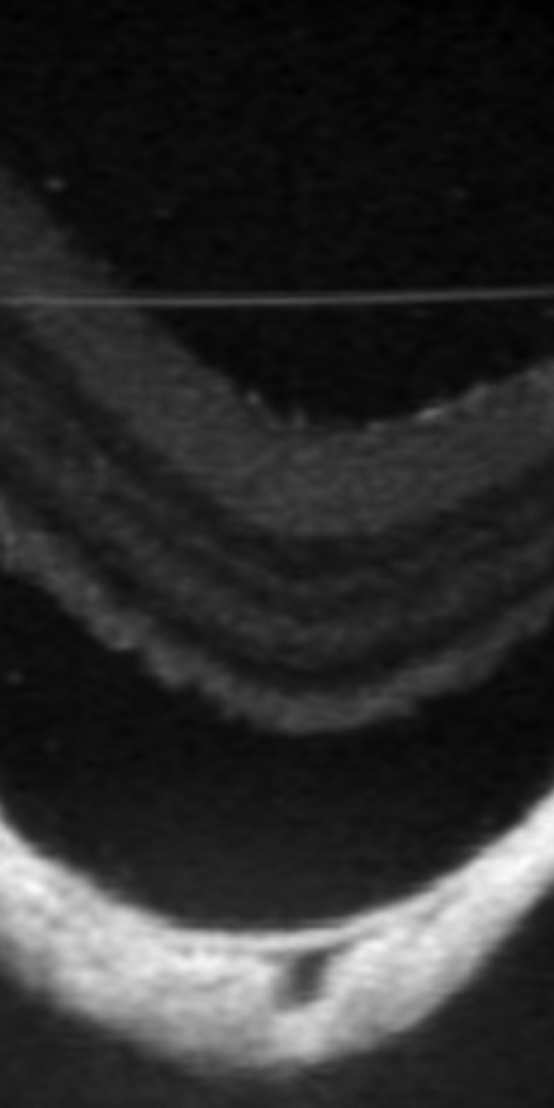} &
        \includegraphics[width=0.17\textwidth]{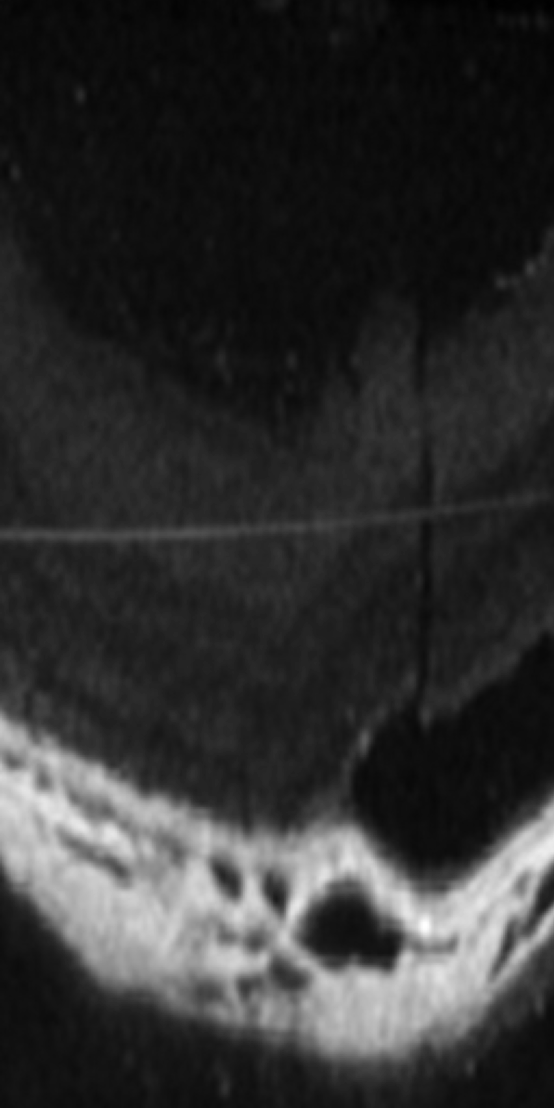} &
        \includegraphics[width=0.17\textwidth]{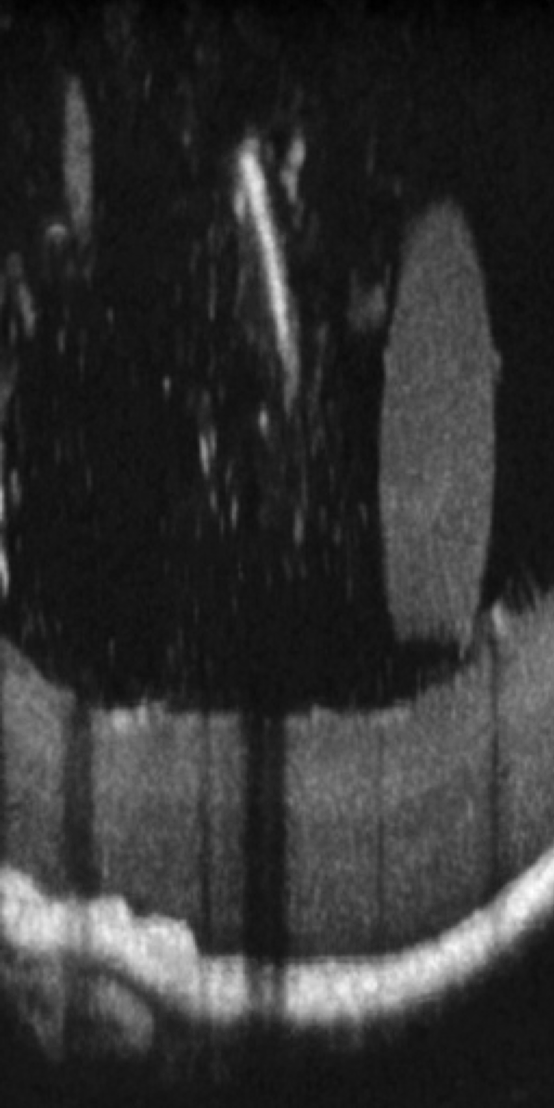} &
        \includegraphics[width=0.17\textwidth]{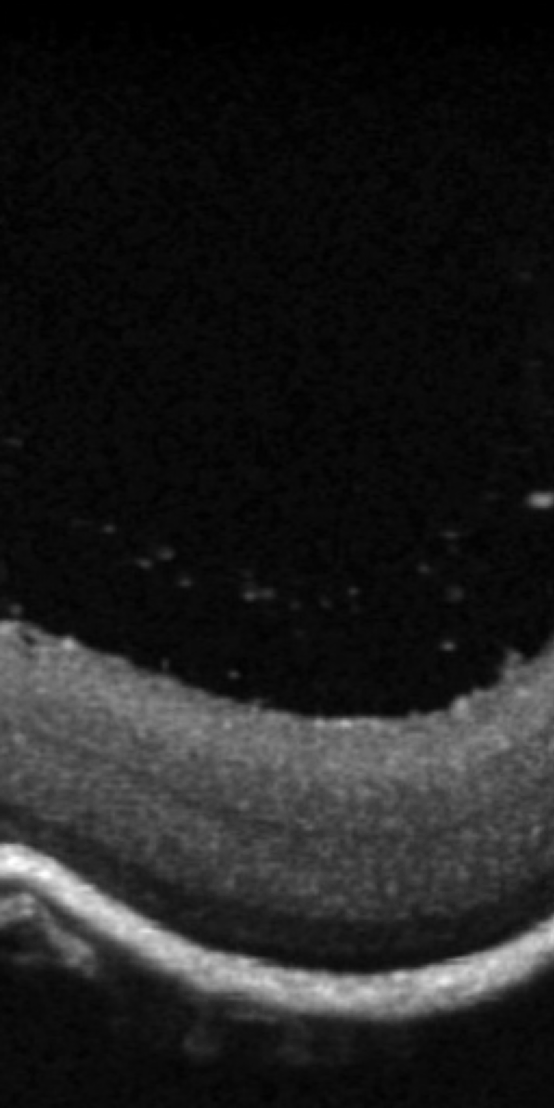} &
        \includegraphics[width=0.17\textwidth]{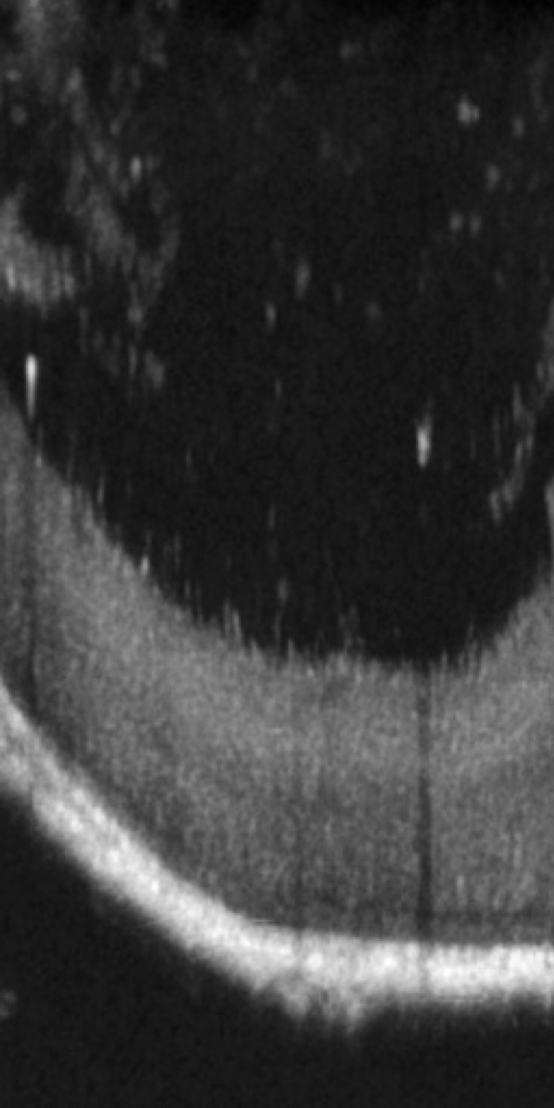} \\

        \raisebox{2cm}{(b)} &
        \includegraphics[width=0.17\textwidth]{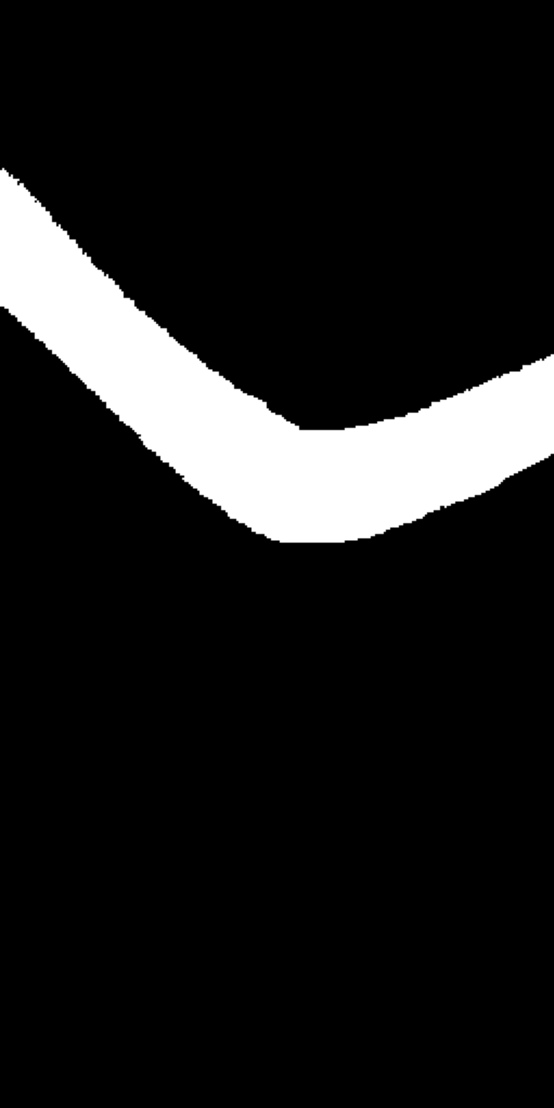} &
        \includegraphics[width=0.17\textwidth]{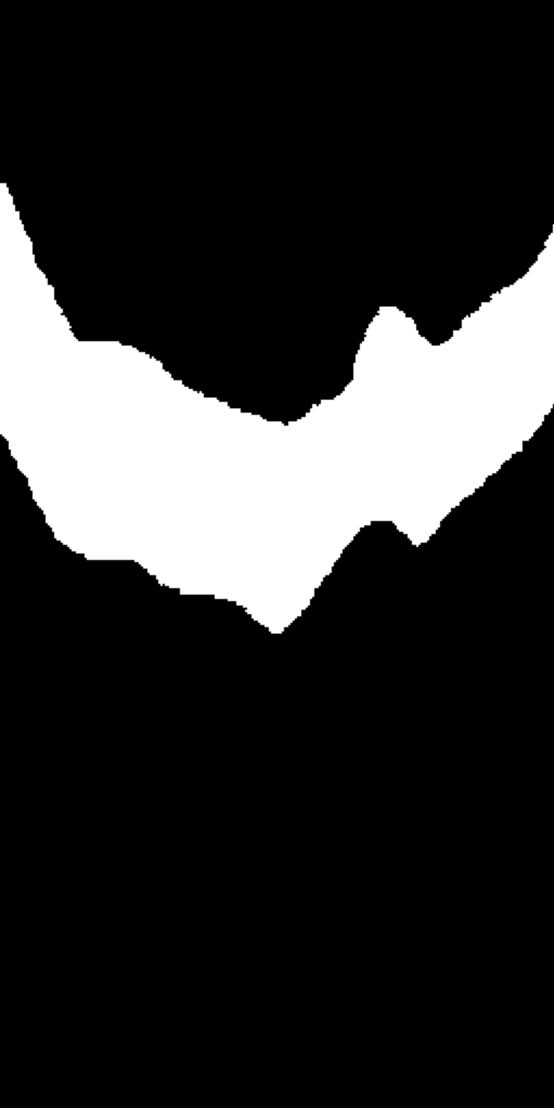} &
        \includegraphics[width=0.17\textwidth]{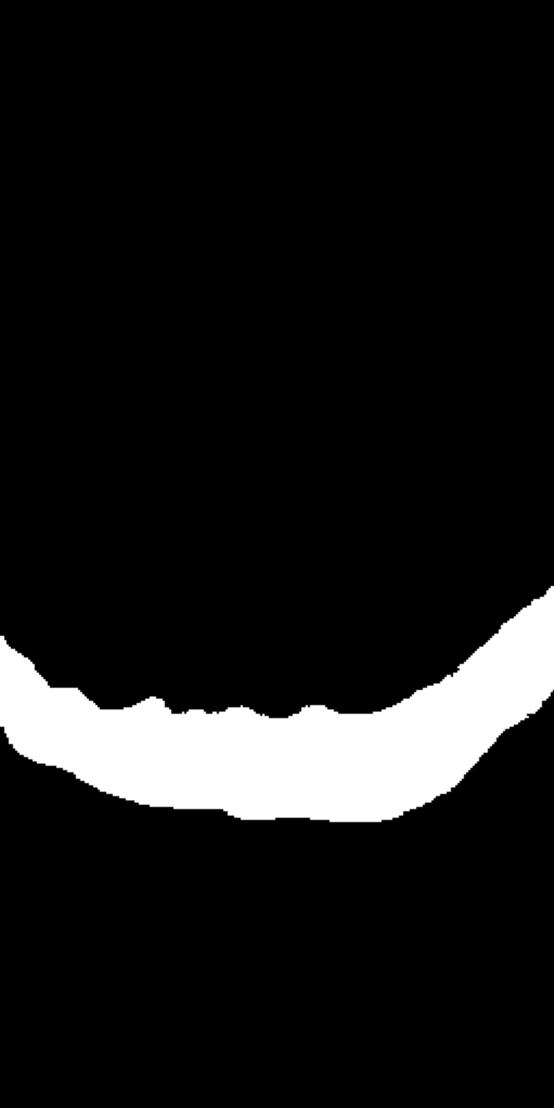} &
        \includegraphics[width=0.17\textwidth]{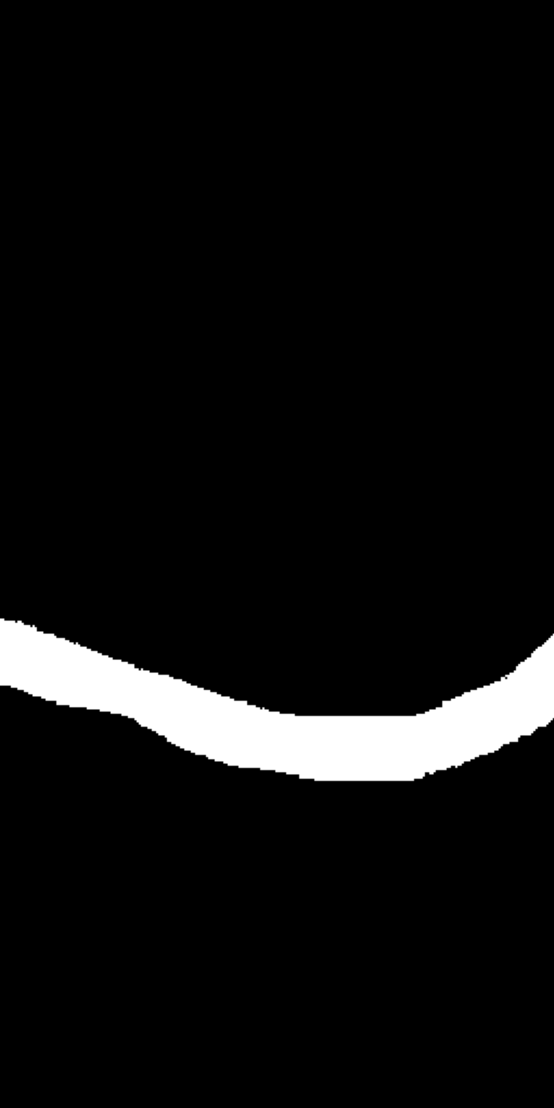} &
        \includegraphics[width=0.17\textwidth]{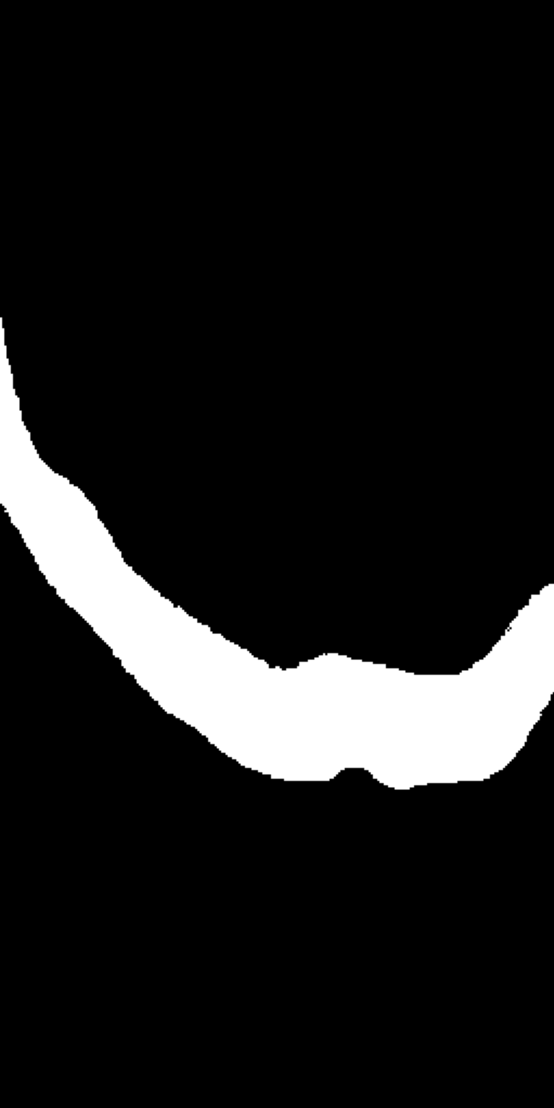} \\

        \raisebox{2cm}{(c)} &
        \includegraphics[width=0.17\textwidth]{figs/appendix/DiT/0/DiT_overlay_1.png} &
        \includegraphics[width=0.17\textwidth]{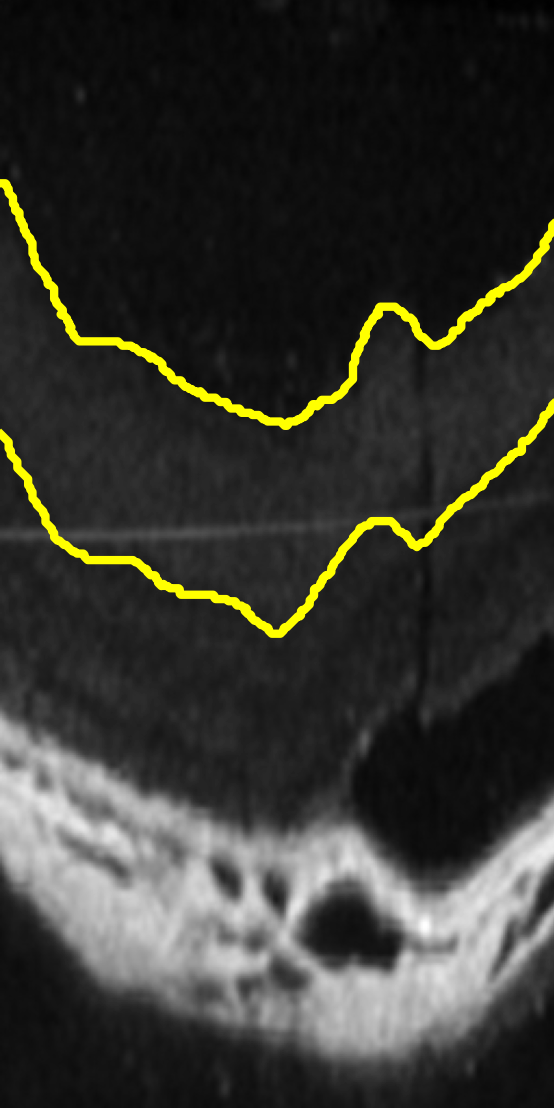} &
        \includegraphics[width=0.17\textwidth]{figs/appendix/DiT/0/DiT_overlay_4.png} &
        \includegraphics[width=0.17\textwidth]{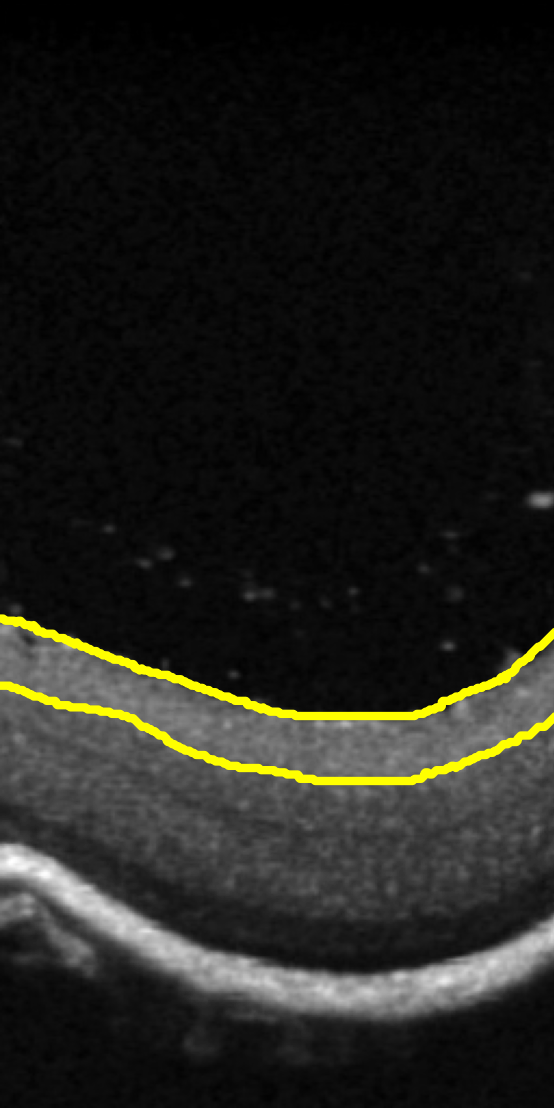} &
        \includegraphics[width=0.17\textwidth]{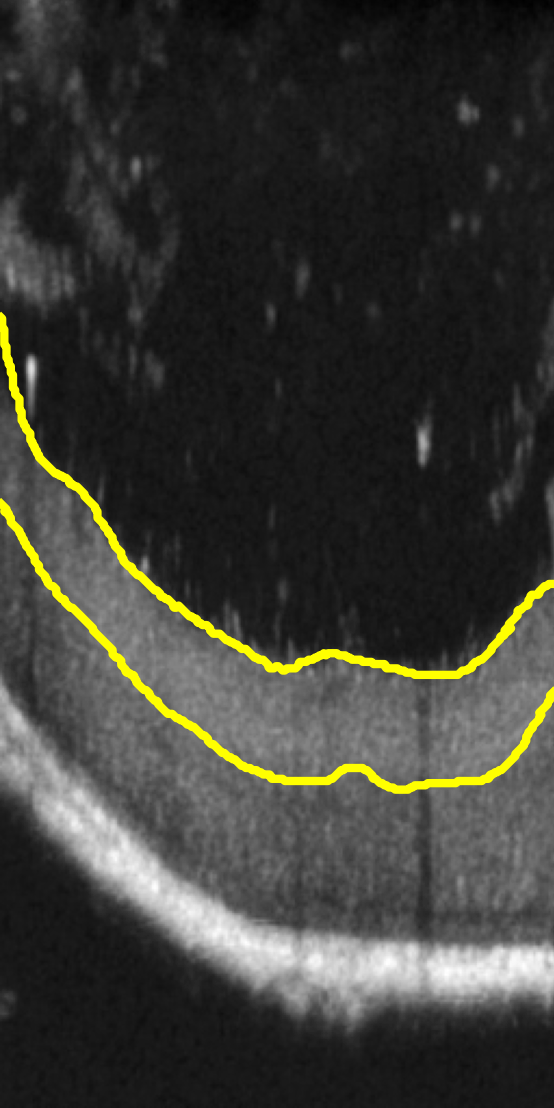} \\
    \end{tabular}

    \caption{Synthetic resin-embedded images generated with DualDiT. (a) Images; (b) masks; (c) overlaying the edges of the mask onto the image.}
    \label{fig:dit_0}
\end{figure*}

\begin{figure*}[H]
    \centering
    \setlength{\tabcolsep}{2pt} 
    \renewcommand{\arraystretch}{1.1} 

    \begin{tabular}{c c c c c c}
        \raisebox{2cm}{(a)} &
        \includegraphics[width=0.17\textwidth]{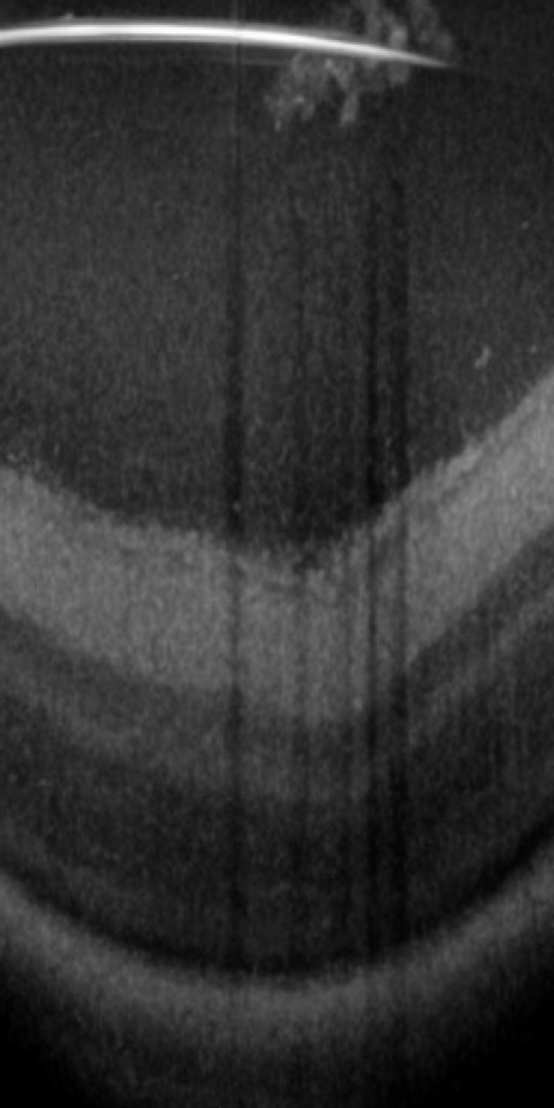} &
        \includegraphics[width=0.17\textwidth]{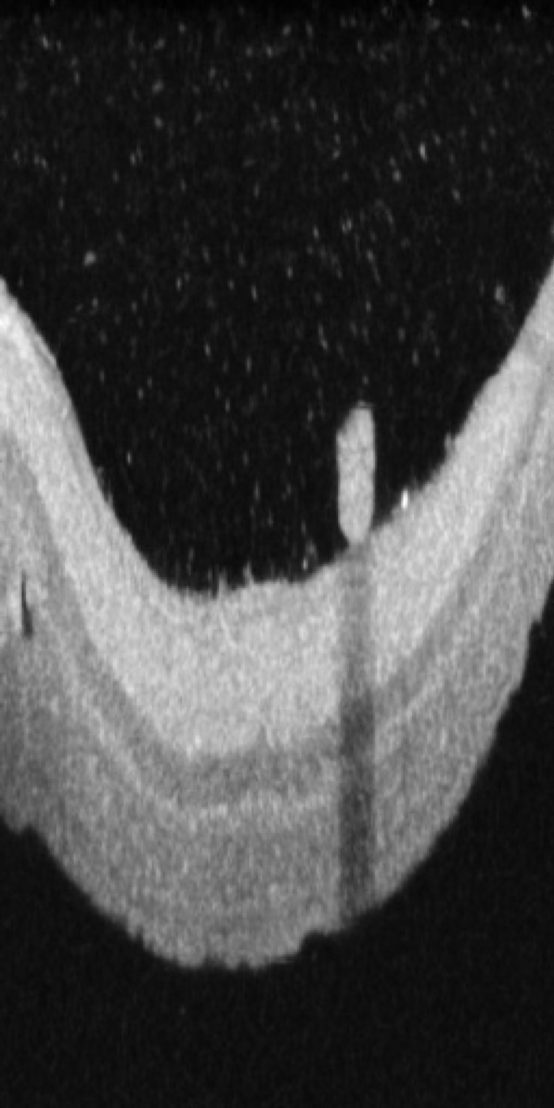} &
        \includegraphics[width=0.17\textwidth]{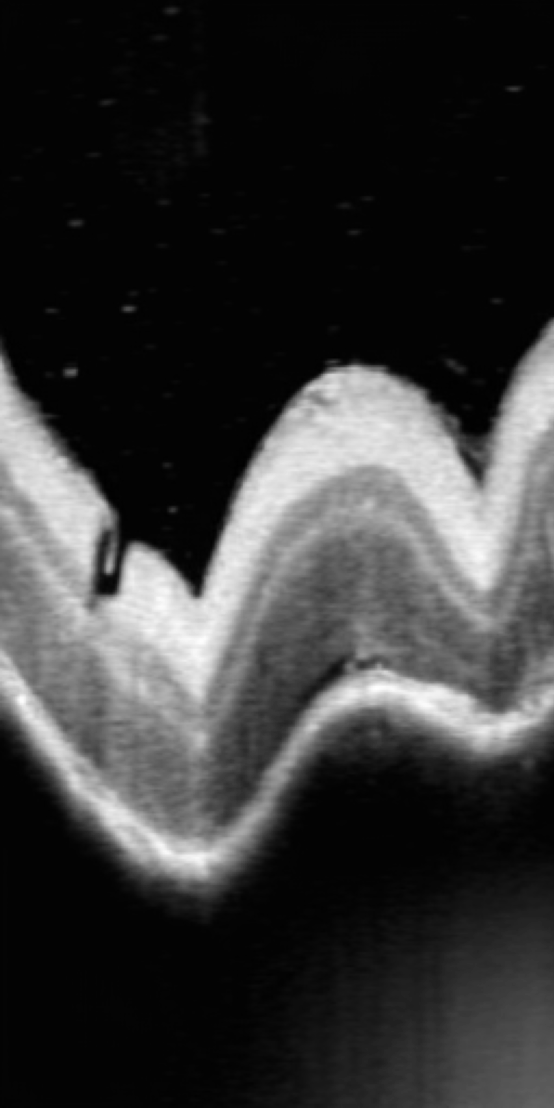} &
        \includegraphics[width=0.17\textwidth]{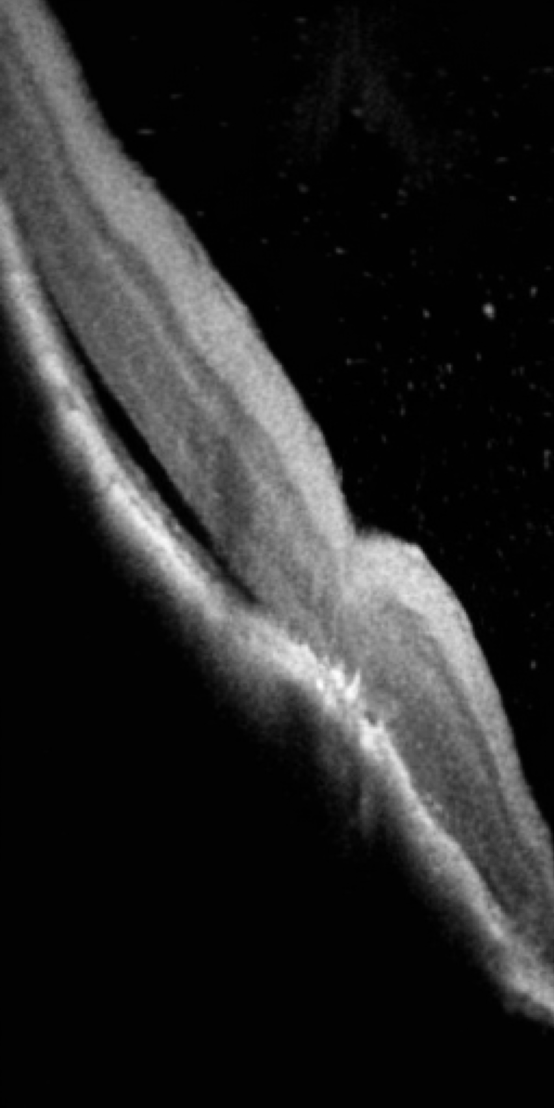} &
        \includegraphics[width=0.17\textwidth]{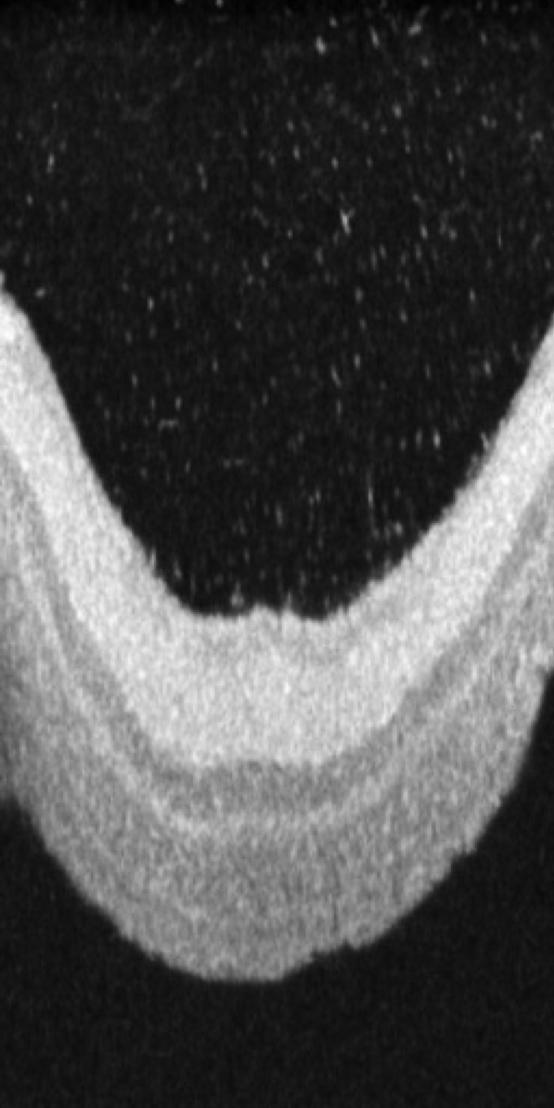} \\

        \raisebox{2cm}{(b)} &
        \includegraphics[width=0.17\textwidth]{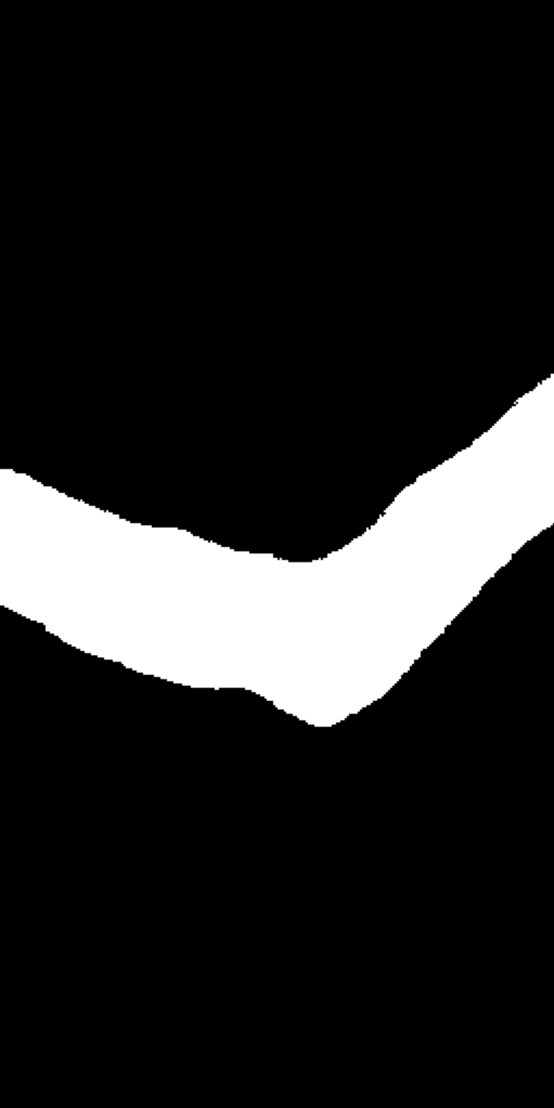} &
        \includegraphics[width=0.17\textwidth]{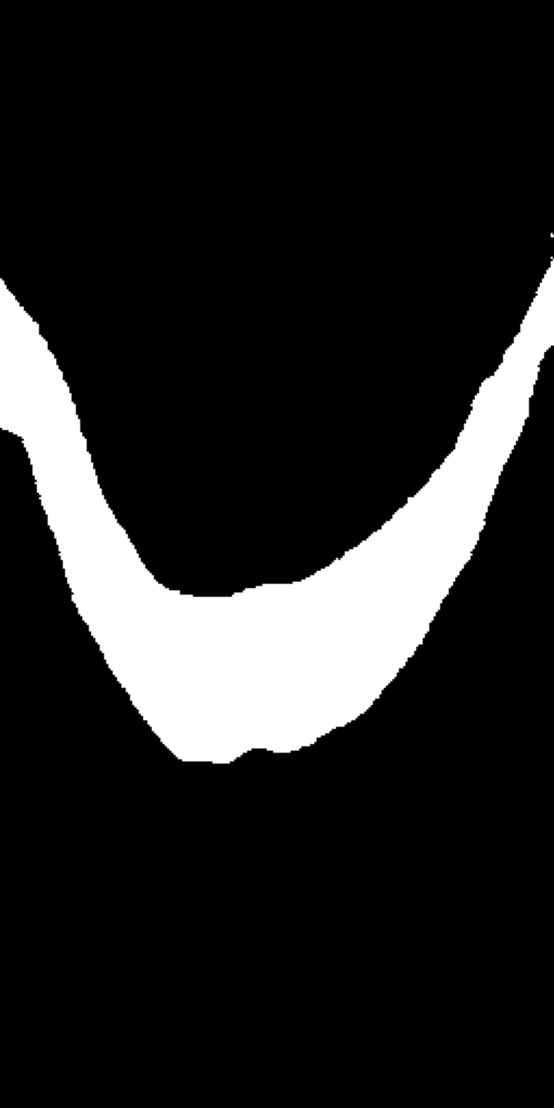} &
        \includegraphics[width=0.17\textwidth]{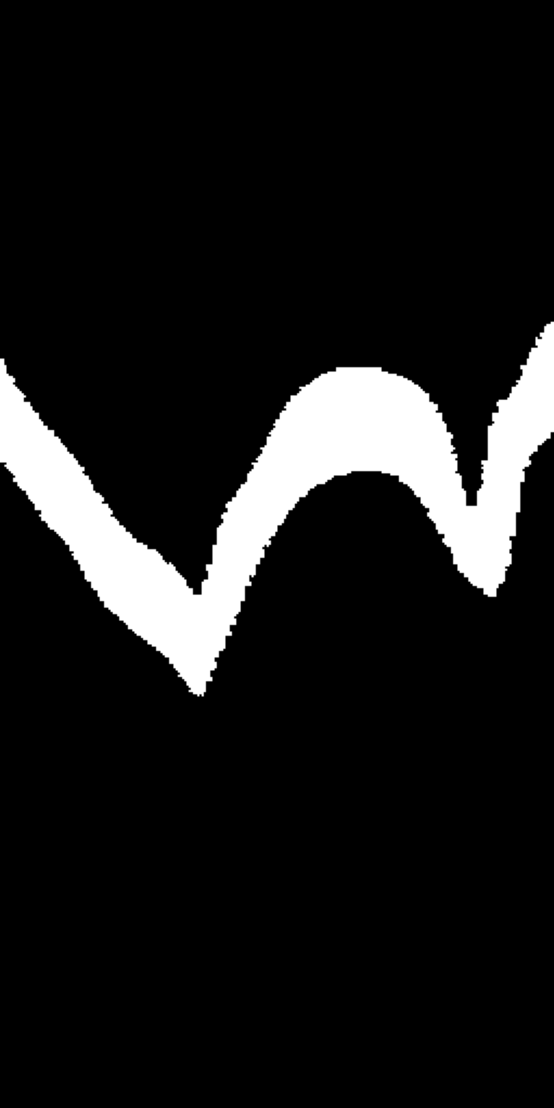} &
        \includegraphics[width=0.17\textwidth]{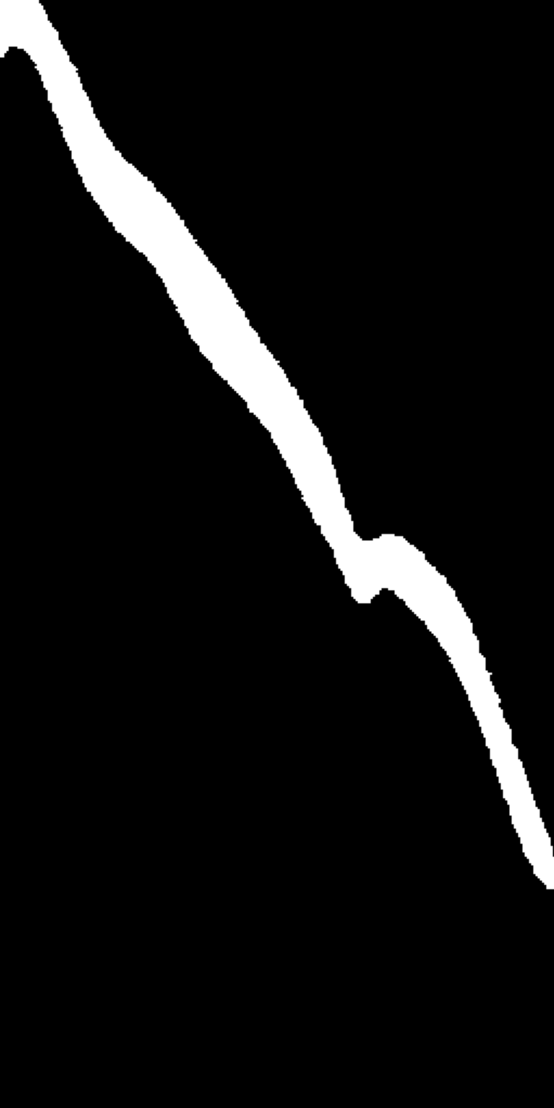} &
        \includegraphics[width=0.17\textwidth]{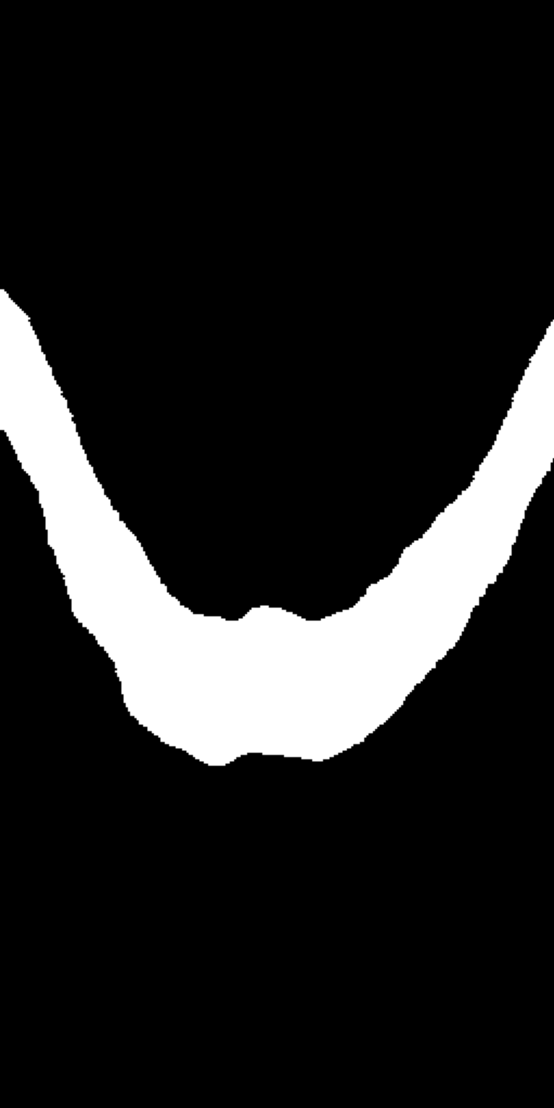} \\

        \raisebox{2cm}{(c)} &
        \includegraphics[width=0.17\textwidth]{figs/appendix/DiT/1/DiT_overlay_1.png} &
        \includegraphics[width=0.17\textwidth]{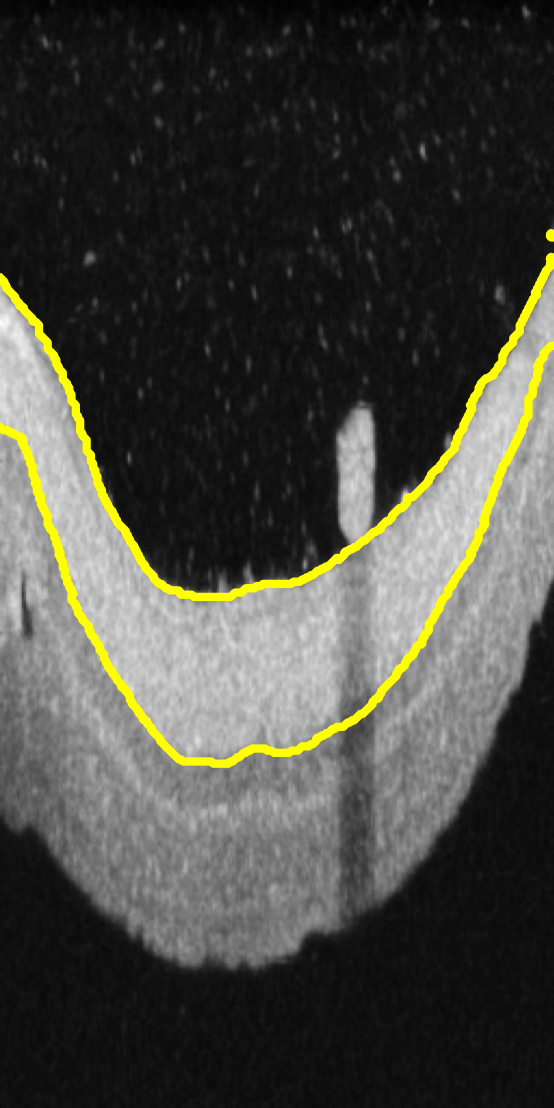} &
        \includegraphics[width=0.17\textwidth]{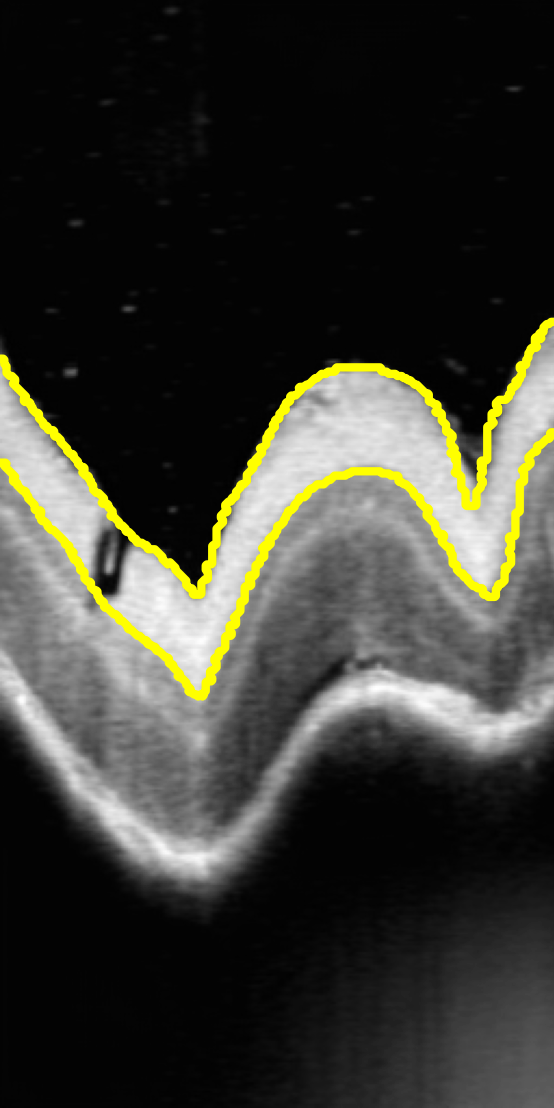} &
        \includegraphics[width=0.17\textwidth]{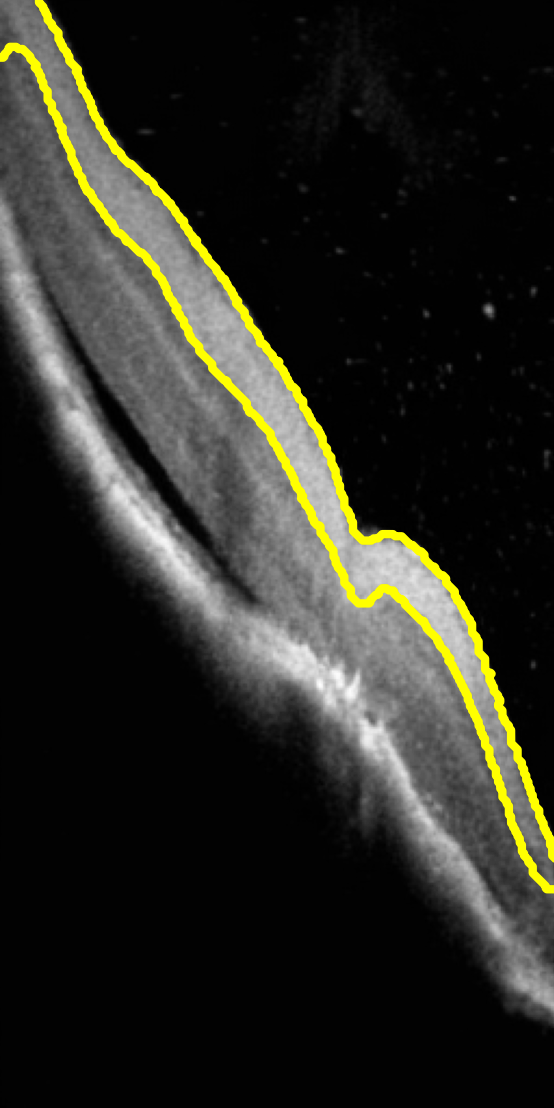} &
        \includegraphics[width=0.17\textwidth]{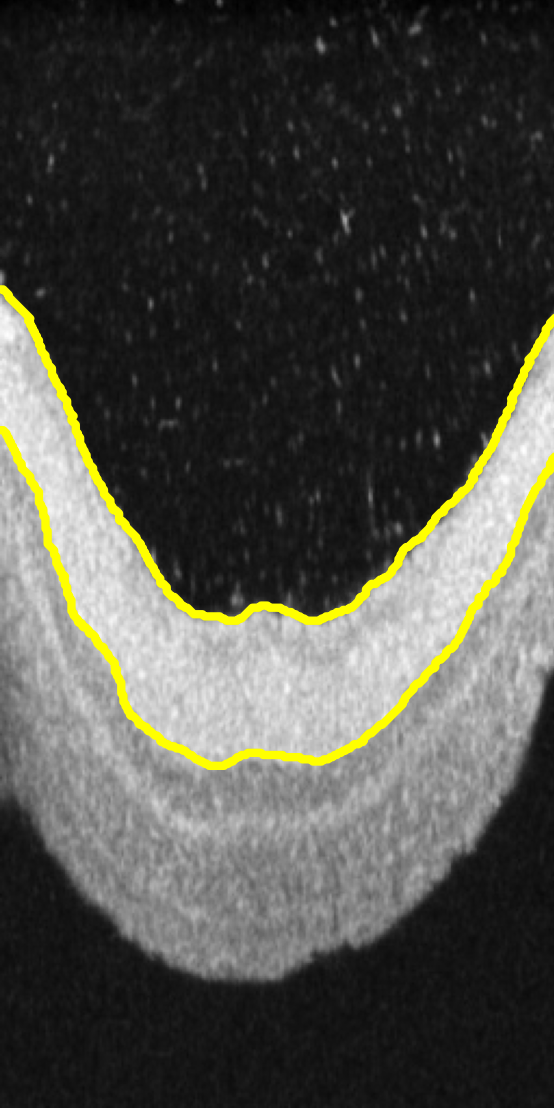} \\
    \end{tabular}

    \caption{Synthetic fluid-embedded images generated with DualDiT. (a) Images; (b) masks; (c) overlaying the edges of the mask onto the image.}
    \label{fig:dit_1}
\end{figure*}

\begin{figure*}[p]
    \centering

    \begin{subfigure}[t]{0.47\textwidth}
        \centering
        \includegraphics[width=\linewidth]{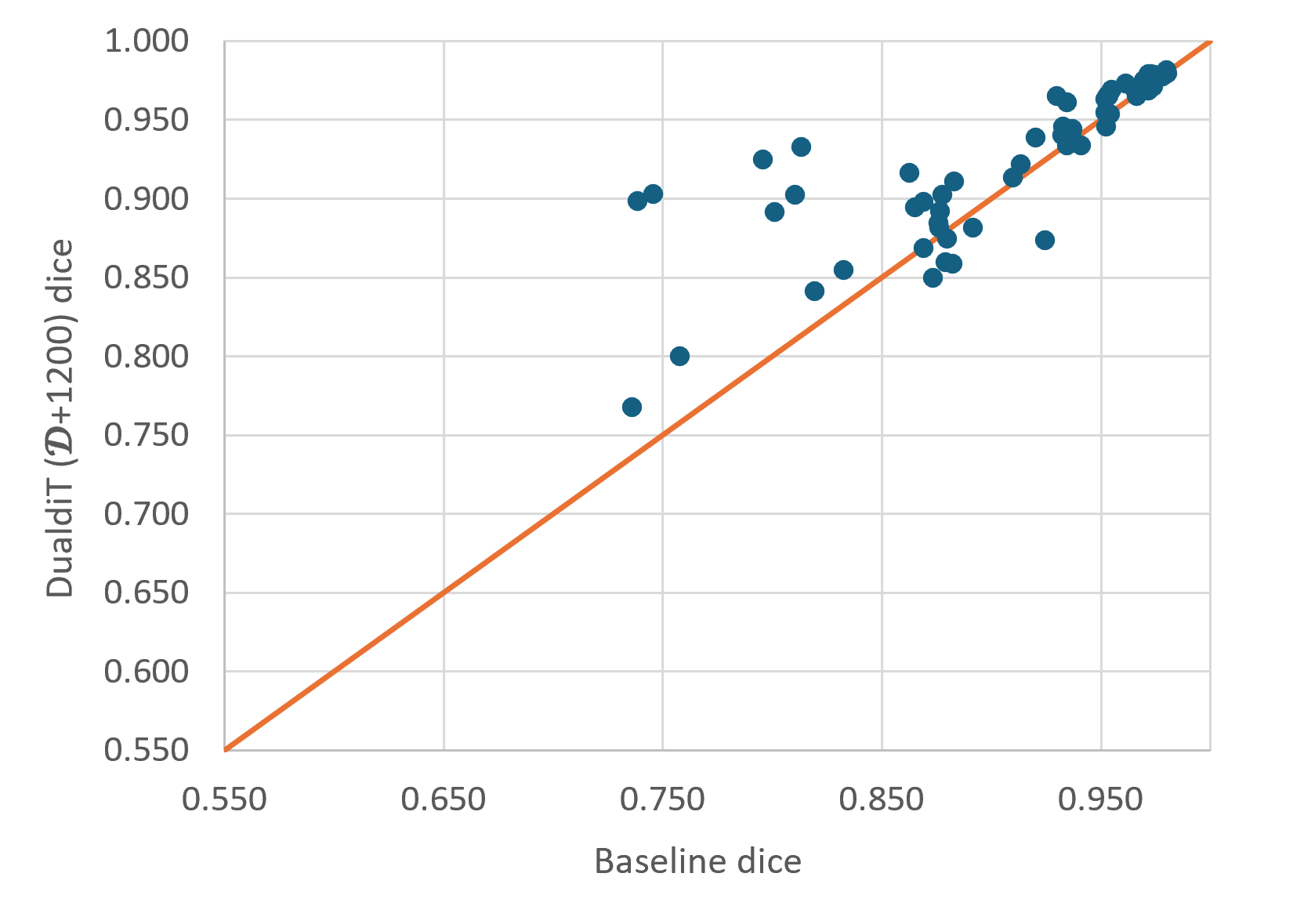}
        \caption{Per-B-scan Dice scores obtained with the Baseline and DualDiT trained with $\mathcal{D}+1200$. The diagonal line denotes equal performance; points above it indicate higher Dice scores for DualDiT.}
        \label{fig:scatter_dice}
    \end{subfigure}
    \hfill
    \begin{subfigure}[t]{0.47\textwidth}
        \centering
        \includegraphics[width=\linewidth]{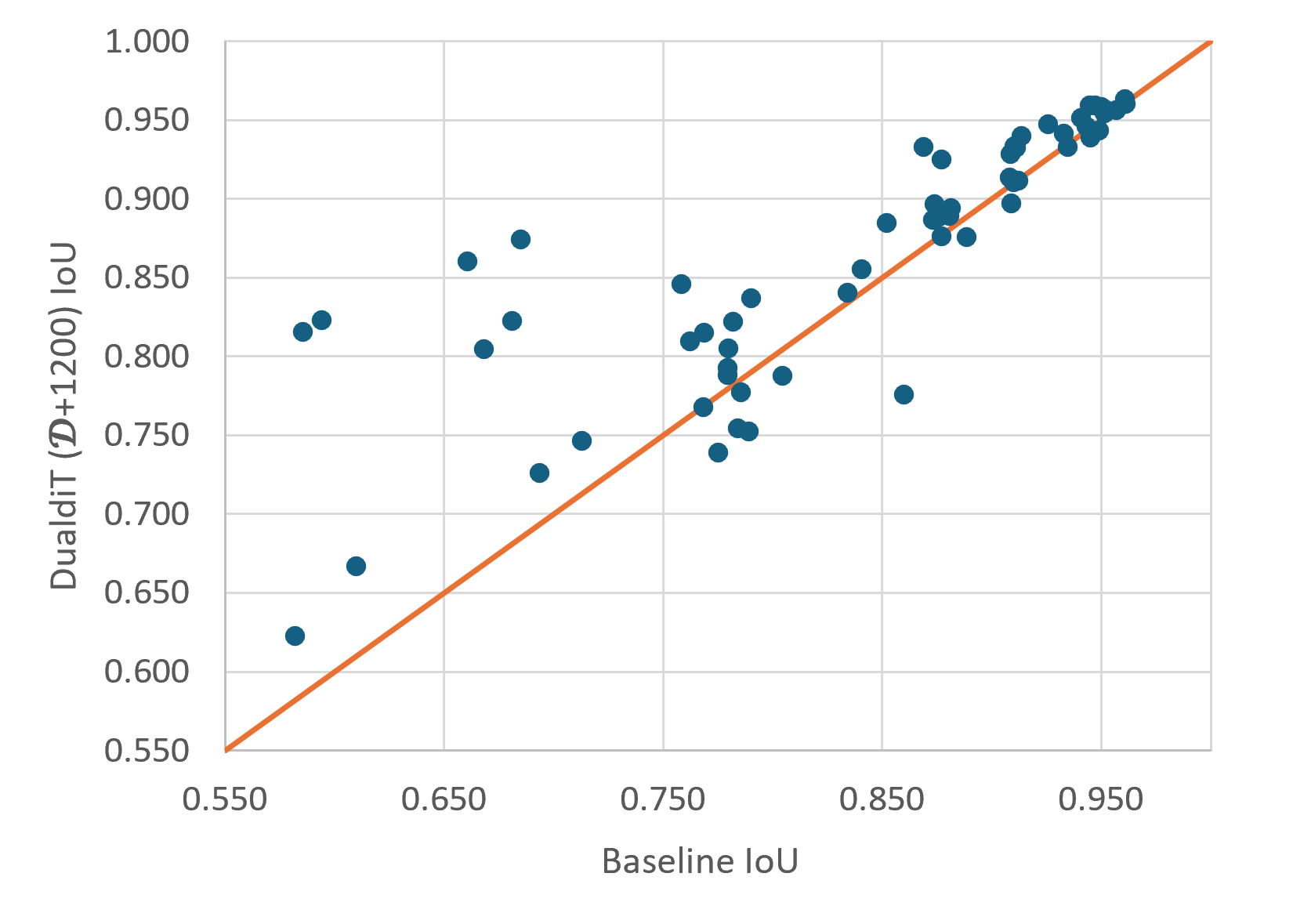}
        \caption{Per-B-scan IoU scores obtained with the Baseline and DualDiT trained with $\mathcal{D}+1200$. The diagonal line denotes equal performance; points above it indicate higher IoU scores for DualDiT.}
        \label{fig:scatter_iou}
    \end{subfigure}

    \vspace{4mm}

    \begin{subfigure}[t]{\textwidth}
        \centering
        \includegraphics[width=\linewidth]{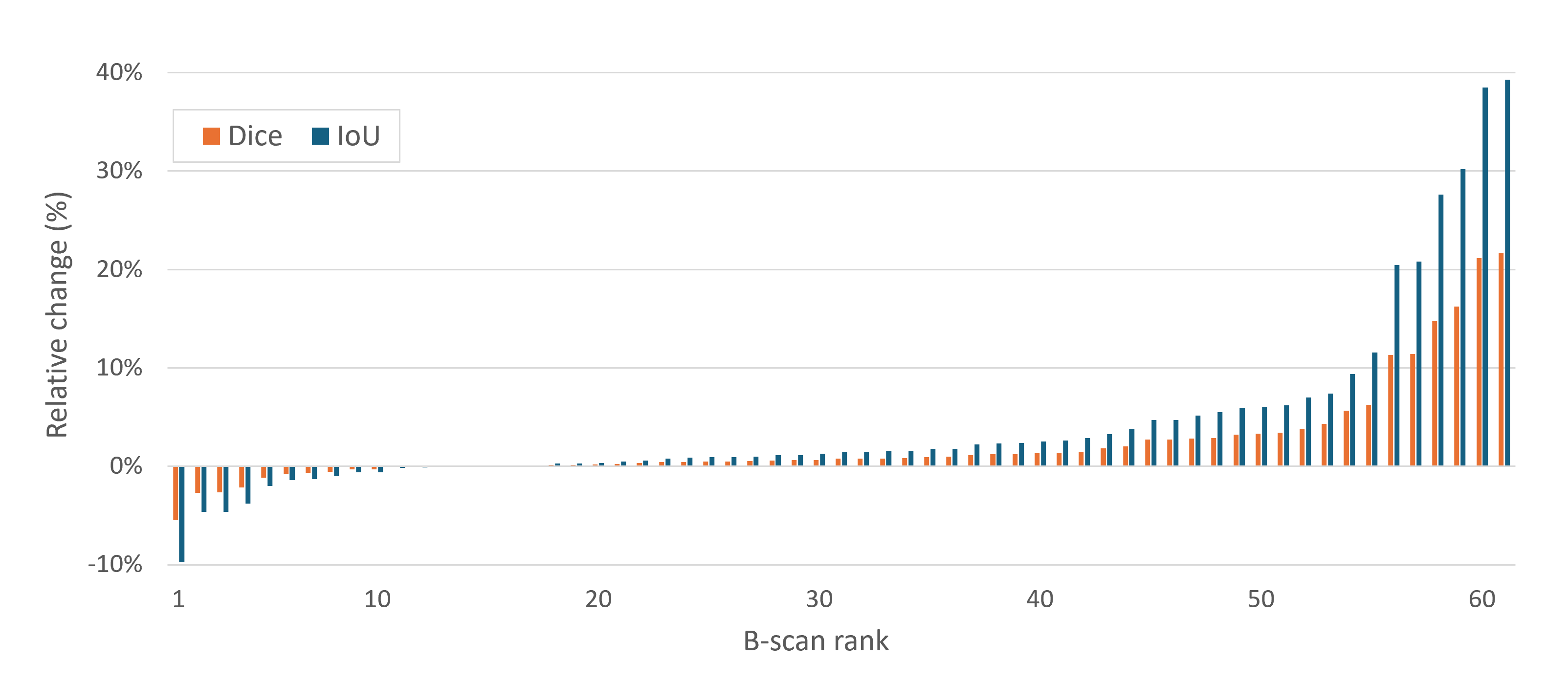}
        \caption{Relative change in Dice and IoU obtained with DualDiT with respect to the Baseline for each test B-scan ($n=61$), ordered by increasing IoU change. Positive values indicate an improvement, whereas negative values indicate a decrease in performance.}
        \label{fig:relative_changes}
    \end{subfigure}

    \caption{Per-B-scan comparison between the Baseline and DualDiT trained with $\mathcal{D}+1200$. The scatter plots show the paired Dice and IoU scores for the complete test set, with the identity line representing equal performance. The bar chart shows the relative percentage change produced by DualDiT for each B-scan, revealing that performance improves for most samples, although decreases are observed in a small number of cases.}
    \label{fig:appendix_dualdit_results}
\end{figure*}








\end{document}